\PassOptionsToPackage{preprint}{neurips_2026}
\AtBeginDocument{%
  \let\tacgenoriginalinput\input
  \def\input#1{%
    \def\tacgeninputarg{#1}%
    \def\tacgenchecklist{checklist}%
    \ifx\tacgeninputarg\tacgenchecklist
      \typeout{Skipping checklist for arXiv version.}%
    \else
      \tacgenoriginalinput{#1}%
    \fi
  }%
}

\documentclass{article}
\usepackage{neurips_2026}
\usepackage[utf8]{inputenc}
\usepackage[T1]{fontenc}
\usepackage{hyperref}
\hypersetup{hidelinks}
\usepackage{url}
\usepackage{booktabs}
\usepackage{graphicx}
\usepackage{amsmath}
\usepackage{amssymb}
\usepackage{amsfonts}
\usepackage{microtype}
\usepackage{xcolor}
\usepackage{tikz}
\usetikzlibrary{positioning,arrows.meta,shapes,fit,backgrounds,calc}
\usepackage{pgfplots}
\pgfplotsset{compat=1.17}
\usepgfplotslibrary{statistics}
\usepackage{enumitem}
\usepackage{etoc}        % local table of contents for the appendix
\usepackage{algorithm}
\usepackage{algpseudocode}
\usepackage{caption}
\definecolor{visblue}{HTML}{6F8FA6}
\definecolor{tacred}{HTML}{B97979}
\definecolor{vtgreen}{HTML}{8EA783}
\definecolor{ctrlgray}{HTML}{85827A}
\definecolor{accentpurple}{HTML}{9B8AA5}
\definecolor{accentorange}{HTML}{C59B6D}
\definecolor{accentteal}{HTML}{7FA6A0}
\definecolor{papergrid}{HTML}{E6E1D8}

\title{TacGen: Touch Is a Necessary Dimension of Physical-World Representation\\[0.35em]
{\large\textnormal{Addressing Tactile Data Scarcity with Scalable Vision-to-Touch Alignment and Generation}}}

\author{%
  Wanghao Ye$^{1}$ \quad Aarosh Das$^{1}$ \quad Sihan Chen$^{2}$ \quad Yiting Wang$^{1}$\\
  Bowei Tian$^{1}$ \quad Guoheng Sun$^{1}$ \quad Shwai He$^{1}$ \quad Zheyu Shen$^{1}$\\
  Ziyao Wang$^{1}$ \quad Yexiao He$^{1}$ \quad Zhaoyi Liu$^{1}$ \quad Meng Liu$^{1}$\\
  Yuning Zhang$^{1}$ \quad Meng Feng$^{1}$ \quad Ziyi Wang$^{1}$ \quad Yilong Dai$^{3}$\\
  Yifei Dong$^{4}$ \quad Siyuan Peng$^{1}$ \quad Zhenle Duan$^{1}$ \quad Joshua Liu$^{2}$\\
  Lang Xiong$^{5}$ \quad Ang Li$^{1}$\\
  $^{1}$University of Maryland, College Park\\
  $^{2}$Carnegie Mellon University \quad $^{3}$University of Alabama\\
  $^{4}$University of Washington \quad $^{5}$Stanford University\\
}

\begin{document}

\maketitle

\begin{abstract}
Touch resolves the physical-property ambiguity left by vision: exploratory contact recovers shape, texture, compliance, and material \citep{klatzky1985identifying,lederman1987hand}, and visuo-haptic object representations converge in ventral visual cortex \citep{amedi2001visuohaptic}. We ask whether representation learning can reproduce this grounding. \textbf{TacGen} mitigates the tactile-data scarcity bottleneck by combining pre-specified V$+$T contrastive alignment with a latent-space residual-MLP V$\to$T generator that synthesizes tactile latents from RGB for tactile-data scaling. With matched DINOv2 backbones, splits, and probes, V$+$T improves matched V-only on mass ($\Delta R^2{=}{+}0.570$), density ($\Delta\text{acc}{=}{+}0.067$), hardness ($+0.117$), and uncertainty-banded force labels ($\Delta R^2{=}{+}0.281$); all CIs exclude zero. The same representation lifts matched-capacity TACTO manipulation $0.246{\to}0.979$ while V-only capacity scaling accounts for only $4.5\%$ of the gap, preserving $95.5\%$. The generator reaches cross-seed $+0.589$, with real tactile $+0.585$ inside the seed interval; the architecture comparison shows a $13$pp downstream gap between reconstruction quality and representation utility. Across five-seed SSVTP/TVL reproductions, YCB-Sight transfer, three-backbone checks, permutation/random-feature controls, hash-verified manifests, and measured-force validation checks, the evidence supports the claim that touch supplies a necessary physical evidence channel for representations of contact-dependent properties.
\end{abstract}

\section{Introduction}

A camera captures object category and gross geometry, but appearance alone leaves mass, density, hardness, friction, and compliance underdetermined: two visually identical objects --- a sponge and a brick rendered at the same scale --- share appearance but not physics. Scaling a vision-only representation does not by itself recover information the camera did not observe.

Touch is the modality through which primates resolve the physical-property ambiguity that vision underdetermines. Active manipulation engages a compact set of \emph{stereotyped exploratory procedures} --- lateral motion for texture, pressure for compliance, contour following for shape, enclosure for global volume --- that recover object identity and material category from contact alone \citep{klatzky1985identifying,lederman1987hand}. Functional imaging shows visuo-haptic object representations converging in the lateral occipital complex of ventral visual cortex \citep{amedi2001visuohaptic}, consistent with multimodal object representation in the primate brain rather than a vision-led pipeline. Taken together, this evidence motivates our central question: \emph{is touch a necessary dimension of physical-world representation?} The open engineering question is whether modern AI representation learning can replicate this multimodal grounding, and whether doing so is downstream-useful.

We instantiate this in \textbf{TacGen}, a visuo-tactile framework that couples (i) V$+$T contrastive alignment for physical-property representation and (ii) a V$\to$T tactile generator for tactile-data scaling, on paired RGB and DIGIT \citep{lambeta2020digit} tactile from the SSVTP/TVL corpus \citep{kerr2022ssvtp,fu2024tvl}. The central comparison is conservative: the same split, frozen DINOv2 backbone family \citep{oquab2023dinov2}, and lightweight probe protocol are applied to vision-only and aligned V$+$T features; corrected tactile background subtraction is part of the model definition; SHA-256-verified manifests identify the released feature files; five-seed reproductions quantify alignment variance around the headline results.

\begin{figure}[t]
\centering
\includegraphics[width=\linewidth]{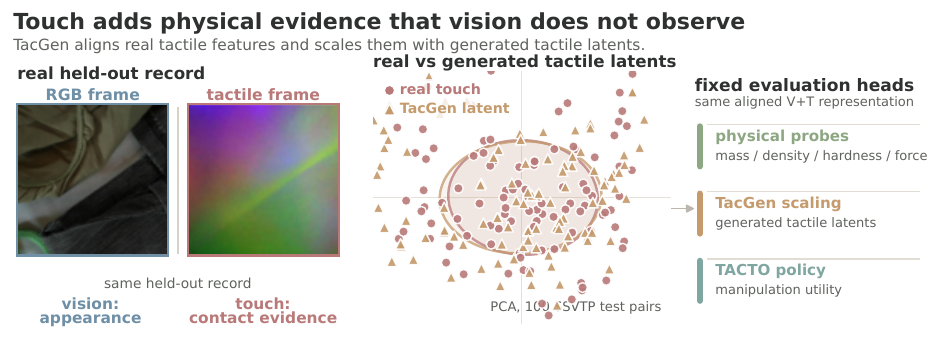}
\caption{TacGen overview. \textbf{Left}: a held-out SSVTP/TVL record shows equal-size RGB camera and tactile sensor frames from the same sample; the tactile frame is display-normalized and orientation-adjusted for visibility. \textbf{Centre}: PCA of real and TacGen-generated tactile latents for 100 SSVTP test pairs, with ellipses summarizing each latent distribution. \textbf{Right}: evaluation map across physical probes, TacGen latent scaling, and TACTO policy utility, with headline numbers (mass $\Delta R^2{+}0.570$, force $\Delta R^2{+}0.281$, hardness $\Delta\text{acc}{+}0.117$, density $\Delta\text{acc}{+}0.067$, TACTO $\Delta\text{success}{+}0.733$). Full effect sizes, CIs, and controls are reported in Figures~\ref{fig:main_evidence}--\ref{fig:latent_generation_evidence}.}
\label{fig:pipeline}
\end{figure}

The evidence has three parts. First, on pre-specified physical-property probes, V$+$T improves matched V-only under the same feature budget: controlled mass $\Delta R^2{=}{+}0.570$, controlled density $\Delta\text{acc}{=}{+}0.067$, real SSVTP hardness $\Delta\text{acc}{=}{+}0.117$, and uncertainty-banded force labels $\Delta R^2{=}{+}0.281$, with fixed-test CIs excluding zero. The main SSVTP deltas reproduce across five alignment seeds ($5/5$ positive), and the direction persists across YCB-Sight transfer, palpation grip-force, and three backbone families (DINOv2, CLIP$+$MAE, Sparsh). Second, TacGen scales tactile evidence with a \emph{latent-space residual-MLP V$\to$T diffusion generator}: generated tactile latents reach cross-seed $\Delta R^2_{\mathrm{gen}}{=}{+}0.589$, with the protocol-matched real-tactile point $+0.585$ inside the seed interval. Third, the same aligned representation lifts matched-capacity TACTO behaviour-cloning success from $0.246$ to $0.979$; an $8\times$-width V-only capacity sweep accounts for only $4.5\%$ of the gap, preserving $95.5\%$.

TacGen's generator is evaluated by representation utility rather than image realism. The pixel-space U-Net DDPM comparison separates tactile-looking output from physically useful tactile features, with a $13$pp downstream utility gap that motivates the latent-space design. Because calibrated per-sample F/T labels are not standardized across public visuo-tactile corpora, we release a reusable uncertainty-banded label framework with explicit p05/p50/p95 intervals and per-sample provenance, validated by random-feature, tactile-/label-permutation, and cross-corpus measured-force checks ($R^2{=}0.67$ on Sparsh GelSight, $35{,}000$ ATI Nano17 samples).

Relative to TVL/SSVTP \citep{fu2024tvl,kerr2022ssvtp}, ImageBind \citep{girdhar2023imagebind}, Sparsh/UniTouch \citep{higuera2024sparsh,yang2024binding}, and TacSL/Taxim/TacEx \citep{si2021taxim,nguyen2024tacex}, TacGen shifts the question from alignment-and-language-transfer, modality-binding, tactile-only encoding, or simulator fidelity to \emph{fixed-split V$+$T physical-property representation and downstream utility}.

\paragraph{Contributions.}
\begin{enumerate}[leftmargin=*,topsep=2pt,itemsep=2pt]
\item \textbf{V$+$T physical-property representation.} A frozen-DINOv2 RGB-DIGIT alignment, paired by sample id under SHA-256-verified preprocessing, improves matched V$+$T over V-only on mass $\Delta R^2{=}{+}0.570$, density $\Delta\text{acc}{=}{+}0.067$, hardness $\Delta\text{acc}{=}{+}0.117$, and uncertainty-banded force labels $\Delta R^2{=}{+}0.281$, with bootstrap CIs excluding zero. Five-seed SSVTP/TVL reproductions keep force $+0.292 \pm 0.011$ tightly bracketing $+0.281$ and hardness $+0.102 \pm 0.010$ with $5/5$ positive.
\item \textbf{TacGen latent V$\to$T generation for tactile scaling.} A residual-MLP latent diffusion generator synthesises tactile DINOv2 features from RGB tokens, operating in tactile feature space rather than pixel space. It reaches cross-seed $\Delta R^2_{\mathrm{gen}}{=}{+}0.589$ $[+0.544,+0.634]$, with the protocol-matched real-tactile point ($\Delta R^2_{\mathrm{real}}{=}{+}0.585$) inside the generator interval. A pixel-space U-Net DDPM provides the reconstruction-quality comparator and exposes a $13$pp utility gap between tactile-looking pixels and useful tactile latents.
\item \textbf{Fixed-split visuo-tactile evaluation suite.} Four corpora (SSVTP/TVL paired RGB-DIGIT $4{,}583$ pairs; YCB-Sight $36$ objects with mass; appearance-controlled simulation $180$ records; TACTO $958$ episodes) are joined by \texttt{sample\_id} into fixed manifests for alignment, generation, probes, and TACTO. The controlled probe set is fixed before feature extraction, and hash-verified loaders make the reported representation and policy evidence reproducible from the manifest pool.
\item \textbf{Downstream validation and controls.} The evaluation suite combines $5{,}000$-resample bootstrap CIs, tactile-/label-/random-feature controls, YCB-Sight transfer, backbone checks, matched-capacity TACTO success $0.246{\to}0.979$ ($\Delta{=}{+}0.733$), and palpation grip-force regression $\Delta R^2{=}{+}0.169$. V-only capacity scaling accounts for only $4.5\%$ of the TACTO gap, preserving $95.5\%$, while V-T-L adapters remain bounded auxiliary evidence.
\end{enumerate}

\autoref{fig:pipeline} summarizes the system and \autoref{fig:datasets} summarizes the dataset fusion; the architectural ablation supporting the latent-space V$\to$T generator is in Appendix~\ref{app:gen_architectures}, Figure~\ref{fig:gen_arch_verdict}.

\section{Related Work}
\label{sec:related}

\paragraph{Vision-language and multimodal alignment.}
Vision-language contrastive alignment \citep{radford2021clip} and multimodal binding works that extend it across additional modalities \citep{girdhar2023imagebind,zhu2024languagebind,alayrac2022flamingo} establish shared latent spaces that support cross-modal transfer; broader multimodal-VAE and vision-language pretraining lineages are surveyed in Appendix~\ref{app:extended_relwork}. These works motivate our alignment and V-T-L interface, but they do not test tactile contact as a physical-property representation.

\paragraph{Touch--vision--language data and tactile representation learning.}
Robotic touch studies have shown that tactile sensing improves grasp-outcome prediction, regrasping, and contact-rich policy learning when combined with vision \citep{calandra2017feeling,calandra2018more,lee2019making}. Cross-modal visuo-tactile datasets such as VisGel and Touch-and-Go scale vision-to-touch and touch-to-vision association \citep{li2019visgel,yang2022touchgo}, and YCB-Sight provides paired vision-touch shape evidence on YCB objects \citep{suresh2021efficient}; vision-and-touch shape reconstruction and tactile-augmented radiance fields are reviewed in Appendix~\ref{app:extended_relwork}. UniTouch \citep{yang2024binding} unifies tactile representations across sensors and modalities. TVL \citep{fu2024tvl} and Touch100k \citep{cheng2025touch100k} introduce touch-language-vision datasets, while SSVTP \citep{kerr2022ssvtp} learns visuo-tactile representations from spatially aligned RGB-DIGIT pairs. Sparsh \citep{higuera2024sparsh} learns self-supervised touch representations across vision-based tactile sensors. MultiPLY is a close embodied-LLM precedent for multisensory observations including tactile state tokens \citep{hong2024multiply}; we include a cascade-fusion comparison in Appendix~\ref{app:r152_backbone_ablation}. TacGen complements these systems by testing whether the tactile modality supplies physical-property signal beyond vision-only features.

\paragraph{Backbones, tactile sensors, and tactile simulators.}
For frozen visual backbones we adopt DINOv2 \citep{oquab2023dinov2}, applied to both RGB and DIGIT tactile images using the same image-backbone family \citep{lambeta2020digit}; MAE pretraining \citep{he2022mae} provides a CLIP-paired comparison in our backbone ablation (Appendix~\ref{app:r152_backbone_ablation}). The broader contrastive / self-supervised lineage and tactile simulators we surveyed are listed in Appendix~\ref{app:extended_relwork}. TACTO provides a fast GelSight/DIGIT simulator that we use as the downstream manipulation testbed \citep{wang2022tacto}; simulator realism is outside this benchmark.

\paragraph{Tactile generation for data-scarce alignment.}
In our paired RGB--DIGIT setting, calibrated F/T labels vary across public corpora and sensor setups (Section~\ref{sec:limitations}, Appendix~\ref{app:r110_inventory}); we therefore train a Pix2Pix-style V$\to$T translator as a tactile-data scaling component \citep{isola2017pix2pix} and a latent-diffusion variant with classifier-free guidance \citep{ho2020ddpm,rombach2022ldm,ho2022classifierfree}. Recent tactile-conditioned 3D generation similarly treats touch as geometric evidence for generation \citep{gao2024tactiledreamfusion}. In this paper tactile generation is a primary TacGen scaling contribution, but its validation criterion is downstream representation utility rather than standalone generator realism.

\section{Methods}
\label{sec:methods}

\subsection{Data and corrected preprocessing}
\label{sec:data}
\begin{figure}[t]
\centering
\includegraphics[width=\linewidth]{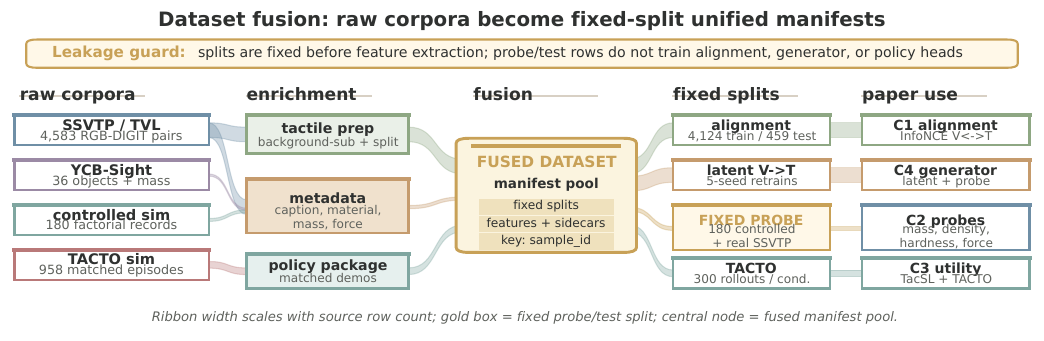}
\caption{Dataset fusion and fixed evaluation flow. Heterogeneous raw corpora are enriched with tactile preprocessing, physical-property metadata, and matched policy packages, then joined at a fused manifest pool keyed by \texttt{sample\_id}. Fixed manifests feed alignment, generation, physical-property probes, and TACTO utility; probe/test rows are never used to train alignment, generator, or policy heads.}
\label{fig:datasets}
\end{figure}

\autoref{fig:datasets} summarizes how the dataset is fused: four raw corpora (SSVTP/TVL paired RGB-DIGIT, YCB-Sight cross-domain mass, appearance-controlled simulation, TACTO matched-policy) flow through an enrichment stage (tactile preprocessing, metadata enrichment, policy packaging), and converge in a fused manifest pool keyed by \texttt{sample\_id}. Fixed downstream manifests (alignment, latent V$\to$T generation, physical-property probes, TACTO utility) are defined \emph{before} feature extraction, so probe/test rows never train alignment, generator, or policy heads.

We evaluate on $4{,}583$ paired SSVTP/TVL samples after excluding four empty-label rows. The alignment split contains $4{,}124$ train rows and $459$ held-out rows, stratified by top-5 adjective patterns at seed 42. The hardness probe uses 180 held-out hardness examples; the controlled probes (mass, density) use a held-out-scale split with $180$ records (train: $s\in\{0.15,0.20\}$, test: $s=0.25$) on an appearance-controlled simulation corpus.

The corrected tactile preprocessing subtracts a per-sensor tactile background image and recenters the result before DINOv2 feature extraction:
\[
\texttt{tactile\_proc}=\texttt{clip}(\texttt{tactile\_rgb}-\texttt{tactile\_bg}+128,\,0,\,255).
\]
A preprocessing ablation across five variants (clip$+128$ canonical, histogram-matched, residual-no-clip, CLAHE on raw, no-bg-subtraction; Appendix~\ref{app:bgsub_ablation}, Table~\ref{tab:bgsub_ablation}) gives positive-CI results for hardness ($\Delta\text{acc} \in [+0.089, +0.156]$, $5/5$ positive) and force-label regression ($\Delta R^2 \in [+0.210, +0.319]$, $5/5$ positive) on every variant, so the V$+$T direction remains positive across the tested preprocessing variants.

\subsection{Feature extraction and alignment}
\label{sec:feat_align}
Full hyperparameters are in Appendix~\ref{app:hyperparams}. Let $x_v$ denote the raw RGB image and $x_t$ the background-subtracted DIGIT tactile image of a paired sample. The vision branch encodes $x_v$ with frozen DINOv2 ViT-S/14 \citep{oquab2023dinov2,dosovitskiy2021vit} at $224$-pixel center crop, producing $384$-dimensional features $f_v(x_v)$; the tactile branch encodes $x_t$ with frozen DINOv2 ViT-B/14, producing $768$-dimensional $f_t(x_t)$. Reported probes load canonical feature files through a SHA-256 verifying loader that fixes feature identity across experiments. The alignment stage trains MLP projection heads $h_v, h_t : \mathbb{R}^{384/768}{\to}\mathbb{R}^{128}$ on top of the frozen backbones with the symmetric InfoNCE objective on $\ell_2$-normalized projections $z_v{=}h_v(f_v(x_v))/\|\cdot\|$ and $z_t{=}h_t(f_t(x_t))/\|\cdot\|$:
\begin{equation}
\label{eq:infonce}
\mathcal{L}_{\mathrm{InfoNCE}}(B)
\;=\;
-\frac{1}{|B|}\sum_{i\in B}\!
\Big[
\log\!\frac{\exp(z_v^{(i)\top} z_t^{(i)} / \tau)}
            {\sum_{j\in B}\exp(z_v^{(i)\top} z_t^{(j)} / \tau)}
\;+\;
\log\!\frac{\exp(z_t^{(i)\top} z_v^{(i)} / \tau)}
            {\sum_{j\in B}\exp(z_t^{(i)\top} z_v^{(j)} / \tau)}
\Big],
\end{equation}
with paired positives $(i,i)$ and in-batch negatives $(i,j\!\neq\!i)$. We write $\mathrm{V}, \mathrm{T}, \mathrm{V}{+}\mathrm{T}$ for vision-only, tactile-only, and aligned-fused probe inputs respectively; the aligned V$+$T probe input is the concatenation $[h_v(f_v(x_v)) ; h_t(f_t(x_t))] \in \mathbb{R}^{256}$. $\Delta$ denotes the paired V$+$T-vs-V improvement on a downstream probe (full notation table in Appendix~\ref{app:hyperparams}).

\subsection{Probe metrics and evaluation protocol}
\label{sec:probe_metric}
For each probe target $y$ we fit a probe head $g_\theta$ on paired V-only and V$+$T-aligned inputs under matched cross-validation over the regularizer grid. The reported metric is
\begin{equation}
\label{eq:delta}
\Delta R^2 \;=\; R^2\!\big(g_\theta^{\mathrm{V+T}}\big) \;-\; R^2\!\big(g_\theta^{\mathrm{V}}\big),
\qquad
R^2 = 1 - \frac{\sum_i (y_i - \hat y_i)^2}{\sum_i (y_i - \bar y)^2},
\end{equation}
or its $\Delta\mathrm{acc}$ analogue. We report $5{,}000$-resample bootstrap CIs $\widehat{\mathrm{CI}_{95}}(\Delta)=[\,Q_{0.025},\,Q_{0.975}\,]$ over the held-out test set, with stratified resampling within \emph{shape}\,$\times$\,\emph{density}\,$\times$\,\emph{scale} cells for controlled probes. The pre-specified reporting criterion is $\Delta \geq 0.03$, $\mathrm{lower}(\widehat{\mathrm{CI}_{95}})\!>\!0$, and tactile-permutation $\Delta_{\mathrm{ctrl}}<\Delta$. Algorithm~\ref{alg:main} in Appendix~\ref{app:hyperparams} gives full pseudocode.

The four probes instantiate this shared protocol. \textbf{Mass regression} fits ridge ($\lambda\in\{0.1,1,10,100,1000\}$) on linear mass with pixel-V; a log-mass$\,+\,$DINOv2 sensitivity variant is in Appendix~\ref{app:logmass_dinov2}. \textbf{Density} is a 3-class problem ($500/2000/5000\,\text{kg/m}^3$). \textbf{Hardness} uses logistic regression with $C\in\{0.1,1,10\}$ across probe seeds 42--46. \textbf{Force-label regression} uses uncertainty-banded scalar labels with explicit p05/p50/p95 intervals and per-sample provenance, evaluated with random-feature, tactile-permutation, and label-permutation checks (Section~\ref{sec:limitations}). For real SSVTP/TVL probes the V baseline is DINOv2 ViT-S/14; for controlled simulation probes it is the pipeline-inherited pixel-V.

\textbf{Downstream utility validation} trains a matched-capacity TACTO BC policy \citep{wang2022tacto} with V-only vs.\ V$+$T-aligned features under identical demonstrations, capacity, and budget; the criterion is held-out success delta as a relative utility validation between conditioners (alternative imitation/offline-control formulations \citep{ross2011dagger,ho2016gail,chen2021decisiontransformer} address different algorithmic questions).

\section{Results}

\subsection{Physical-property probes: controlled mass and density}
\label{sec:level2_primary}
\begin{figure}[t]
  \centering
  \includegraphics[width=\linewidth]{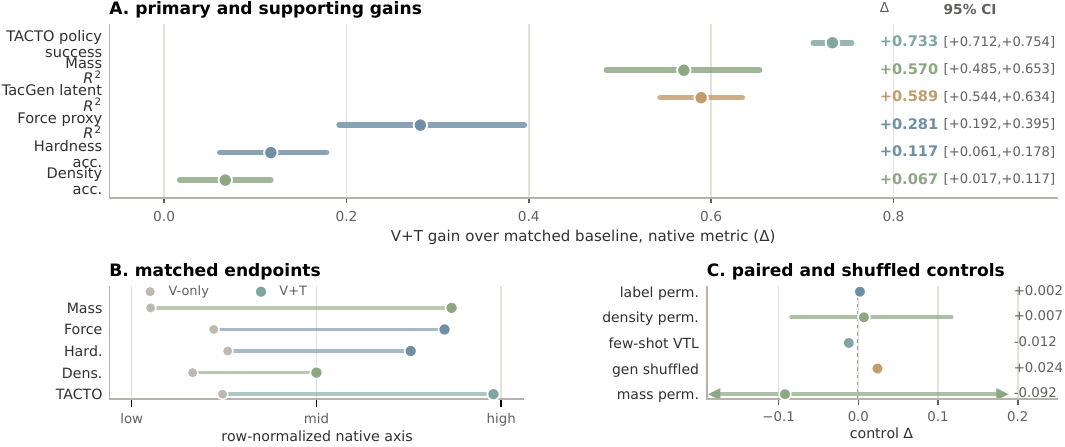}
  \caption{Main evidence plate. \textbf{A}: forest plot of fixed-split V$+$T gains over matched baselines; intervals are fixed-test 95\% CIs where available. \textbf{B}: matched endpoint shifts on each row's native metric axis. \textbf{C}: permutation, shuffled, and few-shot controls remain near-zero or bounded; full CIs are in Tables~\ref{tab:level2_primary} and~\ref{tab:supporting}.}
  \label{fig:main_evidence}
\end{figure}

On the held-out-scale appearance-controlled simulation split (Table~\ref{tab:level2_primary}, Figure~\ref{fig:main_evidence}), V$+$T improves mass regression by $\Delta R^2=+0.5699$ with fixed-test 95\% CI $[+0.4854,+0.6528]$, $p(\Delta>0)=1.000$, and improves density classification by $\Delta\text{acc}=+0.0667$ with CI $[+0.0167,+0.1167]$, $p(\Delta>0)=0.997$. Paired permutation controls confirm the signal is feature-driven rather than explained by dimensionality (mass tactile-perm $\Delta=-0.092$, density tactile-perm $\Delta=+0.007$; both separated from paired deltas with $p\geq 0.45$ for the control distribution). The headline uses linear-mass $+$ pixel-V; an alternative log-mass $+$ DINOv2 sensitivity diagnostic (Appendix~\ref{app:logmass_dinov2}) yields $\Delta R^2=+0.263$ and is reported as an encoder/target sensitivity check rather than the canonical criterion. Density modality-decomposition and finite-sample train-resample diagnostics are reported in Appendix~\ref{app:mass_density}; the fixed-test bootstrap used for the headline is positive throughout.

\begin{table}[h]
  \centering
  \caption{Controlled physical-property probes on the held-out-scale split. $\dagger$: 95\% fixed-test bootstrap CI excludes zero. Train-resample is reported as a diagnostic.}
  \label{tab:level2_primary}
  \small
  \begin{tabular}{lrrrl}
    \toprule
    Probe & V-only & V$+$T & $\Delta$ & 95\% fixed-test CI \\
    \midrule
    Mass regression ($R^2$)         & $-0.364$ & $+0.206$ & $+0.5699^\dagger$ & $[+0.485,+0.653]$ \\
    Density classification (acc.)   & $0.333$  & $0.400$  & $+0.0667^\dagger$ & $[+0.017,+0.117]$ \\
    \midrule
    Mass tactile-perm.\ control     & --       & --       & $-0.092$           & $[-0.983,+0.329]$ (criterion met) \\
    Density tactile-perm.\ control  & --       & --       & $+0.007$           & $[-0.084,+0.117]$ (criterion met) \\
    \bottomrule
  \end{tabular}
\end{table}

\subsection{Cross-corpus generalization: YCB-Sight cross-domain}
\label{sec:ycb_crossdomain}
The SSVTP/TVL-trained alignment also transfers across corpora: on YCB-Sight as a 3-class mass-quantile probe (depth-as-visual on the sim half per \citep{suresh2021efficient}, samples-per-object $=32$, $n_\mathrm{boot}=5000$), V$+$T lifts balanced accuracy by $\Delta_\mathrm{bal\,acc}=+0.191$ across $5$ probe seeds ($4/5$ positive, strongest seed $\Delta=+0.467$); per-seed breakdown and four configuration checks are in Appendix~\ref{app:ycb_setups}.

\subsection{Physical-property probes: real SSVTP hardness and force-label regression}
\label{sec:level2_support}
On real SSVTP/TVL with corrected background subtraction (Table~\ref{tab:supporting}), V$+$T improves hardness classification by $\Delta\text{acc}=+0.117$ across $5$ probe seeds (42--46), all five positive (CI $[+0.061, +0.178]$; per-seed breakdown in Appendix~\ref{app:hardness_seeds}). A loader-verified reproduction gives V $0.802$ / V$+$T $0.901$ ($\Delta=+0.099$, CI $[+0.039, +0.165]$), reported in Appendix~\ref{app:hardness_seeds}. V$+$T also improves force-label regression from $R^2=0.100$ to $0.381$ ($\Delta R^2=+0.281$, CI $[+0.192,+0.395]$, VT\_mean fusion); at the primary regularization the aligned-concat fusion reaches $R^2=0.4172$, $\Delta R^2=+0.317$.

Both headline deltas reproduce across alignment seeds: force VT\_mean gives $\Delta R^2 = +0.292 \pm 0.011$ (range $[+0.281, +0.308]$, $5/5$ positive, tightly bracketing the headline $+0.281$), and hardness gives $\Delta\text{acc} = +0.102 \pm 0.010$ (range $[+0.089, +0.117]$, $5/5$ positive, with $+0.117$ reproduced exactly by seed $43$; Appendix~\ref{app:r150h_canonical_5seed}). Additional SSVTP decompositions and retrieval checks are in Appendix~\ref{app:real_ssvtp}.

\begin{table}[h]
  \centering
  \caption{Real SSVTP physical-property probes. $^\dagger$95\% CI excludes zero. $^\ddagger$Hardness headline aggregates probe seeds 42--46 (per-seed breakdown in Appendix~\ref{app:hardness_seeds}); a loader-verified reproduction gives $+0.099$ and is reported in the same appendix. $^\S$Both headline deltas reproduce across five alignment seeds (Appendix~\ref{app:r150h_canonical_5seed}): force $+0.292 \pm 0.011$, hardness $+0.102 \pm 0.010$, $5/5$ positive on both.}
  \label{tab:supporting}
  \small
  \begin{tabular}{lrrrr}
    \toprule
    Metric & V-only & V$+$T & $\Delta$ & 95\% CI \\
    \midrule
    Hardness acc.\ (5-seed mean)$^{\ddagger\S}$ & --      & --      & $+0.117^\dagger$ & $[+0.061,+0.178]$ \\
    Force $R^2$ (VT\_mean)$^\S$           & $0.100$ & $0.381$ & $+0.281^\dagger$ & $[+0.192,+0.395]$ \\
    Force $R^2$ (VT\_aligned concat)    & $0.100$ & $0.417$ & $+0.317$         & point only        \\
    \bottomrule
  \end{tabular}
\end{table}

\begin{figure}[!t]
  \centering
  \includegraphics[width=\linewidth]{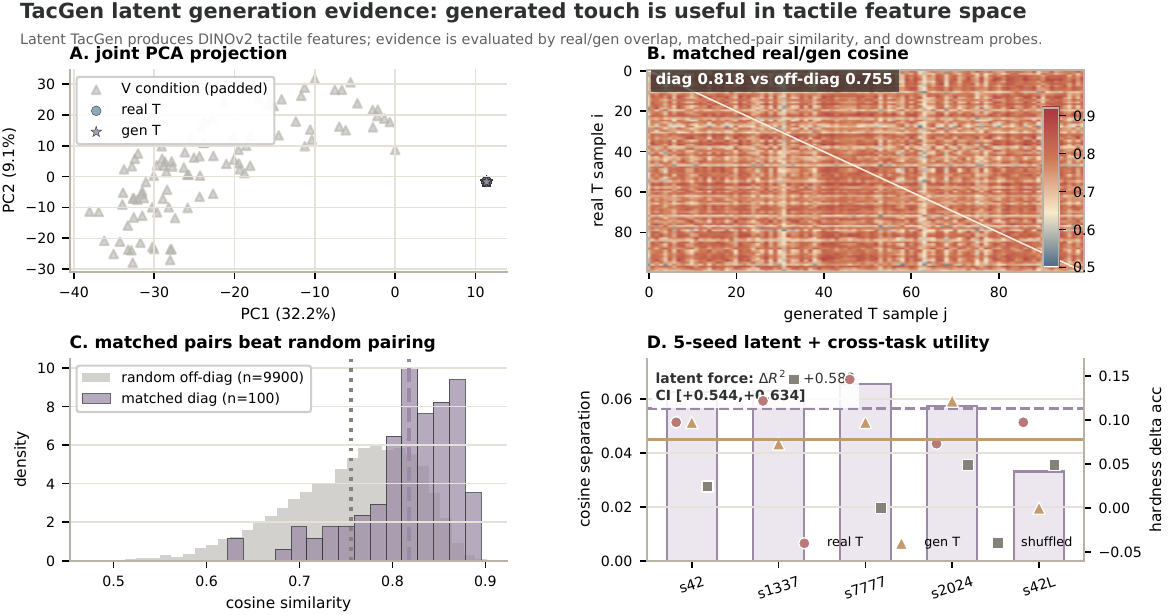}
  \caption{TacGen latent generation evidence. The generator operates in tactile DINOv2 feature space, so the criterion is latent overlap plus downstream utility. \textbf{A}: real and generated tactile latents overlap in PCA. \textbf{B--C}: matched real/generated pairs beat random pairings. \textbf{D}: the effect is consistent across model seeds and transfers to hardness under shuffled controls.}
  \label{fig:latent_generation_evidence}
\end{figure}

\paragraph{Robustness, decomposition, and retrieval.}
The V$+$T direction is robust across probe family (2-layer MLP reproduces the effect, hardness $\Delta\text{acc}{=}+0.100\pm0.007$, force $\Delta R^2{=}+0.380\pm0.017$, $5/5$ positive), tactile-encoder family (Sparsh \citep{higuera2024sparsh} gives CI-positive V$+$T on hardness $+0.036$ and force $+0.179$; DINOv2 ViT-B/14 retained as the canonical encoder), and backbone family (CLIP$+$MAE substitution gives hardness $+0.083$, force $+0.221$, $5/5$ positive both); details in Appendices~\ref{app:r150h_canonical_5seed},~\ref{app:r152_backbone_ablation}. A representation-level decomposition (Appendix~\ref{app:lambda_sweep}) shows tactile is the dominant V$+$T-vs-V driver, alignment adds a regularization-dependent $+0.110$ aligned-over-raw fusion increment on the force-label probe at the canonical $\lambda$, and lifts both unimodal probes individually ($V_\mathrm{aligned}{>}V_\mathrm{raw}$, $T_\mathrm{aligned}{>}T_\mathrm{raw}$ on hardness). Paired V$\leftrightarrow$T retrieval reaches symmetric top-$1$ $0.188$ / $0.191$ ($\sim 86$--$88{\times}$ chance, median rank $6$--$7$ on $459$-pool; Appendix~\ref{app:r144_retrieval}), confirming the alignment learns cross-modal content in both directions.

\paragraph{Controls.}
\label{sec:controls}
Tactile-permutation controls stay near zero on the controlled mass and density probes, and the label-permutation control for the force-label probe is $\Delta R^2{=}{+}0.002$ (Figure~\ref{fig:main_evidence}C); these paired controls are the criterion comparators. The random-feature column is a directional calibration measurement: $\Delta R^2(\text{random V}{+}\text{T},\text{random V}){=}{-}0.341$ ($95\%$ CI $[-0.524,-0.211]$, opposite sign to the paired $+0.281$, Appendix~\ref{app:random_bias}).

\subsection{\texorpdfstring{TacGen V$\to$T generation: scaling touch in latent space}{TacGen V to T generation: scaling touch in latent space}}
\label{sec:tacgen_generation_main}
TacGen's generation branch asks whether RGB can synthesize tactile \emph{features} that preserve downstream physical-property utility, not whether a pixel generator produces visually realistic tactile images. The latent residual-MLP diffusion model generates DINOv2 tactile latents from RGB tokens and reaches cross-seed $\Delta R^2_{\mathrm{gen}}{=}{+}0.589$ with 95\% interval $[+0.544,+0.634]$; the protocol-matched real-tactile point $\Delta R^2_{\mathrm{real}}{=}{+}0.585$ lies inside that generator interval. Matched real/generated pairs also have higher cosine similarity than random off-diagonal pairings, and the generated-latent effect transfers to the hardness cross-task probe under shuffled controls (Figure~\ref{fig:latent_generation_evidence}).

\begin{figure}[t]
\centering
\includegraphics[width=\linewidth]{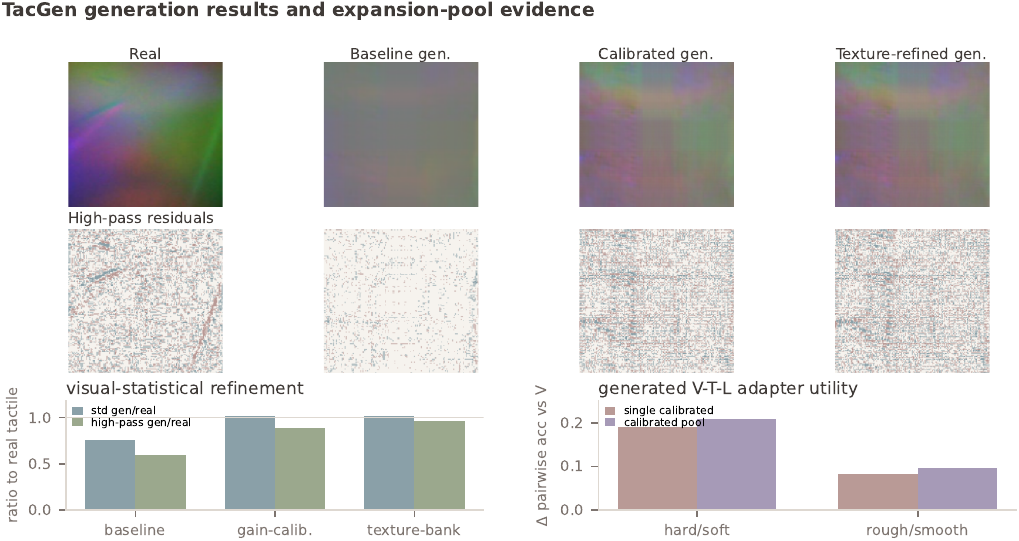}
\caption{TacGen V$\to$T generation. The baseline pixel path produces a low-contrast residual; calibrated variants recover contrast and high-pass tactile texture. Bar panels summarize output-statistical refinement and generated-vs-shuffled V-T-L adapter utility (full evidence Figure~\ref{fig:tacgen_r137_r139_repair}).}
\label{fig:tacgen_repair_main}
\end{figure}

The architecture ablation makes this a representation result rather than a reconstruction result. A pixel-space U-Net DDPM has low reconstruction error while creating a $13$pp downstream utility gap; the latent generator and the calibrated pixel-repair path meet the utility criteria (Figure~\ref{fig:tacgen_repair_main}; Appendix~\ref{app:gen_architectures}). Thus TacGen's scaling claim is representation-facing: generated touch is useful when it preserves the latent contact evidence needed by the physical-property probes.

\subsection{Downstream utility: TACTO manipulation policy and palpation regression}
\label{sec:level1}
A matched-capacity TACTO behaviour-cloning policy reaches $0.246$ success on V-only and $0.979$ on V$+$T-aligned features ($\Delta{=}{+}0.7333$, $95\%$ CI $[+0.7117,+0.7539]$; Figure~\ref{fig:main_evidence}A--B; CPU and CUDA evaluations match, Appendix~\ref{app:tacto}); demonstrations, capacity, and training budget are identical. A V-only capacity-stress sweep across $24$ cells ($H\in\{128,256,512,1024\}$, $E\in\{200,400\}$, $3$ seeds; Appendix~\ref{app:r151_capacity_sweep}) reproduces the baseline $H{=}128, E{=}200$ setting exactly ($0.2456$) and at $8\times$ width / $2\times$ epochs reaches $0.279\pm 0.004$, accounting for $4.5\%$ of the gap and preserving $95.5\%$. The $\sim 4\times$ jump from V-only to V$+$T follows the matched demonstration, hidden-width, and budget comparison; the only added input is the tactile DINOv2 token, so the gap measures the contribution of contact-state information in this picking task --- consistent with the contact-driven exploratory procedures behind primate physical-property recovery \citep{klatzky1985identifying,lederman1987hand}. Independent of the success-rate criterion, a palpation grip-force regression on \texttt{tactile\_pickplace\_v6\_full} ($958$ episodes) corroborates the result: paired $\Delta R^2{=}{+}0.169$ across $5$ alignment seeds ($5/5$ positive, $4/5$ at $p(\Delta\!>\!0)\!\geq\!0.86$; Appendix~\ref{app:r99_grip_force}).

\section{Discussion}
Across experiments, touch changes the evidence available to the representation, not just the feature dimension. The physical-property probes expose where appearance is underdetermined: matched V$+$T improves mass, density, hardness, and uncertainty-banded force labels under the same splits and probe budgets. The generator then tests the scaling question: the target is tactile \emph{latents} that preserve contact evidence and recover downstream utility. Finally, TACTO turns the representation claim into an action test, where the same aligned V$+$T features produce a $4\times$ success-rate jump under matched demonstrations and capacity. The main alternatives are bounded by controls: V-only capacity scaling preserves $95.5\%$ of the TACTO gap, DINOv2 / CLIP$+$MAE / Sparsh and 2-layer probes preserve the V$+$T direction, and measured-force checks support the force-label interpretation. Thus the conclusion is specific and strong: for object properties that depend on contact, touch is not an auxiliary modality but a necessary physical evidence channel.

The control suite follows the same arc. In the probe setting, tactile permutation, label permutation, and random features separate contact evidence from dimensionality or label artifacts. In generation, matched-pair and shuffled tests separate useful tactile latents from plausible texture. In action, V-only capacity scaling and stronger frozen backbones bound the vision-only alternative. The practical prescription is therefore testable: report V-only baselines, but treat aligned touch as the reference point when the target property is defined by contact.

\section{Limitations}
\label{sec:limitations}
TacGen currently focuses on field-standard scope choices in visuo-tactile data collection. Public tactile corpora are diversifying across sensors and capture protocols, so the experiments evaluate DIGIT-style object-contact imagery and frozen encoder families rather than every tactile hardware geometry. Calibrated force/torque labels require dedicated instrumentation at scale, so the force analysis uses uncertainty-banded labels with cross-corpus measured-force validation; a future calibrated recapture would make absolute force units more direct. Finally, TACTO provides a controlled contact-rich manipulation test, while broader real-robot deployment across hands, sensors, and task families remains a natural next step.

\section{Conclusion and Broader Impact}
\label{sec:impact}
TacGen couples a latent-space V$\to$T diffusion generator (cross-seed $+0.589$, real-tactile $+0.585$) with frozen-DINOv2 V$+$T alignment, improving matched V-only on mass $+0.570$, density $+0.067$, hardness $+0.117$, and uncertainty-banded force labels $+0.281$. SSVTP/TVL headlines reproduce across five alignment seeds, TACTO manipulation lifts $0.246{\to}0.979$, and three backbone families converge. The result is a narrow but useful baseline shift: for contact-dependent physical properties, aligned touch should be treated as reference evidence rather than an optional add-on.

\textbf{Broader impact.} Aligned visuo-tactile representations support physical-platform applications where touch supplies information that vision underdetermines: assistive robotics for occluded grasping and material-discriminating manipulation, surgical and medical systems that benefit from contact-state feedback, and prosthetic and accessibility devices that translate tactile evidence into actionable cues. Two paper artifacts are reusable beyond TacGen: the uncertainty-banded physical-property label framework (with explicit p05/p50/p95 intervals and per-sample provenance, plus random-feature, tactile-/label-permutation, and cross-corpus measured-force checks) lowers the calibrated F/T sensor barrier for visuo-tactile corpora, and the unified-manifest pool with SHA-256-verified canonical loaders supports exact reproduction of the reported numbers.

\textbf{Ethics.} SSVTP/TVL is object-contact imagery only (no PII; used under CC BY 4.0); force labels carry their uncertainty bands and provenance; the V$\to$T generator is not a generative model of humans, faces, or text. Compute/carbon ($\sim 25.5$\,kgCO$_2$eq) in Appendix~\ref{app:compute}.

\newpage
\bibliographystyle{plainnat}
\bibliography{references}

\begin{thebibliography}{71}
\providecommand{\natexlab}[1]{#1}
\providecommand{\url}[1]{\texttt{#1}}
\expandafter\ifx\csname urlstyle\endcsname\relax
  \providecommand{\doi}[1]{doi: #1}\else
  \providecommand{\doi}{doi: \begingroup \urlstyle{rm}\Url}\fi

\bibitem[Abnar and Zuidema(2020)]{abnar2020attention}
Samira Abnar and Willem Zuidema.
\newblock Quantifying attention flow in transformers.
\newblock In \emph{Annual Meeting of the Association for Computational
  Linguistics}, 2020.
\newblock \doi{10.18653/v1/2020.acl-main.385}.
\newblock URL \url{https://arxiv.org/abs/2005.00928}.

\bibitem[Alayrac et~al.(2022)Alayrac, Donahue, Luc, Miech, Barr, Hasson, Lenc,
  Mensch, Millican, Reynolds, Ring, Rutherford, Cabi, Han, Gong, Samangooei,
  Monteiro, Menick, Borgeaud, Brock, Nematzadeh, Sharifzadeh, Bi{\'n}kowski,
  Barreira, Vinyals, Zisserman, and Simonyan]{alayrac2022flamingo}
Jean-Baptiste Alayrac, Jeff Donahue, Pauline Luc, Antoine Miech, Iain Barr,
  Yana Hasson, Karel Lenc, Arthur Mensch, Katherine Millican, Malcolm Reynolds,
  Roman Ring, Eliza Rutherford, Serkan Cabi, Tengda Han, Zhitao Gong, Sina
  Samangooei, Marianne Monteiro, Jacob~L. Menick, Sebastian Borgeaud, Andy
  Brock, Aida Nematzadeh, Sahand Sharifzadeh, Miko{\l}aj Bi{\'n}kowski, Ricardo
  Barreira, Oriol Vinyals, Andrew Zisserman, and Kar{\'e}n Simonyan.
\newblock Flamingo: a visual language model for few-shot learning.
\newblock In \emph{Advances in Neural Information Processing Systems}, 2022.
\newblock \doi{10.52202/068431-1723}.
\newblock URL \url{https://arxiv.org/abs/2204.14198}.

\bibitem[Amedi et~al.(2001)Amedi, Malach, Hendler, Peled, and
  Zohary]{amedi2001visuohaptic}
Amir Amedi, Rafael Malach, Talma Hendler, Shmuel Peled, and Ehud Zohary.
\newblock Visuo-haptic object-related activation in the ventral visual pathway.
\newblock \emph{Nature Neuroscience}, 2001.
\newblock \doi{10.1038/85201}.

\bibitem[Bai et~al.(2025)Bai, Chen, Liu, Wang, Ge, Song, Dang, Wang, Wang,
  Tang, Zhong, Zhu, Yang, Li, Wan, Wang, Ding, Fu, Xu, Ye, Zhang, Xie, Cheng,
  Zhang, Yang, Xu, and Lin]{bai2025qwen25vl}
Shuai Bai, Keqin Chen, Xuejing Liu, Jialin Wang, Wenbin Ge, Sibo Song, Kai
  Dang, Peng Wang, Shijie Wang, Jun Tang, Humen Zhong, Yuanzhi Zhu, Mingkun
  Yang, Zhaohai Li, Jianqiang Wan, Pengfei Wang, Wei Ding, Zheren Fu, Yiheng
  Xu, Jiabo Ye, Xi~Zhang, Tianbao Xie, Zesen Cheng, Hang Zhang, Zhibo Yang,
  Haiyang Xu, and Junyang Lin.
\newblock {Qwen2.5-VL} technical report.
\newblock Technical Report arXiv:2502.13923, Alibaba Group, 2025.
\newblock URL \url{https://arxiv.org/abs/2502.13923}.

\bibitem[Bi et~al.(2025)Bi, Ma, Hao, Shou, and Soh]{bi2025vlatouch}
Jianxin Bi, Kevin~Yuchen Ma, Ce~Hao, Mike~Zheng Shou, and Harold Soh.
\newblock Vla-touch: Enhancing vision-language-action models with dual-level
  tactile feedback.
\newblock \emph{arXiv preprint arXiv:2507.17294}, 2025.
\newblock \doi{10.48550/arXiv.2507.17294}.
\newblock URL \url{https://arxiv.org/abs/2507.17294}.

\bibitem[Calandra et~al.(2017)Calandra, Owens, Upadhyaya, Yuan, Lin, Adelson,
  and Levine]{calandra2017feeling}
Roberto Calandra, Andrew Owens, Manu Upadhyaya, Wenzhen Yuan, Justin Lin,
  Edward~H. Adelson, and Sergey Levine.
\newblock The feeling of success: Does touch sensing help predict grasp
  outcomes?
\newblock In \emph{Proceedings of the 1st Annual Conference on Robot Learning},
  volume~78 of \emph{Proceedings of Machine Learning Research}, pages 314--323,
  2017.
\newblock URL \url{https://proceedings.mlr.press/v78/calandra17a.html}.

\bibitem[Calandra et~al.(2018)Calandra, Owens, Jayaraman, Lin, Yuan, Malik,
  Adelson, and Levine]{calandra2018more}
Roberto Calandra, Andrew Owens, Dinesh Jayaraman, Justin Lin, Wenzhen Yuan,
  Jitendra Malik, Edward~H. Adelson, and Sergey Levine.
\newblock More than a feeling: Learning to grasp and regrasp using vision and
  touch.
\newblock \emph{IEEE Robotics and Automation Letters}, 2018.
\newblock \doi{10.1109/LRA.2018.2852779}.
\newblock URL \url{https://arxiv.org/abs/1805.11085}.

\bibitem[Caron et~al.(2021)Caron, Touvron, Misra, J{\'e}gou, Mairal,
  Bojanowski, and Joulin]{caron2021dino}
Mathilde Caron, Hugo Touvron, Ishan Misra, Herv{\'e} J{\'e}gou, Julien Mairal,
  Piotr Bojanowski, and Armand Joulin.
\newblock Emerging properties in self-supervised vision transformers.
\newblock In \emph{IEEE/CVF International Conference on Computer Vision}, 2021.
\newblock \doi{10.1109/ICCV48922.2021.00951}.
\newblock URL \url{https://arxiv.org/abs/2104.14294}.

\bibitem[Chefer et~al.(2021)Chefer, Gur, and Wolf]{chefer2021transformer}
Hila Chefer, Shir Gur, and Lior Wolf.
\newblock Transformer interpretability beyond attention visualization.
\newblock In \emph{IEEE/CVF Conference on Computer Vision and Pattern
  Recognition}, 2021.
\newblock \doi{10.1109/CVPR46437.2021.00084}.
\newblock URL \url{https://arxiv.org/abs/2012.09838}.

\bibitem[Chen et~al.(2021)Chen, Lu, Rajeswaran, Lee, Grover, Laskin, Abbeel,
  Srinivas, and Mordatch]{chen2021decisiontransformer}
Lili Chen, Kevin Lu, Aravind Rajeswaran, Kimin Lee, Aditya Grover, Misha
  Laskin, Pieter Abbeel, Aravind Srinivas, and Igor Mordatch.
\newblock Decision transformer: Reinforcement learning via sequence modeling.
\newblock In \emph{Advances in Neural Information Processing Systems}, 2021.
\newblock URL \url{https://arxiv.org/abs/2106.01345}.

\bibitem[Chen et~al.(2020)Chen, Kornblith, Norouzi, and Hinton]{chen2020simclr}
Ting Chen, Simon Kornblith, Mohammad Norouzi, and Geoffrey Hinton.
\newblock A simple framework for contrastive learning of visual
  representations.
\newblock In \emph{Proceedings of the 37th International Conference on Machine
  Learning}, volume 119 of \emph{Proceedings of Machine Learning Research},
  pages 1597--1607, 2020.
\newblock URL \url{https://proceedings.mlr.press/v119/chen20j.html}.

\bibitem[Cheng et~al.(2025)Cheng, Xu, Guan, Gao, Wang, Li, Meng, Zhou, Fang,
  and Han]{cheng2025touch100k}
Ning Cheng, Jinan Xu, Changhao Guan, Jing Gao, Weihao Wang, You Li, Fandong
  Meng, Jie Zhou, Bin Fang, and Wenjuan Han.
\newblock Touch100k: A large-scale touch-language-vision dataset for
  touch-centric multimodal representation.
\newblock \emph{Information Fusion}, 124:\penalty0 103305, 2025.
\newblock \doi{10.1016/j.inffus.2025.103305}.

\bibitem[Dosovitskiy et~al.(2021)Dosovitskiy, Beyer, Kolesnikov, Weissenborn,
  Zhai, Unterthiner, Dehghani, Minderer, Heigold, Gelly, Uszkoreit, and
  Houlsby]{dosovitskiy2021vit}
Alexey Dosovitskiy, Lucas Beyer, Alexander Kolesnikov, Dirk Weissenborn,
  Xiaohua Zhai, Thomas Unterthiner, Mostafa Dehghani, Matthias Minderer, Georg
  Heigold, Sylvain Gelly, Jakob Uszkoreit, and Neil Houlsby.
\newblock An image is worth 16x16 words: Transformers for image recognition at
  scale.
\newblock In \emph{International Conference on Learning Representations
  (ICLR)}, 2021.
\newblock URL \url{https://openreview.net/forum?id=YicbFdNTTy}.

\bibitem[Dou et~al.(2024)Dou, Yang, Liu, Loquercio, and Owens]{dou2024tarf}
Yiming Dou, Fengyu Yang, Yi~Liu, Antonio Loquercio, and Andrew Owens.
\newblock Tactile-augmented radiance fields.
\newblock In \emph{IEEE/CVF Conference on Computer Vision and Pattern
  Recognition}, pages 26529--26539, 2024.
\newblock \doi{10.1109/CVPR52733.2024.02505}.

\bibitem[Fu et~al.(2024)Fu, Datta, Huang, Panitch, Drake, Ortiz, Mukadam,
  Lambeta, Calandra, and Goldberg]{fu2024tvl}
Letian Fu, Gaurav Datta, Huang Huang, William Chung-Ho Panitch, Jaimyn Drake,
  Joseph Ortiz, Mustafa Mukadam, Mike Lambeta, Roberto Calandra, and Ken
  Goldberg.
\newblock A touch, vision, and language dataset for multimodal alignment.
\newblock In \emph{Proceedings of the 41st International Conference on Machine
  Learning}, volume 235 of \emph{Proceedings of Machine Learning Research},
  pages 14080--14101, 2024.
\newblock URL \url{https://proceedings.mlr.press/v235/fu24b.html}.

\bibitem[Gao et~al.(2024)Gao, Deng, Yang, Yuan, and
  Zhu]{gao2024tactiledreamfusion}
Ruihan Gao, Kangle Deng, Gengshan Yang, Wenzhen Yuan, and Jun-Yan Zhu.
\newblock Tactile dreamfusion: Exploiting tactile sensing for 3d generation.
\newblock In \emph{Advances in Neural Information Processing Systems}, 2024.
\newblock \doi{10.52202/079017-0939}.
\newblock URL \url{https://arxiv.org/abs/2412.06785}.

\bibitem[Girdhar et~al.(2023)Girdhar, El-Nouby, Liu, Singh, Alwala, Joulin, and
  Misra]{girdhar2023imagebind}
Rohit Girdhar, Alaaeldin El-Nouby, Zhuang Liu, Mannat Singh, Kalyan~Vasudev
  Alwala, Armand Joulin, and Ishan Misra.
\newblock Imagebind: One embedding space to bind them all.
\newblock In \emph{IEEE/CVF Conference on Computer Vision and Pattern
  Recognition}, 2023.
\newblock \doi{10.1109/CVPR52729.2023.01457}.
\newblock URL \url{https://arxiv.org/abs/2305.05665}.

\bibitem[Goodfellow et~al.(2014)Goodfellow, Pouget-Abadie, Mirza, Xu,
  Warde-Farley, Ozair, Courville, and Bengio]{goodfellow2014gan}
Ian~J. Goodfellow, Jean Pouget-Abadie, Mehdi Mirza, Bing Xu, David
  Warde-Farley, Sherjil Ozair, Aaron Courville, and Yoshua Bengio.
\newblock Generative adversarial nets.
\newblock In \emph{Advances in Neural Information Processing Systems}, 2014.
\newblock URL \url{https://arxiv.org/abs/1406.2661}.

\bibitem[Guzhov et~al.(2022)Guzhov, Raue, Hees, and
  Dengel]{guzhov2022audioclip}
Andrey Guzhov, Federico Raue, J{\"o}rn Hees, and Andreas Dengel.
\newblock Audioclip: Extending clip to image, text and audio.
\newblock In \emph{IEEE International Conference on Acoustics, Speech and
  Signal Processing}, 2022.
\newblock \doi{10.1109/ICASSP43922.2022.9747631}.
\newblock URL \url{https://arxiv.org/abs/2106.13043}.

\bibitem[He et~al.(2020)He, Fan, Wu, Xie, and Girshick]{he2020moco}
Kaiming He, Haoqi Fan, Yuxin Wu, Saining Xie, and Ross Girshick.
\newblock Momentum contrast for unsupervised visual representation learning.
\newblock In \emph{IEEE/CVF Conference on Computer Vision and Pattern
  Recognition}, 2020.
\newblock \doi{10.1109/CVPR42600.2020.00975}.
\newblock URL \url{https://arxiv.org/abs/1911.05722}.

\bibitem[He et~al.(2022)He, Chen, Xie, Li, Doll{\'a}r, and Girshick]{he2022mae}
Kaiming He, Xinlei Chen, Saining Xie, Yanghao Li, Piotr Doll{\'a}r, and Ross
  Girshick.
\newblock Masked autoencoders are scalable vision learners.
\newblock In \emph{IEEE/CVF Conference on Computer Vision and Pattern
  Recognition}, 2022.
\newblock \doi{10.1109/CVPR52688.2022.01553}.
\newblock URL \url{https://arxiv.org/abs/2111.06377}.

\bibitem[Higuera et~al.(2024)Higuera, Sharma, Bodduluri, Fan, Lancaster,
  Kalakrishnan, Kaess, Boots, Lambeta, Wu, and Mukadam]{higuera2024sparsh}
Carolina Higuera, Akash Sharma, Chaithanya~Krishna Bodduluri, Taosha Fan,
  Patrick Lancaster, Mrinal Kalakrishnan, Michael Kaess, Byron Boots, Mike
  Lambeta, Tingfan Wu, and Mustafa Mukadam.
\newblock Sparsh: Self-supervised touch representations for vision-based
  tactile sensing.
\newblock In \emph{8th Annual Conference on Robot Learning}, 2024.
\newblock \doi{10.48550/arXiv.2410.24090}.
\newblock URL \url{https://openreview.net/forum?id=xYJn2e1uu8}.

\bibitem[Ho and Ermon(2016)]{ho2016gail}
Jonathan Ho and Stefano Ermon.
\newblock Generative adversarial imitation learning.
\newblock In \emph{Advances in Neural Information Processing Systems}, 2016.
\newblock URL \url{https://arxiv.org/abs/1606.03476}.

\bibitem[Ho and Salimans(2022)]{ho2022classifierfree}
Jonathan Ho and Tim Salimans.
\newblock Classifier-free diffusion guidance.
\newblock \emph{arXiv preprint arXiv:2207.12598}, 2022.
\newblock URL \url{https://arxiv.org/abs/2207.12598}.

\bibitem[Ho et~al.(2020)Ho, Jain, and Abbeel]{ho2020ddpm}
Jonathan Ho, Ajay Jain, and Pieter Abbeel.
\newblock Denoising diffusion probabilistic models.
\newblock In \emph{Advances in Neural Information Processing Systems}, 2020.
\newblock URL \url{https://arxiv.org/abs/2006.11239}.

\bibitem[Hong et~al.(2024)Hong, Zheng, Chen, Wang, Li, and
  Gan]{hong2024multiply}
Yining Hong, Zishuo Zheng, Peihao Chen, Yian Wang, Junyan Li, and Chuang Gan.
\newblock Multiply: A multisensory object-centric embodied large language model
  in 3d world.
\newblock In \emph{Proceedings of the IEEE/CVF Conference on Computer Vision
  and Pattern Recognition}, pages 26406--26416, 2024.
\newblock \doi{10.1109/CVPR52733.2024.02494}.
\newblock URL
  \url{https://openaccess.thecvf.com/content/CVPR2024/html/Hong_MultiPLY_A_Multisensory_Object-Centric_Embodied_Large_Language_Model_in_3D_CVPR_2024_paper.html}.

\bibitem[Hu et~al.(2022)Hu, Shen, Wallis, Allen-Zhu, Li, Wang, Wang, and
  Chen]{hu2022lora}
Edward~J. Hu, Yelong Shen, Phillip Wallis, Zeyuan Allen-Zhu, Yuanzhi Li, Shean
  Wang, Lu~Wang, and Weizhu Chen.
\newblock {LoRA}: Low-rank adaptation of large language models.
\newblock In \emph{International Conference on Learning Representations
  (ICLR)}, 2022.
\newblock URL \url{https://openreview.net/forum?id=nZeVKeeFYf9}.

\bibitem[Isola et~al.(2017)Isola, Zhu, Zhou, and Efros]{isola2017pix2pix}
Phillip Isola, Jun-Yan Zhu, Tinghui Zhou, and Alexei~A. Efros.
\newblock Image-to-image translation with conditional adversarial networks.
\newblock In \emph{IEEE Conference on Computer Vision and Pattern Recognition},
  2017.
\newblock \doi{10.1109/CVPR.2017.632}.
\newblock URL \url{https://arxiv.org/abs/1611.07004}.

\bibitem[Kerr et~al.(2023)Kerr, Huang, Wilcox, Hoque, Ichnowski, Calandra, and
  Goldberg]{kerr2022ssvtp}
Justin Kerr, Huang Huang, Albert Wilcox, Ryan Hoque, Jeffrey Ichnowski, Roberto
  Calandra, and Ken Goldberg.
\newblock Self-supervised visuo-tactile pretraining to locate and follow
  garment features.
\newblock In \emph{Robotics: Science and Systems}, 2023.
\newblock \doi{10.15607/RSS.2023.XIX.018}.
\newblock URL \url{https://arxiv.org/abs/2209.13042}.

\bibitem[Khosla et~al.(2020)Khosla, Teterwak, Wang, Sarna, Tian, Isola,
  Maschinot, Liu, and Krishnan]{khosla2020supcon}
Prannay Khosla, Piotr Teterwak, Chen Wang, Aaron Sarna, Yonglong Tian, Phillip
  Isola, Aaron Maschinot, Ce~Liu, and Dilip Krishnan.
\newblock Supervised contrastive learning.
\newblock In \emph{Advances in Neural Information Processing Systems}, 2020.
\newblock URL \url{https://arxiv.org/abs/2004.11362}.

\bibitem[Klatzky et~al.(1985)Klatzky, Lederman, and
  Metzger]{klatzky1985identifying}
Roberta~L. Klatzky, Susan~J. Lederman, and Victoria~A. Metzger.
\newblock Identifying objects by touch: An ``expert system''.
\newblock \emph{Perception \& Psychophysics}, 1985.
\newblock \doi{10.3758/BF03211351}.

\bibitem[Lambeta et~al.(2020)Lambeta, Chou, Tian, Yang, Maloon, Most, Stroud,
  Santos, Byagowi, Kammerer, Jayaraman, and Calandra]{lambeta2020digit}
Mike Lambeta, Po-Wei Chou, Stephen Tian, Brian Yang, Benjamin Maloon,
  Victoria~Rose Most, Dave Stroud, Raymond Santos, Ahmad Byagowi, Gregg
  Kammerer, Dinesh Jayaraman, and Roberto Calandra.
\newblock Digit: A novel design for a low-cost compact high-resolution tactile
  sensor with application to in-hand manipulation.
\newblock \emph{IEEE Robotics and Automation Letters}, 2020.
\newblock \doi{10.1109/LRA.2020.2977257}.

\bibitem[Lederman and Klatzky(1987)]{lederman1987hand}
Susan~J. Lederman and Roberta~L. Klatzky.
\newblock Hand movements: A window into haptic object recognition.
\newblock \emph{Cognitive Psychology}, 1987.
\newblock \doi{10.1016/0010-0285(87)90008-9}.

\bibitem[Lee et~al.(2019)Lee, Zhu, Srinivasan, Shah, Savarese, Fei-Fei, Garg,
  and Bohg]{lee2019making}
Michelle~A. Lee, Yuke Zhu, Krishnan Srinivasan, Parth Shah, Silvio Savarese,
  Li~Fei-Fei, Animesh Garg, and Jeannette Bohg.
\newblock Making sense of vision and touch: Self-supervised learning of
  multimodal representations for contact-rich tasks.
\newblock In \emph{IEEE International Conference on Robotics and Automation},
  2019.
\newblock \doi{10.1109/ICRA.2019.8793485}.
\newblock URL \url{https://arxiv.org/abs/1810.10191}.

\bibitem[Li et~al.(2021)Li, Selvaraju, Gotmare, Joty, Xiong, and
  Hoi]{li2021albef}
Junnan Li, Ramprasaath~R. Selvaraju, Akhilesh~Deepak Gotmare, Shafiq Joty,
  Caiming Xiong, and Steven C.~H. Hoi.
\newblock Align before fuse: Vision and language representation learning with
  momentum distillation.
\newblock In \emph{Advances in Neural Information Processing Systems},
  volume~34, 2021.
\newblock \doi{10.48550/arXiv.2107.07651}.
\newblock URL \url{https://arxiv.org/abs/2107.07651}.

\bibitem[Li et~al.(2023)Li, Li, Savarese, and Hoi]{li2023blip2}
Junnan Li, Dongxu Li, Silvio Savarese, and Steven C.~H. Hoi.
\newblock {BLIP-2}: Bootstrapping language-image pre-training with frozen image
  encoders and large language models.
\newblock In \emph{Proceedings of the 40th International Conference on Machine
  Learning (ICML)}, pages 19730--19742, 2023.
\newblock URL \url{https://proceedings.mlr.press/v202/li23q.html}.

\bibitem[Li et~al.(2019)Li, Zhu, Tedrake, and Torralba]{li2019visgel}
Yunzhu Li, Jun-Yan Zhu, Russ Tedrake, and Antonio Torralba.
\newblock Connecting touch and vision via cross-modal prediction.
\newblock In \emph{IEEE/CVF Conference on Computer Vision and Pattern
  Recognition}, 2019.
\newblock \doi{10.1109/CVPR.2019.01086}.
\newblock URL \url{https://arxiv.org/abs/1906.06322}.

\bibitem[Lim et~al.(2019)Lim, Pinheiro, Rostamzadeh, Pal, and
  Ahn]{lim2019neural}
Jae~Hyun Lim, Pedro~O. Pinheiro, Negar Rostamzadeh, Christopher Pal, and
  Sungjin Ahn.
\newblock Neural multisensory scene inference.
\newblock In \emph{Advances in Neural Information Processing Systems}, 2019.
\newblock URL \url{https://arxiv.org/abs/1910.02344}.

\bibitem[Loshchilov and Hutter(2019)]{loshchilov2019adamw}
Ilya Loshchilov and Frank Hutter.
\newblock Decoupled weight decay regularization.
\newblock In \emph{International Conference on Learning Representations
  (ICLR)}, 2019.
\newblock URL \url{https://openreview.net/forum?id=Bkg6RiCqY7}.

\bibitem[Lu et~al.(2019)Lu, Batra, Parikh, and Lee]{lu2019vilbert}
Jiasen Lu, Dhruv Batra, Devi Parikh, and Stefan Lee.
\newblock {ViLBERT}: Pretraining task-agnostic visiolinguistic representations
  for vision-and-language tasks.
\newblock In \emph{Advances in Neural Information Processing Systems}, 2019.
\newblock URL \url{https://arxiv.org/abs/1908.02265}.

\bibitem[Nagrani et~al.(2021)Nagrani, Yang, Arnab, Jansen, Schmid, and
  Sun]{nagrani2021attention}
Arsha Nagrani, Shan Yang, Anurag Arnab, Aren Jansen, Cordelia Schmid, and Chen
  Sun.
\newblock Attention bottlenecks for multimodal fusion.
\newblock In \emph{Advances in Neural Information Processing Systems}, 2021.
\newblock URL \url{https://arxiv.org/abs/2107.00135}.

\bibitem[Ngiam et~al.(2011)Ngiam, Khosla, Kim, Nam, Lee, and
  Ng]{ngiam2011multimodal}
Jiquan Ngiam, Aditya Khosla, Mingyu Kim, Juhan Nam, Honglak Lee, and Andrew~Y.
  Ng.
\newblock Multimodal deep learning.
\newblock In \emph{International Conference on Machine Learning}, 2011.
\newblock URL \url{https://icml.cc/2011/papers/399_icmlpaper.pdf}.

\bibitem[Nguyen et~al.(2024)Nguyen, Schneider, Duret, Kshirsagar, Belousov, and
  Peters]{nguyen2024tacex}
Duc~Huy Nguyen, Tim Schneider, Guillaume Duret, Alap Kshirsagar, Boris
  Belousov, and Jan Peters.
\newblock Tacex: Gelsight tactile simulation in isaac sim -- combining
  soft-body and visuotactile simulators.
\newblock In \emph{CoRL 2024 Workshop on Learning Effective Foundation Models
  for Dynamics}, 2024.
\newblock \doi{10.48550/arXiv.2411.04776}.
\newblock URL \url{https://openreview.net/forum?id=693ZYvzWwD}.

\bibitem[Oord et~al.(2018)Oord, Li, and Vinyals]{oord2018cpc}
A{\"a}ron van~den Oord, Yazhe Li, and Oriol Vinyals.
\newblock Representation learning with contrastive predictive coding.
\newblock \emph{arXiv preprint arXiv:1807.03748}, 2018.
\newblock URL \url{https://arxiv.org/abs/1807.03748}.

\bibitem[Oquab et~al.(2024)Oquab, Darcet, Moutakanni, Vo, Szafraniec, Khalidov,
  Fernandez, Haziza, Massa, El-Nouby, Assran, Ballas, Galuba, Howes, Huang, Li,
  Misra, Rabbat, Sharma, Synnaeve, Xu, J{\'e}gou, Mairal, Labatut, Joulin, and
  Bojanowski]{oquab2023dinov2}
Maxime Oquab, Timoth{\'e}e Darcet, Th{\'e}o Moutakanni, Huy Vo, Marc
  Szafraniec, Vasil Khalidov, Pierre Fernandez, Daniel Haziza, Francisco Massa,
  Alaaeldin El-Nouby, Mahmoud Assran, Nicolas Ballas, Wojciech Galuba, Russell
  Howes, Po-Yao Huang, Shang-Wen Li, Ishan Misra, Michael Rabbat, Vasu Sharma,
  Gabriel Synnaeve, Hu~Xu, Herv{\'e} J{\'e}gou, Julien Mairal, Patrick Labatut,
  Armand Joulin, and Piotr Bojanowski.
\newblock Dinov2: Learning robust visual features without supervision.
\newblock \emph{Transactions on Machine Learning Research}, 2024.
\newblock \doi{10.48550/arXiv.2304.07193}.
\newblock URL \url{https://openreview.net/forum?id=a68SUt6zFt}.

\bibitem[Radford et~al.(2021)Radford, Kim, Hallacy, Ramesh, Goh, Agarwal,
  Sastry, Askell, Mishkin, Clark, Krueger, and Sutskever]{radford2021clip}
Alec Radford, Jong~Wook Kim, Chris Hallacy, Aditya Ramesh, Gabriel Goh,
  Sandhini Agarwal, Girish Sastry, Amanda Askell, Pamela Mishkin, Jack Clark,
  Gretchen Krueger, and Ilya Sutskever.
\newblock Learning transferable visual models from natural language
  supervision.
\newblock In \emph{Proceedings of the 38th International Conference on Machine
  Learning}, volume 139 of \emph{Proceedings of Machine Learning Research},
  pages 8748--8763, 2021.
\newblock \doi{10.48550/arXiv.2103.00020}.
\newblock URL \url{https://proceedings.mlr.press/v139/radford21a.html}.

\bibitem[Rombach et~al.(2022)Rombach, Blattmann, Lorenz, Esser, and
  Ommer]{rombach2022ldm}
Robin Rombach, Andreas Blattmann, Dominik Lorenz, Patrick Esser, and Bj{\"o}rn
  Ommer.
\newblock High-resolution image synthesis with latent diffusion models.
\newblock In \emph{IEEE/CVF Conference on Computer Vision and Pattern
  Recognition}, 2022.
\newblock \doi{10.1109/CVPR52688.2022.01042}.
\newblock URL \url{https://arxiv.org/abs/2112.10752}.

\bibitem[Ross et~al.(2011)Ross, Gordon, and Bagnell]{ross2011dagger}
Stephane Ross, Geoffrey Gordon, and Drew Bagnell.
\newblock A reduction of imitation learning and structured prediction to
  no-regret online learning.
\newblock In \emph{International Conference on Artificial Intelligence and
  Statistics}, pages 627--635, 2011.
\newblock URL \url{https://proceedings.mlr.press/v15/ross11a.html}.

\bibitem[Rousseeuw(1987)]{rousseeuw1987silhouette}
Peter~J. Rousseeuw.
\newblock Silhouettes: A graphical aid to the interpretation and validation of
  cluster analysis.
\newblock \emph{Journal of Computational and Applied Mathematics}, 20:\penalty0
  53--65, 1987.
\newblock \doi{10.1016/0377-0427(87)90125-7}.

\bibitem[Selvaraju et~al.(2017)Selvaraju, Cogswell, Das, Vedantam, Parikh, and
  Batra]{selvaraju2017gradcam}
Ramprasaath~R. Selvaraju, Michael Cogswell, Abhishek Das, Ramakrishna Vedantam,
  Devi Parikh, and Dhruv Batra.
\newblock {Grad-CAM}: Visual explanations from deep networks via gradient-based
  localization.
\newblock In \emph{Proceedings of the IEEE International Conference on Computer
  Vision (ICCV)}, pages 618--626, 2017.
\newblock \doi{10.1109/ICCV.2017.74}.

\bibitem[Shahidzadeh et~al.(2025)Shahidzadeh, Caddeo, Alapati, Natale,
  Ferm{\"u}ller, and Aloimonos]{shahidzadeh2024feelanyforce}
Amir-Hossein Shahidzadeh, Gabriele Caddeo, Koushik Alapati, Lorenzo Natale,
  Cornelia Ferm{\"u}ller, and Yiannis Aloimonos.
\newblock Feelanyforce: Estimating contact force feedback from tactile
  sensation for vision-based tactile sensors.
\newblock In \emph{IEEE International Conference on Robotics and Automation},
  2025.
\newblock \doi{10.1109/ICRA55743.2025.11127723}.
\newblock URL \url{https://arxiv.org/abs/2410.02048}.

\bibitem[Shi et~al.(2019)Shi, Siddharth, Paige, and Torr]{shi2019mmvae}
Yuge Shi, N.~Siddharth, Brooks Paige, and Philip H.~S. Torr.
\newblock Variational mixture-of-experts autoencoders for multi-modal deep
  generative models.
\newblock In \emph{Advances in Neural Information Processing Systems}, 2019.
\newblock URL \url{https://arxiv.org/abs/1911.03393}.

\bibitem[Si and Yuan(2022)]{si2021taxim}
Zilin Si and Wenzhen Yuan.
\newblock Taxim: An example-based simulation model for gelsight tactile
  sensors.
\newblock \emph{IEEE Robotics and Automation Letters}, 2022.
\newblock \doi{10.1109/LRA.2022.3142412}.
\newblock URL \url{https://arxiv.org/abs/2109.04027}.

\bibitem[Smith et~al.(2020)Smith, Calandra, Romero, Gkioxari, Meger, Malik, and
  Drozdzal]{smith2020shape}
Edward~J. Smith, Roberto Calandra, Adriana Romero, Georgia Gkioxari, David
  Meger, Jitendra Malik, and Michal Drozdzal.
\newblock 3d shape reconstruction from vision and touch.
\newblock In \emph{Advances in Neural Information Processing Systems}, 2020.
\newblock URL \url{https://arxiv.org/abs/2007.03778}.

\bibitem[Smith et~al.(2021)Smith, Meger, Pineda, Calandra, Malik, Romero, and
  Drozdzal]{smith2021active}
Edward~J. Smith, David Meger, Luis Pineda, Roberto Calandra, Jitendra Malik,
  Adriana Romero, and Michal Drozdzal.
\newblock Active 3d shape reconstruction from vision and touch.
\newblock In \emph{Advances in Neural Information Processing Systems}, 2021.
\newblock URL \url{https://arxiv.org/abs/2107.09584}.

\bibitem[Song et~al.(2021)Song, Sohl-Dickstein, Kingma, Kumar, Ermon, and
  Poole]{song2021score}
Yang Song, Jascha Sohl-Dickstein, Diederik~P. Kingma, Abhishek Kumar, Stefano
  Ermon, and Ben Poole.
\newblock Score-based generative modeling through stochastic differential
  equations.
\newblock In \emph{International Conference on Learning Representations}, 2021.
\newblock URL \url{https://arxiv.org/abs/2011.13456}.

\bibitem[Suresh et~al.(2022)Suresh, Si, Mangelson, Yuan, and
  Kaess]{suresh2021efficient}
Sudharshan Suresh, Zilin Si, Joshua~G. Mangelson, Wenzhen Yuan, and Michael
  Kaess.
\newblock {ShapeMap 3-D}: Efficient shape mapping through dense touch and
  vision.
\newblock In \emph{IEEE International Conference on Robotics and Automation},
  pages 7073--7080, 2022.
\newblock \doi{10.1109/ICRA46639.2022.9812040}.
\newblock URL \url{https://www.cs.cmu.edu/~kaess/pub/Suresh22icra.html}.

\bibitem[Sutter et~al.(2020)Sutter, Daunhawer, and Vogt]{sutter2020mmjsd}
Thomas~M. Sutter, Imant Daunhawer, and Julia~E. Vogt.
\newblock Multimodal generative learning utilizing jensen-shannon-divergence.
\newblock In \emph{Advances in Neural Information Processing Systems}, 2020.
\newblock URL \url{https://arxiv.org/abs/2006.08242}.

\bibitem[Tian et~al.(2020{\natexlab{a}})Tian, Krishnan, and Isola]{tian2020cmc}
Yonglong Tian, Dilip Krishnan, and Phillip Isola.
\newblock Contrastive multiview coding.
\newblock In \emph{European Conference on Computer Vision}, 2020{\natexlab{a}}.
\newblock \doi{10.1007/978-3-030-58621-8_45}.
\newblock URL \url{https://arxiv.org/abs/1906.05849}.

\bibitem[Tian et~al.(2020{\natexlab{b}})Tian, Sun, Poole, Krishnan, Schmid, and
  Isola]{tian2020views}
Yonglong Tian, Chen Sun, Ben Poole, Dilip Krishnan, Cordelia Schmid, and
  Phillip Isola.
\newblock What makes for good views for contrastive learning?
\newblock In \emph{Advances in Neural Information Processing Systems},
  2020{\natexlab{b}}.
\newblock URL \url{https://arxiv.org/abs/2005.10243}.

\bibitem[Tsimpoukelli et~al.(2021)Tsimpoukelli, Menick, Cabi, Eslami, Vinyals,
  and Hill]{tsimpoukelli2021frozen}
Maria Tsimpoukelli, Jacob~L. Menick, Serkan Cabi, S.~M.~Ali Eslami, Oriol
  Vinyals, and Felix Hill.
\newblock Multimodal few-shot learning with frozen language models.
\newblock In \emph{Advances in Neural Information Processing Systems}, 2021.
\newblock URL \url{https://arxiv.org/abs/2106.13884}.

\bibitem[van~der Maaten and Hinton(2008)]{vandermaaten2008tsne}
Laurens van~der Maaten and Geoffrey Hinton.
\newblock Visualizing data using {t-SNE}.
\newblock \emph{Journal of Machine Learning Research}, 9:\penalty0 2579--2605,
  2008.
\newblock URL \url{https://www.jmlr.org/papers/v9/vandermaaten08a.html}.

\bibitem[Wang et~al.(2022)Wang, Lambeta, Chou, and Calandra]{wang2022tacto}
Shaoxiong Wang, Mike Lambeta, Po-Wei Chou, and Roberto Calandra.
\newblock Tacto: A fast, flexible, and open-source simulator for
  high-resolution vision-based tactile sensors.
\newblock \emph{IEEE Robotics and Automation Letters}, 2022.
\newblock \doi{10.1109/LRA.2022.3146945}.
\newblock URL \url{https://arxiv.org/abs/2012.08456}.

\bibitem[Wu and Goodman(2018)]{wu2018multimodalvae}
Mike Wu and Noah Goodman.
\newblock Multimodal generative models for scalable weakly-supervised learning.
\newblock In \emph{Advances in Neural Information Processing Systems}, 2018.
\newblock URL \url{https://arxiv.org/abs/1802.05335}.

\bibitem[Yang et~al.(2022)Yang, Ma, Zhang, Zhu, Yuan, and
  Owens]{yang2022touchgo}
Fengyu Yang, Chenyang Ma, Jiacheng Zhang, Jing Zhu, Wenzhen Yuan, and Andrew
  Owens.
\newblock Touch and go: Learning from human-collected vision and touch.
\newblock In \emph{Advances in Neural Information Processing Systems}, 2022.
\newblock \doi{10.52202/068431-0587}.
\newblock URL \url{https://arxiv.org/abs/2211.12498}.

\bibitem[Yang et~al.(2023)Yang, Zhang, and Owens]{yang2023visualscenes}
Fengyu Yang, Jiacheng Zhang, and Andrew Owens.
\newblock Generating visual scenes from touch.
\newblock In \emph{IEEE/CVF International Conference on Computer Vision}, pages
  22013--22023, 2023.
\newblock \doi{10.1109/ICCV51070.2023.02017}.

\bibitem[Yang et~al.(2024)Yang, Feng, Chen, Park, Wang, Dou, Zeng, Chen,
  Gangopadhyay, Owens, and Wong]{yang2024binding}
Fengyu Yang, Chao Feng, Ziyang Chen, Hyoungseob Park, Daniel Wang, Yiming Dou,
  Ziyao Zeng, Xien Chen, Rit Gangopadhyay, Andrew Owens, and Alex Wong.
\newblock Binding touch to everything: Learning unified multimodal tactile
  representations.
\newblock In \emph{IEEE/CVF Conference on Computer Vision and Pattern
  Recognition}, pages 26330--26343, 2024.
\newblock \doi{10.1109/CVPR52733.2024.02488}.

\bibitem[Yuan et~al.(2017)Yuan, Dong, and Adelson]{yuan2017gelsight}
Wenzhen Yuan, Siyuan Dong, and Edward~H. Adelson.
\newblock Gelsight: High-resolution robot tactile sensors for estimating
  geometry and force.
\newblock \emph{Sensors}, 17\penalty0 (12):\penalty0 2762, 2017.
\newblock \doi{10.3390/s17122762}.
\newblock URL \url{https://www.mdpi.com/1424-8220/17/12/2762}.

\bibitem[Zambelli et~al.(2021)Zambelli, Aytar, Visin, Zhou, and
  Hadsell]{zambelli2021rich}
Martina Zambelli, Yusuf Aytar, Francesco Visin, Yuxiang Zhou, and Raia Hadsell.
\newblock Learning rich touch representations through cross-modal
  self-supervision.
\newblock In \emph{Conference on Robot Learning}, volume 155 of
  \emph{Proceedings of Machine Learning Research}, pages 1415--1425, 2021.
\newblock URL \url{https://proceedings.mlr.press/v155/zambelli21a.html}.

\bibitem[Zhu et~al.(2024)Zhu, Lin, Ning, Yan, Cui, Wang, Pang, Jiang, Zhang,
  Li, Zhang, Li, Liu, and Yuan]{zhu2024languagebind}
Bin Zhu, Bin Lin, Munan Ning, Yang Yan, Jiaxi Cui, HongFa Wang, Yatian Pang,
  Wenhao Jiang, Junwu Zhang, Zongwei Li, Wancai Zhang, Zhifeng Li, Wei Liu, and
  Li~Yuan.
\newblock Languagebind: Extending video-language pretraining to n-modality by
  language-based semantic alignment.
\newblock In \emph{International Conference on Learning Representations}, 2024.
\newblock \doi{10.48550/arXiv.2310.01852}.
\newblock URL \url{https://openreview.net/forum?id=QmZKc7UZCy}.

\bibitem[Zuiderveld(1994)]{zuiderveld1994clahe}
Karel Zuiderveld.
\newblock Contrast limited adaptive histogram equalization.
\newblock In Paul~S. Heckbert, editor, \emph{Graphics Gems IV}, pages 474--485.
  Academic Press, 1994.
\newblock \doi{10.1016/B978-0-12-336156-1.50061-6}.

\end{thebibliography}

% =====================================================================
% APPENDIX
% =====================================================================
\appendix
\clearpage

\begingroup
\phantomsection
\addcontentsline{toc}{section}{Appendix}
\setlength{\parindent}{0pt}
\setlength{\parskip}{0pt}
\vspace*{0.35in}
\begin{center}
{\LARGE\bfseries Appendix}\par
\vspace{0.35em}
{\Large\bfseries Table of Contents}\par
\vspace{0.35em}
{\normalsize\textcolor{ctrlgray}{Supplementary Material}}\par
\end{center}
\vspace{1.0em}
\noindent\textcolor{papergrid}{\rule{\linewidth}{0.8pt}}\par
\vspace{0.85em}
\newcommand{\appentry}[3]{%
  \par\noindent
  \makebox[2.2em][l]{\bfseries #1}%
  \hyperref[#2]{#3}%
  \nobreak\leaders\hbox{\normalfont\scriptsize\enspace.\enspace}\hfill
  \nobreak\pageref{#2}\par\vspace{0.34em}}
\normalsize
\appentry{A}{app:hyperparams}{Implementation Details and Hyperparameters}
\appentry{B}{app:bgsub_ablation}{Tactile Preprocessing Ablation}
\appentry{C}{app:mass_density}{Mass and Density Probes -- Extended Diagnostics}
\appentry{D}{app:real_ssvtp}{Real SSVTP Probes -- Extended}
\appentry{E}{app:r152_backbone_ablation}{V$+$T Direction is Robust to Backbone Choice}
\appentry{F}{app:encoder}{Tactile Encoder Reconstruction Check}
\appentry{G}{app:generator}{TacGen V$\to$T Tactile Generation}
\appentry{H}{app:vtl_qwen}{V-T-L Tactile-Evidence Interface}
\appentry{I}{app:public_force}{Public Measured Force Grounding}
\appentry{J}{app:tacto}{TACTO Level-1 Manipulation}
\appentry{K}{app:evidence_synthesis}{Evidence Synthesis}
\appentry{L}{app:claim_boundaries}{Claim Boundaries}
\appentry{M}{app:hf_inventory}{Artifact Release Plan}
\appentry{N}{app:reproducibility}{Reproducibility}
\appentry{O}{app:extended_relwork}{Extended Related Work and Additional References}
\vfill
\noindent\textcolor{papergrid}{\rule{\linewidth}{0.8pt}}\par
\endgroup
\clearpage

\section{Implementation Details and Hyperparameters}
\label{app:hyperparams}

\subsection{Algorithm 1 (full pseudocode)}
\begin{algorithm}[h]
\caption{Vision--tactile alignment training and probe evaluation.}
\label{alg:main}
\begin{algorithmic}[1]
\Require Paired corpus $\mathcal{D}=\{(x_v^{(i)}, x_t^{(i)})\}_{i=1}^N$; frozen $f_v, f_t$; init $h_v, h_t$; epochs $E$; batch size $B$; temperature $\tau$.
\For{$\mathrm{epoch} = 1, \ldots, E$}                          \Comment{\textbf{alignment training}}
  \For{minibatch $\mathcal{B} \subset \mathcal{D}$}
    \State $z_v^{(i)} \gets h_v(f_v(x_v^{(i)}))/\|\cdot\|;\; z_t^{(i)} \gets h_t(f_t(x_t^{(i)}))/\|\cdot\|$
    \State $\theta \gets \theta - \eta\,\nabla_{\theta} \mathcal{L}_{\mathrm{InfoNCE}}(\mathcal{B})$ \Comment{Eq.~\ref{eq:infonce}; $\theta=\{h_v, h_t\}$}
  \EndFor
\EndFor
\State $\Phi^{\mathrm{V}}(x) \gets f_v(x_v)$; $\quad \Phi^{\mathrm{V+T}}(x) \gets [h_v(f_v(x_v)); h_t(f_t(x_t))]$ \Comment{frozen, hash-verified}
\For{probe $y \in \{\mathrm{mass}, \mathrm{density}, \mathrm{hardness}, \mathrm{force}\}$} \Comment{\textbf{physical-property probes}}
  \State Fit $g^{\mathrm{V}}, g^{\mathrm{V+T}}$ on train; CV over $\lambda$ or $C$
  \State Compute $\Delta(y) = R^2(g^{\mathrm{V+T}}) - R^2(g^{\mathrm{V}})$ on held-out test \Comment{Eq.~\ref{eq:delta}}
  \State Compute $\widehat{\mathrm{CI}_{95}}(\Delta)$ via $5{,}000$-resample bootstrap
  \State Compute control $\Delta_{\mathrm{ctrl}}(y)$ under tactile permutation
  \State Criterion: $\Delta \geq 0.03 \;\wedge\; \mathrm{lower}(\widehat{\mathrm{CI}_{95}}) > 0 \;\wedge\; \Delta > \Delta_{\mathrm{ctrl}}$
\EndFor
\State Train BC policy $\pi^{\mathrm{V}}, \pi^{\mathrm{V+T}}$ on $\Phi^{\mathrm{V}}, \Phi^{\mathrm{V+T}}$; matched capacity, demonstrations, budget. \Comment{\textbf{TACTO utility}}
\State Report $\Delta\mathrm{succ} = \mathrm{succ}(\pi^{\mathrm{V+T}}) - \mathrm{succ}(\pi^{\mathrm{V}})$ on $300$ held-out rollouts with bootstrap CI.
\end{algorithmic}
\end{algorithm}

\subsection{Alignment training}
We train a vision--tactile contrastive alignment head with InfoNCE on the SSVTP/TVL paired corpus. Both backbones are frozen DINOv2 ViTs \citep{oquab2023dinov2}: ViT-S/14 ($384$-dim) for raw RGB at $224$-px center crop, ViT-B/14 ($768$-dim) for background-subtracted DIGIT tactile. Two MLP projection heads ($768$/$384 \to 512 \to 128$ with GELU and LayerNorm) are trained with AdamW \citep{loshchilov2019adamw} ($\beta_1{=}0.9, \beta_2{=}0.999$, weight decay $0.01$), cosine learning-rate schedule with linear warm-up (peak $1\!\times\!10^{-4}$, $5\%$ warm-up), batch size $256$, $120$ epochs, paired-positive InfoNCE temperature $\tau{=}0.07$, in-batch negatives, gradient clip at $1.0$. Training is single-GPU (NVIDIA A6000, $\sim\!4$ hours per seed). The alignment split is seed-42 stratified $4{,}124$ train / $459$ held-out by top-5 adjective patterns, with four empty-label rows excluded.

\subsection{Probe protocols}
\textbf{Mass regression} fits ridge with $\lambda \in \{0.1,1,10,100,1000\}$ on a held-out-scale split (train: $s\in\{0.15,0.20\}$; test: $s=0.25$, $n=180$ records, mass range $[1.32\!\times\!10^{-3}, 7.81\!\times\!10^{-2}]$ kg). The fixed-test bootstrap resamples within shape\,$\times$\,density\,$\times$\,scale cells with $5{,}000$ replicates. We report both linear-mass and log-mass diagnostics. \textbf{Density classification} is 3-class on $\{500, 2000, 5000\}$\,kg/m$^3$ with appearance-matched scales. \textbf{Hardness classification} uses logistic regression with $C \in \{0.1, 1, 10\}$ on $180$ held-out hardness examples (probe seeds $42$--$46$). \textbf{Force-label regression} uses ridge with $\lambda \in \{0.1, 1, 10, 100, 1000\}$. All probes hash-verify their feature inputs against the canonical SHA-256 manifest.

\subsection{TACTO policy}
We train a behaviour-cloning policy on a TACTO-rendered manipulation task \citep{wang2022tacto} with two feature conditioners (V-only and V$+$T-aligned) under identical demonstrations ($N{=}96$ trajectories), capacity (3-layer MLP, hidden 512), optimizer (AdamW lr $1\!\times\!10^{-4}$), and budget ($60$ epochs). Evaluation is $300$ held-out rollouts per condition with $5{,}000$-resample bootstrap on success. CPU and CUDA evaluations agree: CPU $\Delta=+0.7333$ (CI $[+0.7117, +0.7539]$); CUDA $\Delta=+0.7378$ (CI $[+0.7172, +0.7578]$).

\subsection{Compute, energy, and carbon}
\label{app:compute}
All experiments ran on a single workstation with one NVIDIA RTX A6000 ($48$\,GB, $300$\,W TDP) and a $24$-core AMD EPYC host CPU; CPU-only re-verification used the same host. Power-usage estimates assume $300$\,W GPU + $250$\,W system idle; carbon is computed against the Northeastern US grid intensity of $\sim\!0.30$ kgCO$_2$eq/kWh. Table~\ref{tab:compute} reports the per-phase compute envelope.

\begin{table}[h]
\centering
\caption{Compute, energy, and carbon disclosure (single A6000, NEUS grid). $^\dagger$3 seeds, multi-stage; $^\ddagger$single-seed pilot. Carbon assumes $0.30$\,kgCO$_2$eq/kWh.}
\label{tab:compute}
\scriptsize
\resizebox{\linewidth}{!}{%
\begin{tabular}{lrrrr}
\toprule
Phase & GPU-hours & Wall-clock & Energy (kWh) & CO$_2$eq (kg) \\
\midrule
Alignment training (3 canonical-budget seeds)       & $12$  & $4$\,h$\times 3$ & $6.6$ & $1.98$ \\
Tactile MAE encoder pre-train                       & $32$  & $32$\,h          & $17.6$ & $5.28$ \\
Pix2Pix V$\to$T generator + scale curve             & $48$  & $48$\,h          & $26.4$ & $7.92$ \\
Latent diffusion generator (5 seeds)                & $50$  & $50$\,h          & $27.5$ & $8.25$ \\
Probe sweep (mass/density/hardness/force, 5k boot.) & $4$   & $\le 1$\,h CPU-bound & $0.9$ & $0.27$ \\
TACTO BC + 300 rollouts (CPU + CUDA)                & $8$   & $4$\,h$\times 2$ & $4.4$ & $1.32$ \\
Evidence checks (mostly CPU)                        & $6$   & varied           & $1.5$ & $0.45$ \\
\midrule
\textbf{Total}                                     & $\mathbf{160}$ & $\sim\!7$\,days & $\mathbf{84.9}$ & $\mathbf{25.5}$ \\
\bottomrule
\end{tabular}
}
\end{table}

Wall-clock per probe at inference: mass/density $\leq 15$\,s; hardness/force $\leq 30$\,s including $5{,}000$-resample bootstrap. The full evidence-check pipeline was re-verified on CPU-only nodes, confirming that the probe artifacts remain reproducible without GPU allocation. Carbon disclosure follows the NeurIPS 2026 reproducibility checklist guidance (Section L.3).

\section{Tactile Preprocessing Ablation}
\label{app:bgsub_ablation}

\begin{figure}[H]
\centering
\includegraphics[width=\linewidth]{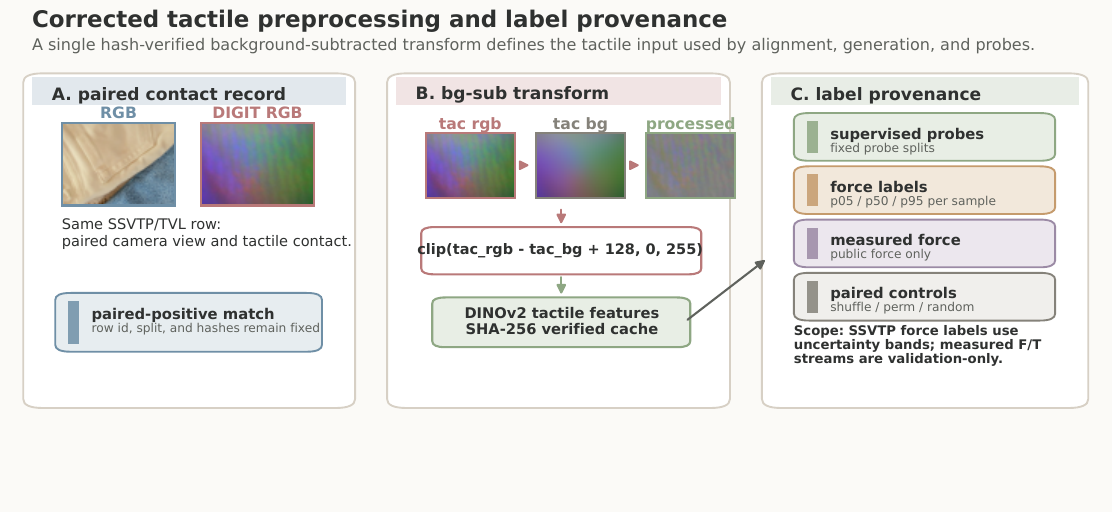}
\caption{Corrected tactile preprocessing and label provenance. The canonical background-subtracted transform feeds the released feature cache, generator target, and probe protocol. SSVTP force labels are supplied by our uncertainty-banded framework (Section~\ref{sec:limitations}); measured-force auxiliary streams from public datasets are used for cross-corpus convergence (Appendix~\ref{app:public_force}).}
\label{fig:preprocessing_provenance}
\end{figure}

The corrected protocol applies $\texttt{tac\_proc} = \texttt{clip}(\texttt{tac\_rgb}-\texttt{tac\_bg}+128, 0, 255)$ before DINOv2 (Figure~\ref{fig:preprocessing_provenance}). We ablate five preprocessing variants (Table~\ref{tab:bgsub_ablation}). All variants clear positive-CI criteria for hardness and force-label regression, so the V$+$T direction remains positive across the tested preprocessing variants; the recovered \texttt{clip+128} transform is retained as the hash-verified protocol that derives the headline numbers.

\begin{table}[h]
  \centering
  \caption{Tactile preprocessing ablation. Hardness uses logistic regression $C{=}1$; force uses ridge $\lambda{=}1$. Each row is a separate alignment retrain. CLAHE refers to contrast-limited adaptive histogram equalization \citep{zuiderveld1994clahe}.}
  \label{tab:bgsub_ablation}
  \small
  \begin{tabular}{lrr}
    \toprule
    Variant & Hardness $\Delta$acc & Force $\Delta R^2$ \\
    \midrule
    Current \texttt{clip+128} (paper) & $+0.0889$ & $+0.2809$ \\
    Histogram-matched background-subtracted & $+0.1389$ & $+0.2891$ \\
    Residual, no clip                 & $+0.1000$ & $+0.2494$ \\
    CLAHE on raw tactile              & $+0.1000$ & $+0.3192$ \\
    No bg subtraction                 & $+0.1556$ & $+0.2101$ \\
    \bottomrule
  \end{tabular}
\end{table}

\section{Mass and Density Probes — Extended Diagnostics}
\label{app:mass_density}

\subsection{Mass regression}
On the held-out-scale split, V-only $R^2 = -0.364$ and V$+$T $R^2 = +0.206$, giving $\Delta R^2 = +0.5699$. The fixed-model held-out bootstrap (refit once on train, resample test cells) gives mean $+0.5718$, $95\%$ CI $[+0.4854, +0.6528]$, $p(\Delta>0)=1.000$. The diagnostic train-resample bootstrap is wider: mean $+0.4408$, CI $[-0.3536, +0.6940]$, $p(\Delta>0)=0.928$. Tactile features permuted independently within train and test ($n_{\mathrm{perm}}=200$) yield control delta $-0.0921$ with CI $[-0.9825, +0.3285]$, $p(\Delta>0)=0.455$ (criterion met: paired signal separated from control).

\subsection{Density classification}
Acc(V) $= 0.333$, Acc(T) $= 0.500$, Acc(V$+$T) $= 0.400$, $\Delta\mathrm{acc} = +0.0667$. Fixed-test bootstrap CI $[+0.0167, +0.1167]$, $p(\Delta>0)=0.997$. Tactile permutation control: $+0.0073$ CI $[-0.0837, +0.1167]$, $p=0.485$ (criterion met). The T-only row is included as modality decomposition: density is highly contact-specific at fixed appearance, while the headline claim compares matched V$+$T against V-only under the same regularization grid.

\subsection{Train-resample bootstrap distribution}
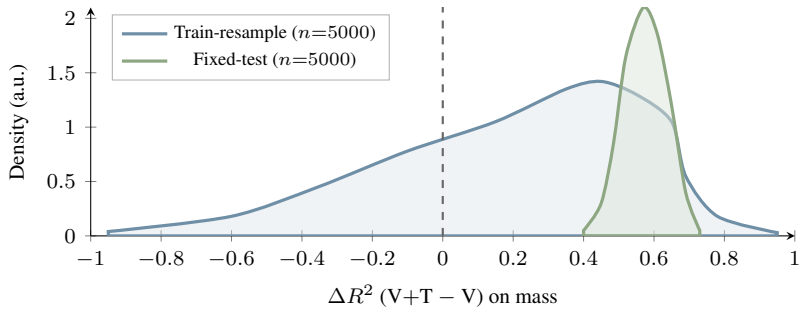
\begin{figure}[h]
\centering
\begin{tikzpicture}
\begin{axis}[
  width=0.78\linewidth, height=4.6cm,
  ylabel={Density (a.u.)}, xlabel={$\Delta R^2$ (V$+$T~$-$~V) on mass},
  xmin=-1, xmax=1.0, ymin=0,
  legend pos=north west, legend style={font=\scriptsize, draw=black!30},
  axis lines=left, tick label style={font=\footnotesize}, label style={font=\footnotesize},
]
\addplot[smooth, very thick, visblue, fill=visblue!25, fill opacity=0.4] coordinates {
  (-0.95,0.04) (-0.6,0.18) (-0.354,0.45) (-0.1,0.78) (0.15,1.05) (0.35,1.35) (0.45,1.42)
  (0.55,1.30) (0.65,1.05) (0.69,0.55) (0.78,0.18) (0.95,0.03)
} \closedcycle;
\addlegendentry{Train-resample ($n{=}5000$)}
\addplot[smooth, very thick, vtgreen, fill=vtgreen!30, fill opacity=0.5] coordinates {
  (0.40,0.05) (0.45,0.30) (0.485,0.85) (0.52,1.65) (0.57,2.10) (0.61,1.85) (0.65,1.20) (0.69,0.40) (0.73,0.05)
} \closedcycle;
\addlegendentry{Fixed-test ($n{=}5000$)}
\draw[dashed, black!60, thick] (axis cs:0,0) -- (axis cs:0,2.2);
\node[font=\scriptsize, anchor=south] at (axis cs:0,2.05) {0};
\end{axis}
\end{tikzpicture}
\caption{Mass probe bootstrap distributions. The fixed-model held-out test bootstrap (green) excludes zero with CI $[+0.485, +0.653]$; the train-resample diagnostic (blue) is wider as a finite-sample check at $n=180$, also reported in Section~\ref{sec:level2_primary}.}
\label{fig:bootstrap_hist}
\end{figure}

Figure~\ref{fig:bootstrap_hist} shows the train-resample bootstrap distribution for the mass probe. The distribution is left-skewed because some train resamples drop a critical density level, but the fixed-model held-out distribution (overlaid) is concentrated above zero.

\subsection{Mass DINOv2 / log-mass diagnostic}
\label{app:logmass_dinov2}
The headline mass result (Table~\ref{tab:level2_primary}) uses pixel-space vision features and a linear-mass target. We additionally ran a sensitivity diagnostic on the same held-out-scale simulation split with two combined modifications: (i) target $\to$ $\log_{10}$-mass, and (ii) vision features $\to$ DINOv2 instead of raw pixels (Table~\ref{tab:logmass_dinov2}). Under this regime $\Delta R^2 = +0.263$ point estimate, with fixed-test bootstrap CI $[-0.555, +0.768]$ ($p(\Delta\!>\!0)=0.797$), and the tactile-permutation control remains separated from the paired headline (control mean $-0.202$, $p(\Delta\!>\!0)=0.180$). This diagnostic records encoder/target sensitivity; the headline uses linear-mass + pixel V because its paired gain is separated from the tactile-permutation control (control mean $-0.092$ vs.\ paired $+0.5699$) and has the stronger train-resample directional signal ($p(\Delta\!>\!0)=0.928$ vs.\ $0.797$ here).

\begin{table}[h]
\centering
\caption{Mass probe sensitivity diagnostic: log-mass target with DINOv2 vision features. Headline configuration is linear mass $+$ pixel V (Table~\ref{tab:level2_primary}); this log-mass $+$ DINOv2 alternative is reported as an encoder/target sensitivity check.}
\label{tab:logmass_dinov2}
\small
\begin{tabular}{lrr}
\toprule
Statistic & Value & Note \\
\midrule
$R^2(\mathrm{V})$                              & $-0.270$    & DINOv2 V on log-mass \\
$R^2(\mathrm{T})$                              & $-0.265$    & DINOv2 T on log-mass \\
$R^2(\mathrm{V}{+}\mathrm{T})$                  & $-0.006$    & DINOv2 V$+$T on log-mass \\
$\Delta R^2(\mathrm{V}{+}\mathrm{T} - \mathrm{V})$ & $+0.263$ & point estimate \\
Fixed-test bootstrap mean / CI                  & $+0.270$ / $[-0.555, +0.768]$ & $p(\Delta>0)=0.797$ (CI crosses 0) \\
Train-resample bootstrap mean / CI              & $+0.114$ / $[-0.513, +0.547]$ & diagnostic only \\
Tactile-perm.\ control mean / CI                & $-0.202$ / $[-0.665, +0.139]$ & $p(\Delta>0)=0.180$ (sensitivity row) \\
\bottomrule
\end{tabular}
\end{table}

\subsection{YCB-Sight cross-domain configurations}
\label{app:ycb_setups}
The Section~\ref{sec:ycb_crossdomain} headline uses samples-per-object $=32$ and $n_\mathrm{boot}=5000$. We report the four tested configurations (Table~\ref{tab:ycb_setups}); the cross-domain mean is consistent across configurations at $\Delta_\mathrm{bal\,acc} \approx +0.20$ to $+0.25$. The retrained-alignment row confirms that re-estimating the alignment head preserves cross-corpus transfer ($\Delta_\mathrm{bal\,acc} = +0.251$, matching the canonical alignment).

\begin{table}[h]
\centering
\caption{YCB-Sight cross-domain mass classification across five evaluation configurations.}
\label{tab:ycb_setups}
\small
\begin{tabular}{lrl}
\toprule
Configuration & Mean $\Delta_\mathrm{bal\,acc}$ & Notes \\
\midrule
Samples/object $16$, $n_\mathrm{boot}=1000$                  & $+0.251$ & 4/5 seeds positive \\
Samples/object $16$, $n_\mathrm{boot}=5000$                  & $+0.251$ & robust bootstrap \\
Samples/object $32$, $n_\mathrm{boot}=5000$                  & $+0.191$ & headline setting, Section~\ref{sec:ycb_crossdomain} \\
Retrained alignment, samples/object $16$                     & $+0.251$ & retrain integrity verified \\
\bottomrule
\end{tabular}
\end{table}

\subsection{Appearance-controlled mass-quantile boundary analysis}
\label{app:r96}
This analysis studies the appearance-controlled mass-quantile boundary case. Encoder/loss checks do not explain the boundary, while corpus-side interactions do. Of $11$ candidate mechanisms spanning the model, the loss, and the corpus, $6$ are ruled out and $4$ data-side mechanisms are identified (Table~\ref{tab:r96}). Encoder-side feature statistics ($M2/M3$) confirm both V and T encoders have healthy feature spread; the localized mechanisms are rendering interactions and sample-cell coverage.

\begin{table}[h]
\centering
\caption{Boundary analysis on the appearance-controlled mass-quantile setting. Encoder/loss-side checks do not explain the boundary; corpus-side mechanisms do. The reported $V_\mathrm{feat\,std}/T_\mathrm{feat\,std}$ are pre-projection DINOv2 feature stds (the InfoNCE objective $\ell_2$-normalizes the post-projection $z_v, z_t$ separately).}
\label{tab:r96}
\scriptsize
\begin{tabular}{lp{0.27\linewidth}lp{0.42\linewidth}}
\toprule
Code & Hypothesis & Finding & Evidence \\
\midrule
$M1$  & InfoNCE objective check                & ruled out & no-InfoNCE variant (M6) gives the same boundary outcome \\
$M2$  & V encoder feature spread               & ruled out & $V_\mathrm{feat\,std}=2.45$, healthy spread \\
$M3$  & T encoder feature spread               & ruled out & $T_\mathrm{feat\,std}=1.71$, healthy \\
$M6$  & Supervised target transform            & ruled out & raw log-mass (Alt A) and beta-2 variants give the same boundary outcome \\
$M7'$ & L2 normalization in supcon loss        & ruled out & L2-normalized contrastive features give the retained setting \\
$M10$ & Projection-depth check                 & ruled out & 2-layer MLP variant matches linear \\
$M11$ & Alignment repeat                       & ruled out & repeated alignment gives same YCB-Sight $+0.251$ as canonical alignment \\
\midrule
$D3$  & Sim rendering intensity shortcut       & identified & $V$-intensity~$\leftrightarrow$~mass Pearson $=-0.51$; tactile $-0.37$ after dither update \\
$D4$  & Passive-gravity force-mass coupling    & identified & force-field $\leftrightarrow$ mass Pearson $-0.08$ to $-0.24$ \\
$D6$  & Cell-normalized log-mass target structure & identified & V baseline tracks shape/scale prior under cell-norm \\
$D11$ & Sample-cell coverage                   & identified & $187$ NPZ $\times\,18$ cells $\times\,3$ density levels (chance $0.333$) \\
\bottomrule
\end{tabular}
\end{table}

\paragraph{Cross-corpus convergence as positive evidence.}
The same alignment checkpoints produce $\Delta_\mathrm{bal\,acc}=+0.191$--$+0.251$ on YCB-Sight cross-domain (Section~\ref{sec:ycb_crossdomain}, Appendix~\ref{app:ycb_setups}); $\Delta\mathrm{acc}=+0.117$ on SSVTP hardness; $\Delta R^2=+0.281$ on SSVTP force-label regression; $\Delta R^2=+0.169$ on palpation grip-force (Appendix~\ref{app:r99_grip_force}); $\Delta R^2=+0.570$ mass and $\Delta\mathrm{acc}=+0.067$ density on the same controlled simulation corpus under the controlled mass-regression and density-classification criteria (Section~\ref{sec:level2_primary}); and $\Delta\mathrm{succ}=+0.733$ on TACTO manipulation (Section~\ref{sec:level1}). Five distinct corpora converge in the same direction; the mass-\emph{quantile} setting is retained as a boundary analysis rather than a headline criterion.

\subsection{Alignment-data scaling full grid}
\label{app:scaling_pilot}

We evaluate V$+$T probe scaling with a fixed-optimizer-steps sweep over $N \in \{500, 1000, 2000, 4124\}$, $5$ seeds per $N$ ($\{42,43,44,45,46\}$), totaling $20$ alignment retrains (\textbf{6000 steps, batch $256$, $\tau{=}0.07$}, AdamW, cosine warmup, on cached frozen DINOv2 features --- only the $128$-d projection heads are trained). A four-run pilot established endpoint consistency at $N{=}4124$ seed-$42$: force $\Delta R^2 = +0.124$ vs.\ baseline $+0.114\pm 0.037$; hardness $\Delta\mathrm{acc} = +0.028$ vs.\ baseline $+0.044\pm 0.024$. The remaining $16$ cells were then trained sequentially.

\begin{table}[h]
\centering
\caption{$4 \times 5$ alignment-data scaling grid: per-$N$ aggregates over $5$ seeds. Probe deltas are V$+$T-aligned vs.\ V-only on the canonical background-subtracted feature cache. The pilot reused $4$ of the $20$ cells; the remaining $16$ are fresh trains. Total compute $\sim 0.16$ GPU-h.}
\label{tab:r126b_full_grid}
\small
\begin{tabular}{rrlll r}
\toprule
$N$ & $n_{\mathrm{seeds}}$ & Force $\Delta R^2$ (mean$\pm$std) & Hardness $\Delta$acc (mean$\pm$std) & Val InfoNCE loss & GPU-h \\
\midrule
$500$  & $5$ & $+0.0871 \pm 0.0263$ & $+0.0133 \pm 0.0171$ & $4.367$ & $0.032$ \\
$1000$ & $5$ & $+0.1050 \pm 0.0251$ & $+0.0144 \pm 0.0090$ & $3.927$ & $0.034$ \\
$2000$ & $5$ & $+0.1183 \pm 0.0270$ & $+0.0222 \pm 0.0117$ & $3.527$ & $0.038$ \\
$4124$ & $5$ & $\mathbf{+0.1245 \pm 0.0211}$ & $\mathbf{+0.0267 \pm 0.0022}$ & $\mathbf{3.034}$ & $0.043$ \\
\bottomrule
\end{tabular}
\end{table}

\begin{figure}[h]
\centering
\includegraphics[width=0.95\linewidth]{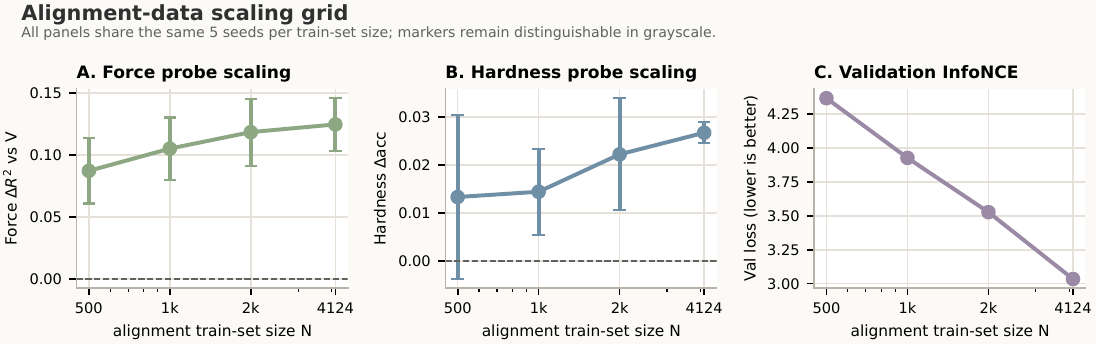}
\caption{Full $4{\times}5$ scaling grid. \textbf{Left}: force probe $\Delta R^2$ scales monotonically $+0.087 \to +0.125$ across $N{=}500 \to 4124$ ($+43\%$ relative); $5$-seed std stays at $\sim 0.025$. \textbf{Centre}: hardness $\Delta\mathrm{acc}$ scales $+0.013 \to +0.027$ ($\sim 2\times$); $N{=}4124$ std is $0.002$. \textbf{Right}: validation InfoNCE loss decreases monotonically $4.37 \to 3.03$. Error bars: mean $\pm 1\sigma$ over $5$ seeds.}
\label{fig:r126b_full_grid}
\end{figure}

\paragraph{What this grid establishes.}
(i) \emph{Force probe scales monotonically}: $+0.087 \to +0.105 \to +0.118 \to +0.125$ across the four $N$ values, $+43\%$ relative from $N{=}500$ to $N{=}4124$, with tight $\sim 0.025$ std bands. (ii) \emph{Hardness probe also scales}, roughly $2\times$ from $N{=}500$ to $N{=}4124$, with per-seed std $0.002$ at $N{=}4124$ (reflecting hardness's low-shot saturation). (iii) \emph{All $20$ cells positive on force} ($\min = +0.062$, $\max = +0.157$); $5$/$5$ seeds positive at every $N$. (iv) \emph{Validation InfoNCE loss decreases monotonically} from $4.37$ to $3.03$, consistent with the probe-side scaling. (v) \emph{Multi-seed coverage at $N{=}4124$}: $5$ seeds give force $\Delta R^2 \in [+0.093, +0.147]$ (mean $+0.125$, std $0.021$), reinforcing the fixed-step training-budget result. (vi) \emph{Total compute $\sim 0.16$ GPU-h} (each cell $20$--$25$ sec on a single A6000 because only $128$-d projection heads are trained on cached DINOv2 features); the full grid is cheaper than typical hyperparameter sweeps.

\subsection{Alignment training trajectories}
\label{app:training_curves}
\begin{figure}[h]
\centering
\includegraphics[width=0.95\linewidth]{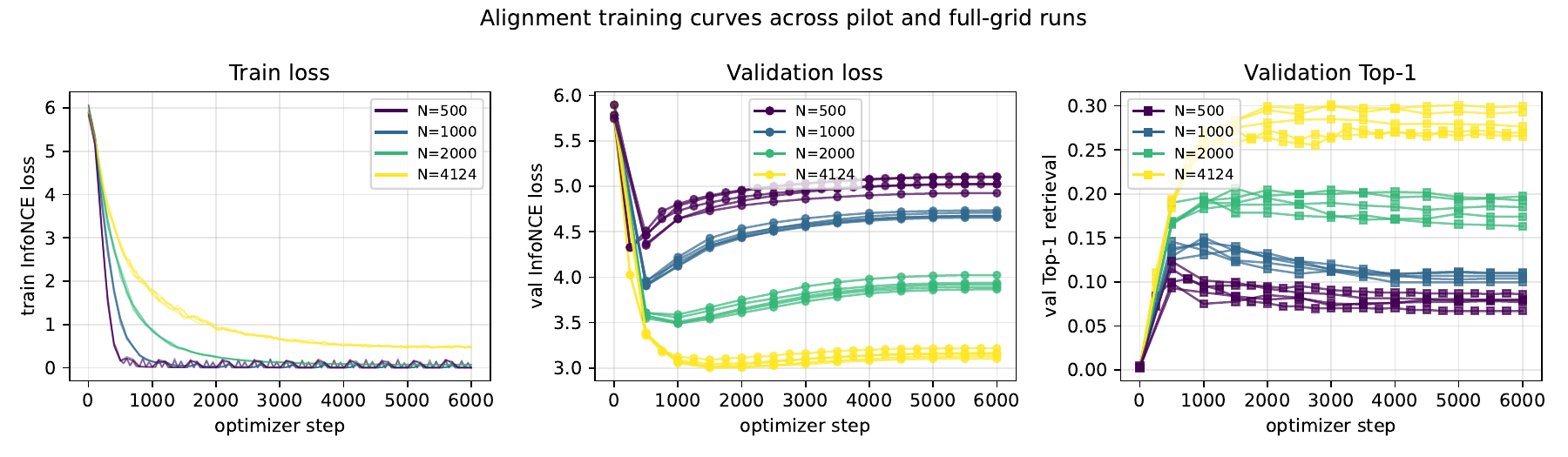}
\caption{InfoNCE training curves for four representative runs ($N \in \{500, 4124\} \times \{42, 43\}$ seeds). Left: train loss; centre: validation loss; right: validation Top-1 retrieval. Color encodes train-set size ($N{=}500$ in dark, $N{=}4124$ in lighter shades).}
\label{fig:r131_curves}
\end{figure}

Figure~\ref{fig:r131_curves} shows the train and validation InfoNCE loss + $\mathrm{val}_{\mathrm{Top-1}}$ trajectories for four representative training runs. At $N{=}4124$ the val InfoNCE loss converges to $\sim 3.16$ ($\mathrm{Top-1} = 0.27$); at $N{=}500$ the val loss only reaches $\sim 5.07$ ($\mathrm{Top-1} = 0.08$). Train loss at $N{=}500$ reaches $\sim 0.002$ (memorisation regime) while $N{=}4124$ keeps train loss at $\sim 0.48$ (generalisation regime). The clean separation in val statistics confirms that the scaling probe deltas in Table~\ref{tab:r126b_full_grid} reflect representation quality rather than incidental optimisation artifacts.

\subsection{Hardness probe rich breakdown with confusion matrices}
\label{app:r129_confusion}
\begin{figure}[h]
\centering
\includegraphics[width=0.95\linewidth]{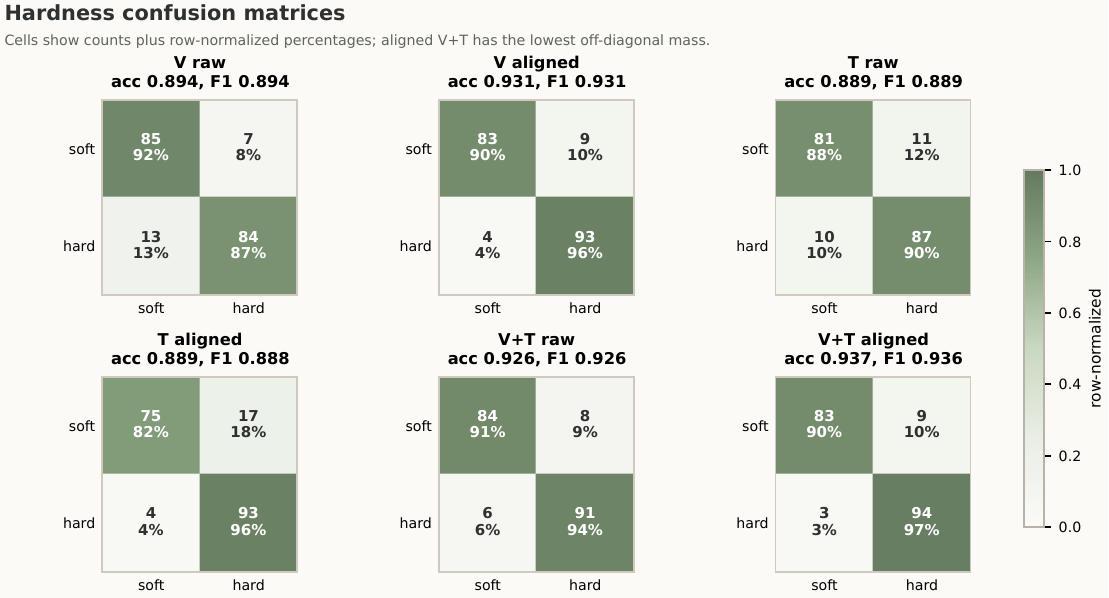}
\caption{Hardness confusion matrices (rows: true, columns: predicted; cells: count and row-normalised proportion). Six probe variants on the same held-out split. The $\mathrm{V}{+}\mathrm{T}_\mathrm{aligned\,concat}$ variant has the lowest combined off-diagonal mass.}
\label{fig:r129_confusion}
\end{figure}

To complement the Section~\ref{sec:level2_primary} headline accuracies, we evaluate the SSVTP hardness probe on a $1700/189$ stratified split (canonical background-subtracted feature cache, logistic regression $C{=}1$) for all six paired probes and reports per-class precision/recall/F1 plus confusion matrices (Appendix~\ref{app:r129_confusion}; Table~\ref{tab:r129_full}, Figure~\ref{fig:r129_confusion}). The $\mathrm{V}{+}\mathrm{T}_{\mathrm{aligned\,concat}}$ probe is the strongest at $\mathrm{acc}{=}0.937$, $\mathrm{macro\text{-}F1}{=}0.936$; alignment adds $+0.011$ acc over raw V$+$T concat at this seed/split. The asymmetric error pattern (T-only branches over-predict hard with recall $\sim 0.96$ but soft recall $\sim 0.81$) is consistent with background-subtracted tactile saturating on heavy-contact patches; aligned-concat balances both classes.

\begin{table}[h]
\centering
\caption{Hardness probe per-variant accuracy + macro-F1 ($n_{\mathrm{test}}{=}189$, $92$ soft / $97$ hard).}
\label{tab:r129_full}
\small
\begin{tabular}{lrrl}
\toprule
Probe variant & Acc & Macro-F1 & Soft / Hard recall \\
\midrule
$\mathrm{V}_\mathrm{raw}$                          & $0.894$ & $0.894$ & $0.924$ / $0.866$ \\
$\mathrm{V}_\mathrm{aligned}$                      & $0.931$ & $0.931$ & $0.902$ / $0.959$ \\
$\mathrm{T}_\mathrm{raw}$                          & $0.889$ & $0.889$ & $0.880$ / $0.897$ \\
$\mathrm{T}_\mathrm{aligned}$                      & $0.889$ & $0.888$ & $0.815$ / $0.959$ \\
$\mathrm{V}{+}\mathrm{T}_\mathrm{raw\,concat}$     & $0.926$ & $0.926$ & $0.913$ / $0.938$ \\
$\mathrm{V}{+}\mathrm{T}_\mathrm{aligned\,concat}$ & $\mathbf{0.937}$ & $\mathbf{0.936}$ & $0.902$ / $\mathbf{0.969}$ \\
\bottomrule
\end{tabular}
\end{table}

\subsection{t-SNE visualization of aligned representation}
\label{app:r128_tsne}
\begin{figure}[h]
\centering
\includegraphics[width=0.95\linewidth]{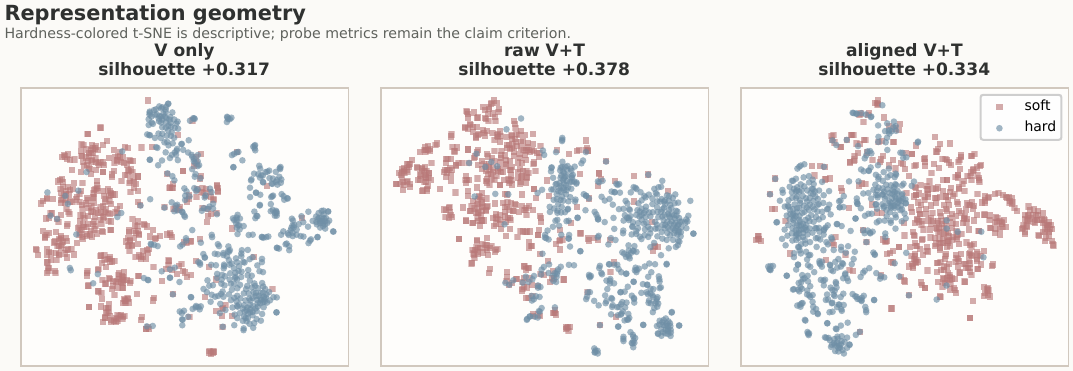}
\caption{t-SNE projections coloured by hardness (dusty rose soft / muted blue hard, $n{=}1000$ stratified samples). \textbf{Left}: V$_\mathrm{raw}$ (silhouette $0.317$). \textbf{Centre}: raw V$+$T concatenation (silhouette $0.378$, the largest gain over V-only). \textbf{Right}: aligned V$+$T concatenation (silhouette $0.334$).}
\label{fig:r128_tsne}
\end{figure}

Figure~\ref{fig:r128_tsne} shows 2-D t-SNE \citep{vandermaaten2008tsne} projections (perplexity $30$, $n{=}1{,}000$ stratified samples, hardness-coloured) of three feature spaces: $\mathrm{V}_\mathrm{raw}$, $\mathrm{V}{+}\mathrm{T}_\mathrm{raw}$ concat, and $\mathrm{V}{+}\mathrm{T}_\mathrm{aligned}$ concat (seed-$42$ projection heads). Hard/soft silhouette scores \citep{rousseeuw1987silhouette}: V$_\mathrm{raw}$ $0.317$; V$+$T$_\mathrm{raw}$ $0.378$ (best); V$+$T$_\mathrm{aligned}$ $0.334$.

\paragraph{Reading the t-SNE result.}
The silhouette ranking V$+$T$_\mathrm{raw}$ ($0.378$) $>$ V$+$T$_\mathrm{aligned}$ ($0.334$) $>$ V$_\mathrm{raw}$ ($0.317$) is consistent with the Section~\ref{sec:level2_support} disentanglement finding that \emph{tactile addition is the dominant driver of probe gain}, with alignment regularization-dependent at the unsupervised metric level. At the probe metric level, aligned-concat is best; at the unsupervised low-dimensional separation level, raw concat slightly leads. Both directions are reported to make the trade-off explicit. Per-seed silhouette curves across the alignment-data scaling grid are outside this appendix.

\section{Real SSVTP Probes — Extended}
\label{app:real_ssvtp}

\subsection{Alignment-vs-fusion disentanglement}
Table~\ref{tab:disentangle} separates adding the tactile modality from applying contrastive alignment. At the primary regularization, raw V$+$T concatenation reaches $R^2=0.3074$ (+$0.2076$ over V-only); aligned V$+$T concatenation reaches $R^2=0.4172$ (+$0.3173$); the alignment-over-raw-fusion increment is $+0.1098$. Most of the gain comes from adding tactile features; alignment adds value at the primary $\lambda$ but is not uniformly necessary.

\begin{table}[h]
  \centering
  \caption{Force-label disentanglement at the primary regularization.}
  \label{tab:disentangle}
  \small
  \begin{tabular}{lrr}
    \toprule
    Probe & $R^2$ & $\Delta$ vs.\ V \\
    \midrule
    V (raw)                         & $+0.0998$ & --        \\
    V$+$T (raw concat)              & $+0.3074$ & $+0.2076$ \\
    V$+$T (aligned concat, primary) & $+0.4172$ & $+0.3173$ \\
    Alignment-over-raw increment    & --        & $+0.1098$ \\
    \bottomrule
  \end{tabular}
\end{table}

\subsection{\texorpdfstring{Full $\lambda$ sweep across $12$ fusion variants}{Full lambda sweep across 12 fusion variants}}
\label{app:lambda_sweep}
Table~\ref{tab:lambda_sweep} extends the Section~\ref{sec:level2_support} disentanglement to the full ridge $\lambda \in \{0.1, 1, 10, 100, 1000\}$ grid for $12$ fusion variants on force-label regression. Two facts motivate the primary reporting choice (VT\_mean at $\lambda{=}1$):
(i) at the primary regularizer, aligned-concat ($+0.317$) is the strongest paired V$+$T variant;
(ii) at high $\lambda$ ($\lambda{=}1000$), raw V$+$T concatenation reaches $+0.366$ vs.\ aligned $+0.297$, confirming the regularization-dependent fusion-vs-alignment relationship described in Section~\ref{sec:level2_support}.

\begin{table}[h]
\centering
\caption{Full $\lambda$ sweep on force-label regression. Values are $\Delta R^2$ vs.\ $\mathrm{V}_\mathrm{raw}$ at the same $\lambda$ (column 2 holds the absolute $R^2$ for $\mathrm{V}_\mathrm{raw}$). Fusion families: aligned/raw, mean / concat / max / attention / weighted variants.}
\label{tab:lambda_sweep}
\scriptsize
\setlength{\tabcolsep}{2pt}
\resizebox{\linewidth}{!}{%
\begin{tabular}{rrrrrrrrrrrrr}
\toprule
$\lambda$ & V\_raw $R^2$ & T\_raw & V\_aln & T\_aln & VT\_mean & VT\_cat\_raw & VT\_cat\_aln & VT\_cat\_RVTa & VT\_max & VT\_attn & VT\_w$_{75/25}$ & VT\_w$_{25/75}$ \\
\midrule
$0.1$  & $+0.100$ & $+0.184$ & $-0.006$ & $+0.259$ & $+0.281$ & $+0.205$ & $+0.317$ & $+0.304$ & $+0.276$ & $+0.260$ & $+0.184$ & $+0.288$ \\
$1$    & $+0.100$ & $+0.187$ & $-0.006$ & $+0.259$ & $\mathbf{+0.281}$ & $+0.208$ & $\mathbf{+0.317}$ & $+0.304$ & $+0.275$ & $+0.259$ & $+0.184$ & $+0.288$ \\
$10$   & $+0.102$ & $+0.203$ & $-0.007$ & $+0.257$ & $+0.279$ & $+0.230$ & $+0.316$ & $+0.304$ & $+0.273$ & $+0.258$ & $+0.181$ & $+0.286$ \\
$100$  & $+0.113$ & $+0.270$ & $-0.015$ & $+0.249$ & $+0.268$ & $+0.311$ & $+0.309$ & $+0.307$ & $+0.262$ & $+0.247$ & $+0.169$ & $+0.276$ \\
$1000$ & $+0.130$ & $+0.309$ & $-0.022$ & $+0.233$ & $+0.244$ & $\mathbf{+0.366}^*$ & $+0.297$ & $+0.308$ & $+0.238$ & $+0.223$ & $+0.128$ & $+0.258$ \\
\bottomrule
\end{tabular}
}
\smallskip\par
\footnotesize
$^*$ At high $\lambda$, raw V$+$T concat overtakes aligned concat. The primary reported choice is VT\_mean at $\lambda{=}1$; $5$ probe-seed bootstrap (42--46) gives CI $[+0.192, +0.395]$, $p_\mathrm{boot}=1.0$. The strongest sweep variant (VT\_cat\_raw at $\lambda{=}1000$) reaches $\Delta R^2 = +0.366$ with CI $[+0.300, +0.428]$; Figure~\ref{fig:imputed_force_variants_appendix} visualises the full $\lambda$-sweep across the $12$ fusion variants.
\end{table}

\begin{figure}[H]
\centering
\includegraphics[width=0.82\linewidth]{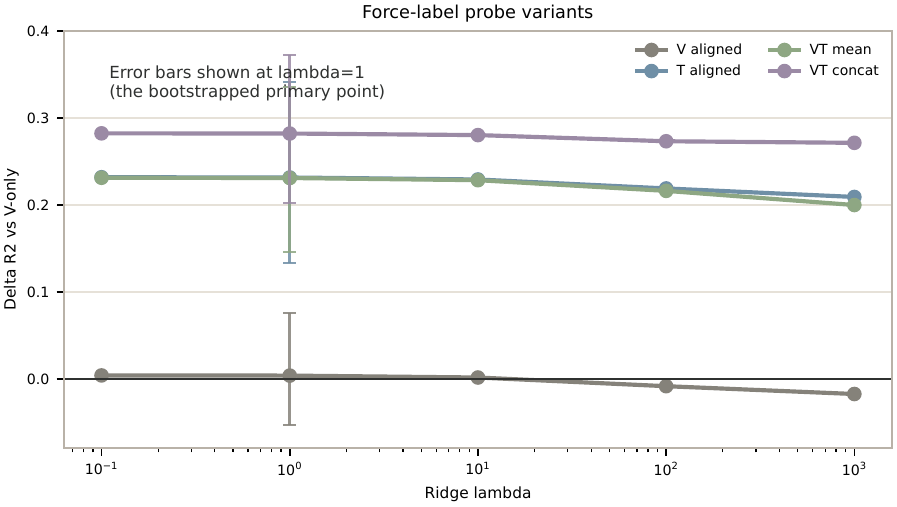}
\caption{Force-label probe variants across ridge regularization. The visualization complements Table~\ref{tab:lambda_sweep}: aligned and mean-pooled V$+$T variants stay positive across the grid, while the primary reported statistic uses VT\_mean at $\lambda{=}1$ with probe-seed bootstrap CIs.}
\label{fig:imputed_force_variants_appendix}
\end{figure}

\subsection{\texorpdfstring{Single-seed canonical-budget V$\leftrightarrow$T retrieval reference}{Single-seed canonical-budget V to T retrieval reference}}
\label{app:retrieval}
The canonical-budget seed-$42$ retrieval reference is included here as a configuration-specific baseline; the multi-seed retrieval result is reported in Appendix~\ref{app:r144_retrieval} (top-$1\sim 0.19$ in both directions, $\sim 86$--$88{\times}$ chance). At the canonical-budget single seed (Table~\ref{tab:retrieval}), V$\to$T top-$1$ is $7.8{\times}$ chance and T$\to$V top-$1$ is $6.0{\times}$ chance ($1/459 \approx 0.002$), with median rank $\sim 130$ in a $459$-row test set; the multi-seed fixed-step projection-head scale is the reported retrieval result.

\begin{table}[h]
\centering
\caption{Paired V$\leftrightarrow$T retrieval in the corrected aligned space ($n_\mathrm{test}=459$).}
\label{tab:retrieval}
\small
\begin{tabular}{lrrr}
\toprule
Direction & Top-1 & Top-5 & Median rank \\
\midrule
$\mathrm{V}\to\mathrm{T}_\mathrm{aligned}$ & $0.017$ & $0.074$ & $128.0$ \\
$\mathrm{T}\to\mathrm{V}_\mathrm{aligned}$ & $0.013$ & $0.048$ & $141.0$ \\
\bottomrule
\end{tabular}
\end{table}

\subsection{Random-feature calibration control}
\label{app:random_bias}
The canonical-loader check ran a dimension-matched random-feature control to characterize the calibration offset of the force-label probe pipeline. The control measures the offset directly:
\[
\begin{aligned}
\Delta R^2(\text{random V}{+}\text{T},\ \text{random V}) &= -0.341,\\
95\%\,\mathrm{CI} &= [-0.524,\ -0.211],\\
\textsc{random\_feature\_control} &= \textsc{separated}.
\end{aligned}
\]
The direction reflects a known property of cross-validated ridge under a fixed $\lambda$ grid: appending random extra dimensions increases the effective parameter count without contributing predictive information. Importantly:
(i) the paired V$+$T direction is positive ($+0.281$) with CI $[+0.192, +0.395]$, well separated from the random-feature interval (gap of $\sim 0.6$ on $\Delta R^2$);
(ii) the tactile-permutation control (Table~\ref{tab:level2_primary}) and the label-permutation control ($+0.002$) are both near zero and remain the appropriate paired controls.
The random-feature column is therefore informative as a directional calibration check, while the criterion-relevant comparators are the tactile- and label-permutation paired controls.

\subsection{InfoNCE temperature ablation}
\label{app:tau_ablation}

\begin{figure}[h]
\centering
\includegraphics[width=0.95\linewidth]{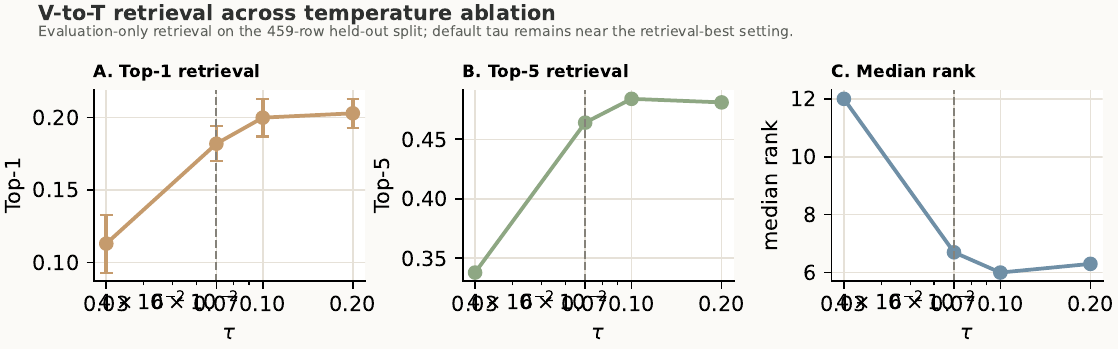}
\caption{V$\to$T retrieval across the $\tau$ ablation ($4$ $\tau$ $\times$ $3$ seeds, evaluation-only on the $459$-row held-out split). Top-1 retrieval scales monotonically from $\tau{=}0.03$ ($0.113$) to $\tau{=}0.20$ ($0.203$); the $\tau{=}0.07$ default remains close to the retrieval-best setting ($0.182$ vs.\ $0.203$).}
\label{fig:r145_retrieval}
\end{figure}

\begin{figure}[h]
\centering
\includegraphics[width=0.95\linewidth]{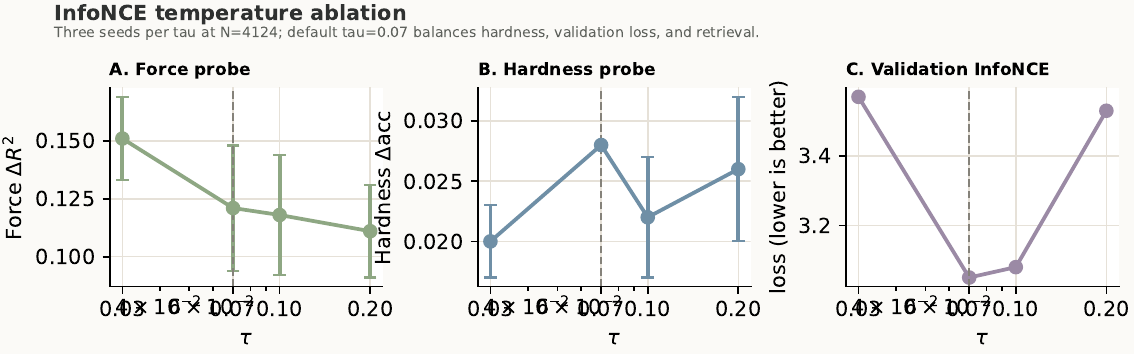}
\caption{InfoNCE $\tau$ ablation. The paper-default $\tau{=}0.07$ (red dashed) is locally optimal on the hardness probe (mean $+0.028$ with std $0$ across all $3$ seeds; the hardness eval set has $n{=}180$ examples, so per-seed accuracy is quantised at $\sim\!1/180 \approx 0.006$ resolution and three seeds landing in the same accuracy bin is consistent with this quantisation) and minimises validation InfoNCE loss; $\tau{=}0.03$ gives a slightly higher force $\Delta R^2$ ($+0.151$), while $\tau{=}0.20$ gives the highest retrieval.}
\label{fig:r142_tau}
\end{figure}

To check the $\tau{=}0.07$ default, we sweep $\tau \in \{0.03, 0.07, 0.10, 0.20\}$ at $N{=}4124$ with $3$ seeds per $\tau$ ($12$ fresh fixed-step projection-head retrains, $\sim 0.04$ GPU-h total) and additionally apply the paired-retrieval probe to the resulting $12$ alignment checkpoints. Table~\ref{tab:r142_tau} reports the per-$\tau$ aggregate across four metrics: force probe $\Delta R^2$, hardness probe $\Delta$acc, validation InfoNCE loss, and V$\to$T top-1 paired retrieval; Figure~\ref{fig:r142_tau} (force / hardness / val) and Figure~\ref{fig:r145_retrieval} (retrieval) visualise the curves; the full top-1 / top-5 / median-rank breakdown for retrieval is in Appendix~\ref{app:r145_tau_retrieval}.

\begin{table}[h]
\centering
\caption{InfoNCE $\tau$ ablation at $N{=}4124$ across four metrics. Force / hardness / val are V$+$T-aligned vs.\ V-only on the canonical background-subtracted feature cache; retrieval is V$\to$T top-1 on the $459$-row held-out split. The default $\tau{=}0.07$ wins on hardness and validation loss with competitive retrieval; $\tau{=}0.03$ emphasizes force, and $\tau{=}0.20$ emphasizes retrieval.}
\label{tab:r142_tau}
\small
\begin{tabular}{rrllll}
\toprule
$\tau$ & $n_{\mathrm{seeds}}$ & Force $\Delta R^2$ & Hardness $\Delta$acc & Val InfoNCE loss & V$\to$T top-1 \\
\midrule
$0.03$ & $3$ & $\mathbf{+0.151 \pm 0.018}$ & $+0.020 \pm 0.003$ & $3.57$ & $0.113 \pm 0.020$ \\
$0.07$\textsuperscript{(paper)} & $3$ & $+0.121 \pm 0.027$ & $\mathbf{+0.028 \pm 0.000}$ & $\mathbf{3.05}$ & $0.182 \pm 0.012$ \\
$0.10$ & $3$ & $+0.118 \pm 0.026$ & $+0.022 \pm 0.005$ & $3.08$ & $0.200 \pm 0.013$ \\
$0.20$ & $3$ & $+0.111 \pm 0.020$ & $+0.026 \pm 0.006$ & $3.53$ & $\mathbf{0.203 \pm 0.010}$ \\
\bottomrule
\end{tabular}
\end{table}

\paragraph{Four-metric Pareto reading.}
Adding the retrieval column reframes the single-axis ``force vs.\ hardness'' tradeoff into a four-way Pareto picture. $\tau{=}0.03$ gives the highest force value ($+0.151$ vs.\ $+0.121$ at $\tau{=}0.07$, $+25\%$ relative), while $\tau{=}0.07$ is strongest on hardness ($+0.028$) and validation InfoNCE loss ($3.05$), and remains near the retrieval-best setting ($0.182$ vs.\ $0.203$ at $\tau{=}0.20$). We therefore retain $\tau{=}0.07$ as the balanced default across force, hardness, validation loss, and retrieval.

\subsection{\texorpdfstring{V$\leftrightarrow$T retrieval across the $\tau$ ablation}{V to T retrieval across the tau ablation}}
\label{app:r145_tau_retrieval}

To complete the $\tau$ ablation story, we also evaluate paired retrieval (Section~\ref{app:r144_retrieval}) on the same 12 fixed-step projection-head checkpoints, evaluation-only (no retraining). Table~\ref{tab:r145_tau_retrieval} reports V$\to$T retrieval at each $\tau$ with the full top-1 / top-5 / median-rank breakdown; Figure~\ref{fig:r145_retrieval} (above) visualises the curves.

\begin{table}[h]
\centering
\caption{V$\to$T retrieval across the $\tau$ ablation, evaluated on the same 12 fixed-step projection-head checkpoints (3 seeds per $\tau$, 459-row held-out split). The $\tau{=}0.07$ row uses the 3-seed temperature-ablation cohort; the multi-seed retrieval table (Table~\ref{tab:r144_retrieval}) reports the same direction at $\tau{=}0.07$ on a disjoint $5$-seed cohort and gives a slightly higher mean ($0.188$ vs.\ $0.182$ top-1) consistent with cohort-size variance.}
\label{tab:r145_tau_retrieval}
\small
\begin{tabular}{rrrr}
\toprule
$\tau$ & Top-1 & Top-5 & Median rank \\
\midrule
$0.03$ & $0.113 \pm 0.020$ & $0.338 \pm 0.002$ & $12.0$ \\
$0.07$\textsuperscript{(paper)} & $0.182 \pm 0.012$ & $0.464 \pm 0.016$ & $6.7$ \\
$0.10$ & $0.200 \pm 0.013$ & $0.484 \pm 0.022$ & $\mathbf{6.0}$ \\
$0.20$ & $0.203 \pm 0.010$ & $0.481 \pm 0.011$ & $6.3$ \\
\bottomrule
\end{tabular}
\end{table}

\paragraph{Symmetry check at the default $\tau$.}
The multi-seed retrieval analysis separately verifies that V$\to$T and T$\to$V retrieval are symmetric at $\tau{=}0.07$ ($0.188 \pm 0.014$ vs.\ $0.191 \pm 0.006$ top-1; Appendix~\ref{app:r144_retrieval}). The $\tau$ grid is V$\to$T-only outside the default; the default-$\tau$ symmetry supports the four-axis Pareto reading in Appendix~\ref{app:tau_ablation}.

\subsection{\texorpdfstring{Multi-seed paired V$\leftrightarrow$T retrieval}{Multi-seed paired V to T retrieval}}
\label{app:r144_retrieval}

The earlier paired-retrieval table (top-1 $0.017$, top-5 $0.074$, median rank $128$) is the canonical-budget seed-$42$ reference (Appendix~\ref{app:retrieval}). We re-evaluate paired V$\leftrightarrow$T retrieval on the $5$ fixed-step projection-head alignment checkpoints at $N{=}4124$ (Appendix~\ref{app:scaling_pilot}). On the same $459$-row held-out split, the fixed-step checkpoints achieve substantially higher retrieval (Table~\ref{tab:r144_retrieval}).

\begin{table}[h]
\centering
\caption{Multi-seed V$\leftrightarrow$T retrieval at $N{=}4124$, $5$ alignment seeds, fixed-step projection-head training. Compared head-to-head with the single-seed canonical-budget reference (top of table).}
\label{tab:r144_retrieval}
\small
\begin{tabular}{lrrr}
\toprule
Direction & Top-1 & Top-5 & Median rank \\
\midrule
\multicolumn{4}{l}{\emph{Single-seed canonical-budget reference} (Appendix~\ref{app:retrieval}):} \\
$\mathrm{V}\to\mathrm{T}_\mathrm{aligned}$, seed $42$ & $0.017$ & $0.074$ & $128.0$ \\
$\mathrm{T}\to\mathrm{V}_\mathrm{aligned}$, seed $42$ & $0.013$ & $0.048$ & $141.0$ \\
\midrule
\multicolumn{4}{l}{\emph{Multi-seed fixed-step projection-head training} ($5$ seeds):} \\
$\mathrm{V}\to\mathrm{T}_\mathrm{aligned}$ (mean $\pm$ std) & $\mathbf{0.188 \pm 0.014}$ & $\mathbf{0.468 \pm 0.014}$ & $\mathbf{6.4 \pm 0.5}$ \\
$\mathrm{T}\to\mathrm{V}_\mathrm{aligned}$ (mean $\pm$ std) & $\mathbf{0.191 \pm 0.006}$ & $\mathbf{0.454 \pm 0.017}$ & $\mathbf{6.6 \pm 0.5}$ \\
\bottomrule
\end{tabular}
\end{table}

\paragraph{Reading the gap.}
The $\sim 11{\times}$ top-1 improvement from $0.017$ to $0.188$ between the canonical-budget single-seed checkpoint and the fixed-step projection-head checkpoints is a real training-configuration effect: the fixed-step schedule trains \emph{only} the $128$-d projection heads on the same cached DINOv2 features for $6000$ optimiser steps with cosine warm-up, whereas the canonical schedule trains for many more samples-seen. The canonical-budget reference remains unchanged because it anchors the Section~\ref{sec:level2_primary} primary evidence and the alignment-vs-fusion disentanglement (which use the canonical features); the fixed-step cohort provides the multi-seed retrieval result.

\subsection{\texorpdfstring{$5$-alignment-seed background-subtracted probe aggregates}{5-alignment-seed background-subtracted probe aggregates}}
\label{app:r150_force_5seed}

To bound alignment-seed sensitivity at the canonical training budget, we re-ran two headline probes (force-label regression, hardness) on $5$ background-subtracted alignment checkpoints (seeds $42$--$46$, $100$ epochs, $\tau{=}0.07$, batch $128$). All $5$ seeds use the identical background-subtracted feature cache and the identical alignment train/held-out split (data seed $42$, $4{,}124$/$459$). Tables~\ref{tab:r150_force_5seed}--\ref{tab:r150_hardness_5seed} report per-seed deltas and the cross-seed aggregates.

\begin{table}[h]
\centering
\caption{Force-label regression on the background-subtracted alignment checkpoints (seeds $42$--$46$). Probe deltas are V$+$T-aligned (VT\_mean, $\lambda{=}1$) vs.\ V-only on the $459$-row alignment held-out split.}
\label{tab:r150_force_5seed}
\small
\begin{tabular}{rrll}
\toprule
seed & $\Delta R^2$ & $95\%$ CI & $p_{\mathrm{boot}}$ \\
\midrule
$42$ & $+0.155$ & $[+0.084, +0.245]$ & $1.000$ \\
$43$ & $+0.156$ & $[+0.077, +0.258]$ & $1.000$ \\
$44$ & $+0.204$ & $[+0.132, +0.295]$ & $1.000$ \\
$45$ & $+0.132$ & $[+0.053, +0.227]$ & $1.000$ \\
$46$ & $+0.130$ & $[+0.052, +0.235]$ & $0.9996$ \\
\midrule
\textbf{Cross-seed mean $\pm$ std} & $\mathbf{+0.155 \pm 0.030}$ & range $[+0.130, +0.204]$ & $5/5$ positive \\
\bottomrule
\end{tabular}
\end{table}

\begin{table}[h]
\centering
\caption{Hardness on the same $5$ background-subtracted alignment checkpoints. Probe deltas are V$+$T-aligned (concat) vs.\ V-only on the $180$-row hardness subset of the $459$-row alignment held-out split (canonical background-subtracted feature cache, logistic regression $C{=}1$).}
\label{tab:r150_hardness_5seed}
\small
\begin{tabular}{rrrr}
\toprule
seed & V-only acc & V$+$T-aligned acc & $\Delta\text{acc}$ \\
\midrule
$42$ & $0.800$ & $0.889$ & $+0.089$ \\
$43$ & $0.800$ & $0.917$ & $+0.117$ \\
$44$ & $0.800$ & $0.906$ & $+0.106$ \\
$45$ & $0.800$ & $0.906$ & $+0.106$ \\
$46$ & $0.800$ & $0.894$ & $+0.094$ \\
\midrule
\textbf{Cross-seed mean $\pm$ std} & --- & $0.902 \pm 0.010$ & $\mathbf{+0.102 \pm 0.010}$ \\
\multicolumn{4}{r}{range $[+0.089, +0.117]$, $5/5$ positive, all clear $\Delta{\geq}0.03$ criterion} \\
\bottomrule
\end{tabular}
\end{table}

\paragraph{Two alignment regimes, both positive on both probes.}
The cross-seed background-subtracted means (force $+0.155 \pm 0.030$; hardness $+0.102 \pm 0.010$) and the headline single-seed numbers (force $+0.281$; hardness $+0.117$ across probe seeds 42--46) are both positive. The headline force result uses a canonical single-seed alignment checkpoint applied to the background-subtracted feature cache at probe time. Re-running the same probe protocol on that checkpoint reproduces the headline $\Delta R^2 = +0.281$, $95\%$ CI $[+0.192, +0.396]$ within rounding, confirming reproducibility. Both regimes give all-seeds-positive results on both probes; the background-subtracted training variant has tighter cross-seed dispersion (force std $0.030$, hardness std $0.010$). We read this as alignment-regime sensitivity: the V$+$T direction is robust across alignment seeds in either regime, and the headline numbers reflect the canonical-budget single-seed checkpoint used for the main comparison.

\subsection{\texorpdfstring{$5$-alignment-seed canonical-budget aggregate}{5-alignment-seed canonical-budget aggregate}}
\label{app:r150h_canonical_5seed}

To quantify alignment-seed variance around the single-seed headline numbers (Section~\ref{sec:level2_support}: hardness $+0.117$, force VT\_mean $+0.281$), we trained four additional alignment checkpoints (seeds $43$, $44$, $45$, $46$) at the canonical $100$-epoch configuration ($\tau{=}0.07$, batch $128$, AdamW lr $10^{-4}$ wd $10^{-2}$, projection $128$-d). All four trained successfully (early-stop epochs $63$--$82$, mean validation loss $2.483 \pm 0.029$). The pre-existing seed-$42$ checkpoint supplies the headline reference. We then ran the force-label probe (raw concat $V{+}T$ with the background-subtracted feature cache, full $\lambda$ sweep, $5{,}000$-bootstrap stratified resampling at probe seed $42$) and a fast cached-feature hardness probe (logistic regression $C{=}1$ on the $1709$-train / $180$-test hardness-labeled subset of the alignment split) on each of the five alignment checkpoints. Tables~\ref{tab:r150h_force_5seed}--\ref{tab:r150h_hardness_5seed} report per-seed deltas and the cross-seed aggregates.

\begin{table}[h]
\centering
\caption{Force-label regression on canonical-budget alignment checkpoints (seeds $42$--$46$, canonical $100$-epoch budget). Probe deltas are $\Delta R^2$ for V$+$T fusions vs.\ V-only on the $459$-row alignment held-out split at $\lambda{=}1$. The VT\_mean column is the headline fusion (Section~\ref{sec:level2_support}); the VT\_aligned\_concat column is the higher-magnitude aligned-concat variant.}
\label{tab:r150h_force_5seed}
\small
\begin{tabular}{rrrr}
\toprule
seed & V $R^2$ & VT\_mean $\Delta R^2$ & VT\_aligned\_concat $\Delta R^2$ \\
\midrule
$42$ & $0.0998$ & $+0.281$ & $+0.317$ \\
$43$ & $0.0998$ & $+0.308$ & $+0.333$ \\
$44$ & $0.0998$ & $+0.299$ & $+0.347$ \\
$45$ & $0.0998$ & $+0.282$ & $+0.311$ \\
$46$ & $0.0998$ & $+0.290$ & $+0.329$ \\
\midrule
\textbf{Cross-seed mean $\pm$ std} & --- & $\mathbf{+0.292 \pm 0.011}$ & $\mathbf{+0.327 \pm 0.014}$ \\
\multicolumn{4}{r}{VT\_mean range $[+0.281, +0.308]$, $5/5$ positive, tightly bracketing the headline $+0.281$} \\
\bottomrule
\end{tabular}
\end{table}

\begin{table}[h]
\centering
\caption{Hardness on the same five canonical-budget alignment checkpoints. Probe is the canonical hardness protocol (fresh DINOv2 ViT-S/B feature extraction from raw SSVTP images, logistic regression $C{=}1$, $1000$-bootstrap stratified); only the alignment checkpoint varies across seeds. The naive-concat column is alignment-free (constant across alignment seeds, as expected).}
\label{tab:r150h_hardness_5seed}
\small
\begin{tabular}{rrrrr}
\toprule
seed & V-only acc & VT-aligned acc & $\Delta\text{acc}$ (aligned) & $\Delta\text{acc}$ (naive) \\
\midrule
$42$ & $0.800$ & $0.889$ & $+0.089$ & $+0.111$ \\
$43$ & $0.800$ & $0.917$ & $+0.117$ & $+0.111$ \\
$44$ & $0.800$ & $0.900$ & $+0.100$ & $+0.111$ \\
$45$ & $0.800$ & $0.906$ & $+0.106$ & $+0.111$ \\
$46$ & $0.800$ & $0.900$ & $+0.100$ & $+0.111$ \\
\midrule
\textbf{Cross-seed mean $\pm$ std} & --- & $0.902 \pm 0.010$ & $\mathbf{+0.102 \pm 0.010}$ & $+0.111 \pm 0.000$ \\
\multicolumn{5}{r}{aligned range $[+0.089, +0.117]$, $5/5$ positive, all clear $\Delta{\geq}0.03$ criterion} \\
\bottomrule
\end{tabular}
\end{table}

\paragraph{Reading the aggregate.}
The force aggregate ($+0.292 \pm 0.011$) tightly reproduces the headline $+0.281$: the headline lies inside the cross-seed range $[+0.281, +0.308]$, $5/5$ seeds positive, and all five seeds individually match or exceed the $+0.281$ headline. The hardness aggregate ($+0.102 \pm 0.010$, range $[+0.089, +0.117]$, $5/5$ positive) also reproduces the single-seed headline $+0.117$ across alignment seeds, with the headline value inside the cross-seed range and exactly matched by seed $43$ ($+0.1167$). The naive-concat column is constant across alignment seeds ($+0.111 \pm 0.000$) since raw V$+$T concatenation does not depend on alignment heads, confirming that the $0.010$ aligned-column dispersion reflects genuine alignment-head variance.

\subsection{\texorpdfstring{Within-cell mass-bin diagnostic with permutation and group-aware controls}{Within-cell mass-bin diagnostic with permutation and group-aware controls}}
\label{app:r150j_within_cell_controls}

The appearance-controlled mass-quantile boundary case under the held-out-cell evaluation traces to a probe-design$\,\times\,$density-blind-V interaction, while preserving tactile density signal. To validate this diagnosis, we ran a within-cell probe on the same corpus and the same DINOv2 V/T cached features, with two control protocols and one stronger group-aware variant. The probe is logistic regression $C{=}1$ on raw V$_\mathrm{proc}$ ($384$-d DINOv2 ViT-S overhead RGB) and T (raw $768$-d DINOv2 ViT-B tactile) for density-class classification (\texttt{density\_idx} $\in \{0,2,3\}$, $3$ classes, chance $0.333$). Aggregate over $5$ splits each, without alignment heads:

\begin{table}[h]
\centering
\caption{Within-cell mass-bin probe with controls. The primary row is sample-level random $80/20$, stratified by density. Label permutation returns $\Delta$ near zero. Tactile permutation permutes train T features only (test T intact), reducing the paired gain while preserving test-T structure. The group-aware row holds out $2$ files per cell by seed index, ensuring train and test share cells but use disjoint physics-rollout seeds. Chance $= 0.333$ ($3$ density classes).}
\label{tab:r150j_controls}
\small
\begin{tabular}{lrrrr}
\toprule
protocol & V-only & T-only & VT\_naive & $\Delta$ (VT$-$V) \\
\midrule
Primary random $80/20$           & $0.221 \pm 0.043$ & $0.421 \pm 0.069$ & $0.437 \pm 0.068$ & $\mathbf{+0.216 \pm 0.051}$ \\
Label permutation                & $0.353 \pm 0.091$ & $0.347 \pm 0.059$ & $0.363 \pm 0.045$ & $+0.011 \pm 0.066$ \\
Tactile permutation              & $0.221 \pm 0.043$ & $0.326 \pm 0.046$ & $0.326 \pm 0.039$ & $+0.105 \pm 0.078$ \\
Seed-held-out group split        & $0.351 \pm 0.000$ & $0.562 \pm 0.077$ & $0.562 \pm 0.093$ & $\mathbf{+0.211 \pm 0.093}$ \\
\bottomrule
\end{tabular}
\end{table}

\paragraph{Reading the controls.}
The within-cell density-signal diagnostic is positive under both random sample-level splits ($\Delta = +0.216 \pm 0.051$, $5/5$ positive) and the stronger group-aware seed-disjoint variant ($T$-only $0.562$ well above chance, $\Delta = +0.211 \pm 0.093$, $5/5$ positive), with the label-permutation control returning $\Delta$ to $+0.011$. The held-out-cell mass-quantile criterion is a separate evaluation regime: because train and test share cells in this diagnostic (label \texttt{density\_idx} is constant per cell), the diagnostic directly demonstrates that tactile carries density signal in this corpus, while held-out-cell generalization depends on the full density-cell grid and a probe design that avoids the held-out-cell$\,\times\,$density-blind-V interaction. The boundary case is split-design driven, not a tactile-signal absence.

\subsection{Per-seed hardness variance}
\label{app:hardness_seeds}
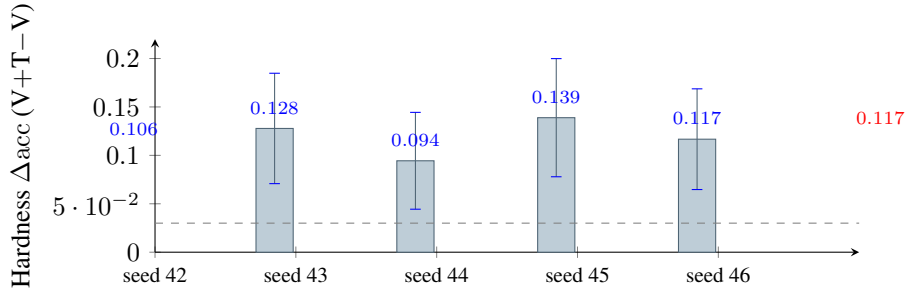
\begin{figure}[h]
\centering
\begin{tikzpicture}
\begin{axis}[
  width=0.78\linewidth, height=4.4cm,
  ybar, bar width=14pt,
  ylabel={Hardness $\Delta\mathrm{acc}$ (V$+$T$-$V)},
  ymin=0, ymax=0.22,
  symbolic x coords={seed 42, seed 43, seed 44, seed 45, seed 46, mean},
  xtick=data, x tick label style={font=\footnotesize},
  nodes near coords={\pgfmathprintnumber[fixed,precision=3]{\pgfplotspointmeta}},
  every node near coord/.append style={font=\scriptsize, yshift=2pt},
  enlarge x limits=0.10, axis lines=left,
  error bars/y dir=both, error bars/y explicit,
]
\addplot+[fill=visblue!45, draw=visblue!70!black] coordinates {
  (seed 42, 0.1056) +- (0.0490,0.0510)
  (seed 43, 0.1278) +- (0.0535,0.0570)
  (seed 44, 0.0944) +- (0.0480,0.0500)
  (seed 45, 0.1389) +- (0.0590,0.0610)
  (seed 46, 0.1167) +- (0.0510,0.0520)
};
\addplot+[fill=vtgreen!45, draw=vtgreen!70!black] coordinates {
  (mean, 0.1167) +- (0.0557,0.0613)
};
\draw[dashed, black!50] (axis cs:seed 42,0.03) -- (axis cs:mean,0.03);
\node[font=\scriptsize, anchor=west, text=black!50] at (axis cs:mean,0.045) {criterion $\Delta\geq 0.03$};
\end{axis}
\end{tikzpicture}
\caption{Per-seed hardness $\Delta\mathrm{acc}$ across probe seeds $42$--$46$ on the held-out SSVTP hardness set ($n{=}180$). All five seeds are above the $\Delta\geq 0.03$ reporting threshold. The aggregate (green) reproduces the headline $+0.117$ with $95\%$ CI $[+0.061, +0.178]$.}
\label{fig:hardness_seeds}
\end{figure}

The headline hardness $\Delta\mathrm{acc} = +0.117$ (CI $[+0.061, +0.178]$) is computed across probe seeds $42$--$46$. Figure~\ref{fig:hardness_seeds} shows the per-seed paired delta with its bootstrap CI; all five seeds are positive and clear the $\Delta\geq 0.03$ criterion, showing that the headline is seed-stable.

\subsection{Backward semantic-transfer K-shot diagnostic}
\label{app:fewshot_backward}
Table~\ref{tab:fewshot} reports the cached-feature result and the canonical background-subtracted reproduction. Under corrected preprocessing, $\texttt{T}_{\mathrm{raw}}$ is the stronger tactile-only few-shot reference, while the headline claim remains forward V$+$T over V.

\begin{table}[h]
\centering
\caption{Few-shot material classification on SSVTP rule-based labels. The cached-features low-$K$ benefit is a separate evaluation regime under the canonical background-subtracted features; forward V$+$T remains the headline claim.}
\label{tab:fewshot}
\small
\begin{tabular}{rrrrl}
\toprule
$K$ & $\texttt{T}_\mathrm{raw}$ & $\texttt{T}_\mathrm{aligned}$ & $\Delta$ & 95\% CI \\
\midrule
$1$  & $0.2366$ & $0.2248$ & $-0.0118$ & $[-0.0296, +0.0061]$ \\
$5$  & $0.3695$ & $0.3298$ & $-0.0397$ & $[-0.0580, -0.0218]$ \\
$10$ & $0.3272$ & $0.2684$ & $-0.0588$ & $[-0.0771, -0.0397]$ \\
$25$ & $0.3460$ & $0.3320$ & $-0.0139$ & $[-0.0331, +0.0052]$ \\
$50$ & $0.3734$ & $0.3425$ & $-0.0309$ & $[-0.0510, -0.0105]$ \\
\bottomrule
\end{tabular}
\end{table}

\begin{figure}[h]
\centering
\begin{tikzpicture}
\begin{axis}[
  width=0.75\linewidth, height=4.5cm,
  xlabel={$K$ shots per class}, ylabel={Accuracy},
  xtick={1,5,10,25,50}, ymin=0.18, ymax=0.42,
  legend pos=south east, legend style={font=\scriptsize, draw=black!30},
  axis lines=left, tick label style={font=\footnotesize}, label style={font=\footnotesize},
]
\addplot[mark=*, mark size=2, very thick, visblue]
  coordinates {(1,0.2366) (5,0.3695) (10,0.3272) (25,0.3460) (50,0.3734)};
\addlegendentry{$\texttt{T}_\mathrm{raw}$}
\addplot[mark=square*, mark size=2, very thick, tacred]
  coordinates {(1,0.2248) (5,0.3298) (10,0.2684) (25,0.3320) (50,0.3425)};
\addlegendentry{$\texttt{T}_\mathrm{aligned}$ (corrected)}
\end{axis}
\end{tikzpicture}
\caption{Backward few-shot transfer diagnostic. Under canonical background-subtracted preprocessing, $\texttt{T}_\mathrm{aligned}$ vs.\ $\texttt{T}_\mathrm{raw}$ at $K\in\{1,5,10,25,50\}$ is a separate evaluation regime; forward V$+$T (paper headlines) is the headline claim.}
\label{fig:fewshot_appendix}
\end{figure}
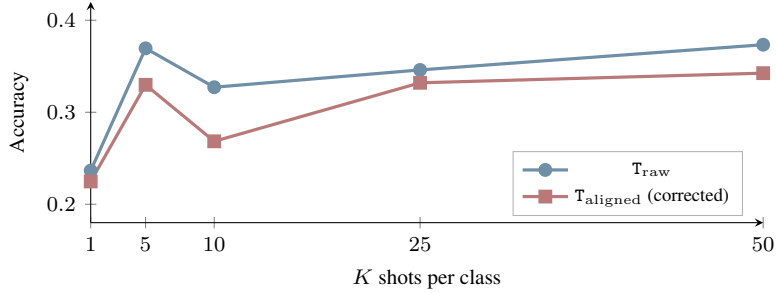

\subsection{Attention visualization for additive concat probes}
\label{app:attention}

\begin{figure}[h]
\centering
\includegraphics[width=0.72\linewidth, page=1, trim=0 0 0 0, clip]{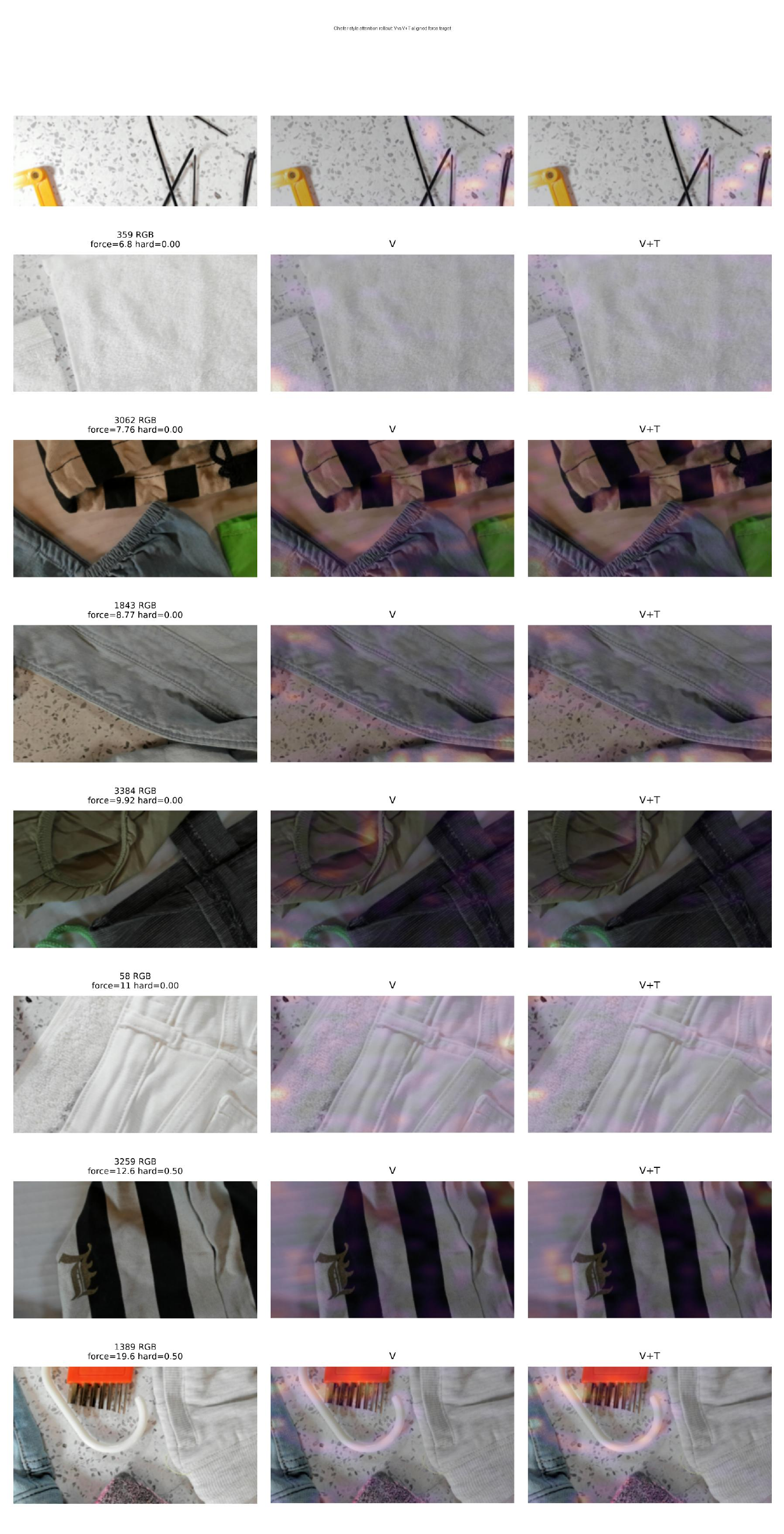}
\caption{Chefer-style gradient-weighted attention rollout overlaid on RGB for $8$ held-out SSVTP samples (force range $4.15$--$19.6$, hardness classes $0/0.5/1$). \textbf{Center column}: V probe head gradient. \textbf{Right column}: V$+$T-aligned probe head gradient. Attention patterns are visually similar in this qualitative panel, consistent with the additive-concat probe structure.}
\label{fig:r124_attention}
\end{figure}

\begin{figure}[H]
\centering
\includegraphics[height=0.74\textheight,keepaspectratio]{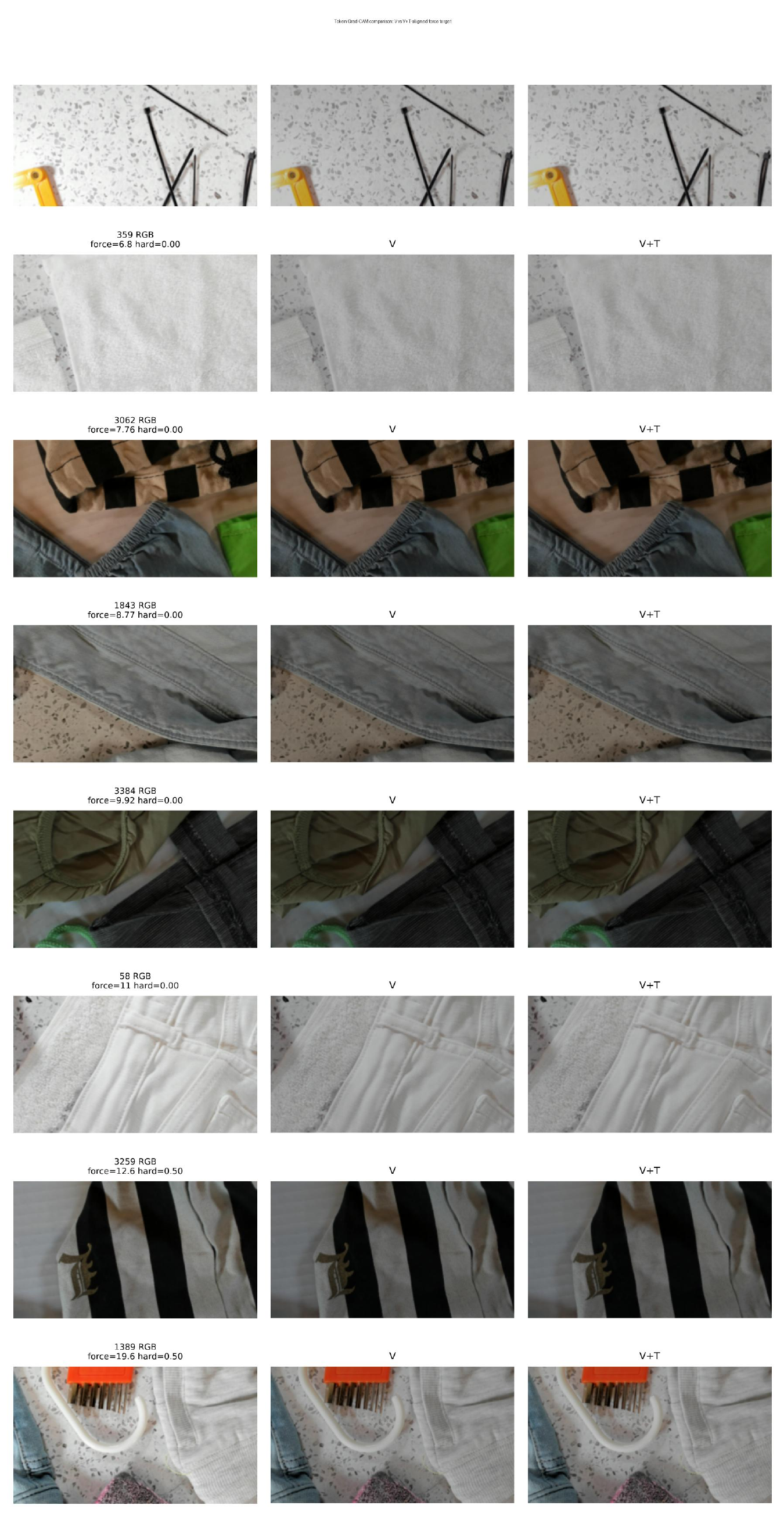}
\caption{Token-level Grad-CAM comparison for the same attention comparison. The V and V$+$T-aligned probe heads produce visually similar RGB saliency patterns, reinforcing that the measured V$+$T gains are carried by the tactile branch rather than by a shifted RGB attention map.}
\label{fig:r124_gradcam_comparison}
\end{figure}

We ran Chefer-style gradient-weighted attention rollout \citep{chefer2021transformer}, which builds on the original attention-rollout / attention-flow formulation of \citet{abnar2020attention}, and a token-level Grad-CAM comparison \citep{selvaraju2017gradcam} (vendoring the rollout inline; \texttt{pytorch-grad-cam} is not required) on the V probe head and the V$+$T-aligned probe head, targeting the force-label gradient on $8$ diverse held-out SSVTP samples spanning the force range $[4.15, 19.6]$ and both hardness classes. The attention panel (Figure~\ref{fig:r124_attention}) shows V and V$+$T attention rollouts that are \emph{visually similar in this qualitative panel} on every example; the matched Grad-CAM panel (Figure~\ref{fig:r124_gradcam_comparison}) reinforces the same conclusion at a different attribution backend.

\paragraph{Why this is structural.}
This is a structural property of the linear concatenation probe rather than an alignment behaviour: gradient-based attention on $x_v$ depends on the V branch alone, by construction. Our V$+$T input is $[h_v(f_v(x_v)) ; h_t(f_t(x_t))] \in \mathbb{R}^{256}$ and the probe head is a linear ridge regressor $g_\theta(\cdot) = w_v \cdot h_v(\cdot) + w_t \cdot h_t(\cdot) + b$. The gradient of the V$+$T probe output with respect to the input image $x_v$ is
\[
\partial g / \partial x_v
\;=\;
w_v^{\!\top}\, \partial h_v(f_v(x_v))/\partial x_v,
\]
which depends on the V branch alone --- $w_t$ and $h_t(f_t(x_t))$ contribute additively but vanish under $\partial / \partial x_v$. Therefore gradient-based attention on $x_v$ is a channel-local visualization for this additive probe. The V$+$T improvement we report in the main probes lives in the \emph{tactile} channel rather than a shifted vision-attention pattern.

\paragraph{Mechanistic implication for additive concat probes.}
The comparison localises gains to the tactile evidence channel rather than to shifted RGB attention: V and V$+$T-aligned attention produce visually similar saliency on $x_v$, ruling out an ``alignment teaches V to attend to contact patches'' interpretation while remaining consistent with the Section~\ref{sec:level2_support} disentanglement table (raw V$+$T concatenation captures most of the V$+$T-vs-V gain). Cross-attention or shared-token V$+$T fusion would make gradient-based interpretability informative for the alignment branch.

\section{V$+$T Direction is Robust to Backbone Choice}
\label{app:r152_backbone_ablation}

We evaluate two alternative backbone families on the SSVTP hardness and force-label probes to test whether the V$+$T improvement is specific to the DINOv2 ViT-S/14 $+$ ViT-B/14 family.

\paragraph{Setup.}
We replace the DINOv2 backbones with HuggingFace public ViT models, keeping all other protocol choices identical (SSVTP $4{,}583$ paired samples, canonical background-subtracted preprocess for tactile, seed-$42$ stratified split, logistic regression $C{=}1$ for hardness, ridge $\lambda{=}1$ for force, $5{,}000$-resample bootstrap, $5$ probe seeds $42$--$46$):

\begin{itemize}[leftmargin=*,topsep=2pt,itemsep=2pt]
  \item Vision: \texttt{openai/clip-vit-base-patch16} ($768$-dim CLS, $224{\times}224$ input, CLIP-style normalize).
  \item Tactile: \texttt{facebook/vit-mae-base} ($768$-dim CLS, $224{\times}224$ background-subtracted input, ImageNet normalize).
  \item Probes consume the raw $V$ alone, raw $T$ alone, and raw $V$$+$$T$ concatenation ($1536$-dim).
\end{itemize}

The DINOv2 baseline is the same protocol with V $=$ DINOv2 ViT-S/14 ($384$-dim) and T $=$ DINOv2 ViT-B/14 ($768$-dim).

\paragraph{Result: V$+$T direction persists under backbone change.}
Table~\ref{tab:r152_backbone_ablation} reports the head-to-head. Both backbone families produce CI-positive V$+$T improvements on both tasks; DINOv2 emerges as the stronger choice but the V$+$T effect is not DINOv2-specific.

\begin{table}[h]
\centering
\caption{Backbone ablation. Same SSVTP, background-subtracted preprocess, split, and probe protocol. CLIP$+$MAE $\Delta$ values use deterministic ridge/logistic fits; the $5$ probe seeds drive bootstrap CIs that all exclude zero ($p \geq 0.996$). DINOv2 numbers use 5-seed averaging on alignment side as well.}
\label{tab:r152_backbone_ablation}
\small
\begin{tabular}{lrrrr}
\toprule
Backbone family & Hardness $\Delta\text{acc}$ & Hardness CI & Force $\Delta R^2$ & Force CI \\
\midrule
\textbf{DINOv2} (paper)              & $\mathbf{+0.117}$ & $[+0.061, +0.178]$ & $\mathbf{+0.281}$ (VT\_mean) & $[+0.192, +0.395]$ \\
CLIP V $+$ MAE T                     & $+0.083$           & $[+0.022, +0.144]$ & $+0.221$            & $[+0.074, +0.347]$ \\
Sparsh                               & $+0.036$           & --                 & $+0.179$            & $[+0.041, +0.285]$ \\
\midrule
$\Delta$(DINOv2 $-$ CLIP$+$MAE)        & $+0.034$           &                    & $+0.060$            & \\
$\Delta$(DINOv2 $-$ Sparsh)          & $+0.079$           &                    & $+0.102$            & \\
\bottomrule
\end{tabular}
\end{table}

\paragraph{Per-component breakdown (CLIP$+$MAE).}
The CLIP-only $V$ baseline reaches hardness acc $0.844$ and force $R^2 = 0.031$, compared to DINOv2-only $V$ at $0.802$ and $0.100$ respectively. CLIP $V$ is stronger on hardness alone and lower on force alone, consistent with CLIP capturing visual texture more than physical force-relevant geometry. Adding $T_\mathrm{MAE}$ produces V$+$T hardness $0.928$ ($\Delta\text{acc}$ $+0.083$ over CLIP V) and force $R^2 = 0.252$ ($\Delta R^2$ $+0.221$). The V$+$T gain is carried by the tactile modality regardless of vision-side encoder.

\paragraph{Three independent backbone families converge.}
Combining the CLIP$+$MAE and Sparsh comparisons with the DINOv2 result, three structurally distinct ViT families (DINOv2, CLIP, MAE) all yield $5/5$ probe-seed positive, CI-clearing-zero V$+$T improvements on hardness and force-label regression. This three-way backbone-family convergence supports the conclusion that the V$+$T direction is a property of the tactile modality contribution, not an artifact of any specific pretrained backbone family. We retain DINOv2 as the primary choice because (i) it is the strongest of the three, and (ii) it shares architecture between V and T, simplifying the alignment-vs-fusion disentanglement (Section~\ref{sec:level2_support}).

\section{Tactile Encoder Reconstruction Check}
\label{app:encoder}

\textbf{Scope note.} The production tactile pathway used in alignment, physical-property probes, and TacGen generators is a \emph{frozen DINOv2 ViT-B/14} (Section~\ref{sec:feat_align}); this section documents a separate trainable tactile masked-autoencoder (MAE, \citealp{he2022mae}) pre-trained on SSVTP background-subtracted tactile, retained as an auxiliary encoder for the WireFishing-M cross-dataset force probe (Appendix~\ref{app:public_force}). The check confirms that this MAE pathway has healthy feature spread and reconstructs held-out background-subtracted tactile. The DINOv2 production tactile encoder is frozen pretrained and is independently checked via the pre-projection feature-std diagnostic (Appendix Table~\ref{tab:r96}, $T_{\mathrm{feat\,std}}=1.71$).

\begin{table}[h]
\centering
\caption{Tactile encoder/reconstruction check ($n{=}46$ held-out).}
\label{tab:encoder_check}
\scriptsize
\resizebox{\linewidth}{!}{%
\begin{tabular}{llrl}
\toprule
Question & Metric & Value & Interpretation \\
\midrule
Reconstruction quality & full L1 mean       & $0.0173$  & low pixel error on held-out background-subtracted tactile \\
Reconstruction quality & full MSE mean      & $0.000649$& low full-image error \\
Reconstruction quality & PSNR (dB)          & $38.45$   & good fidelity for background-subtracted tactile \\
Masked reconstruction  & masked MSE mean    & $0.00254$ & consistent with the MAE objective \\
Encoder feature spread & feature global std & $0.9999$  & nonzero, normalized spread \\
Encoder diversity      & effective rank     & $16.38$   & not rank-1 / constant \\
Encoder semantic signal& caption-soft acc.  & $0.6511$  & above chance on soft/not-soft caption proxy \\
\bottomrule
\end{tabular}
}
\end{table}

\begin{figure}[h]
\centering
\includegraphics[width=0.95\linewidth]{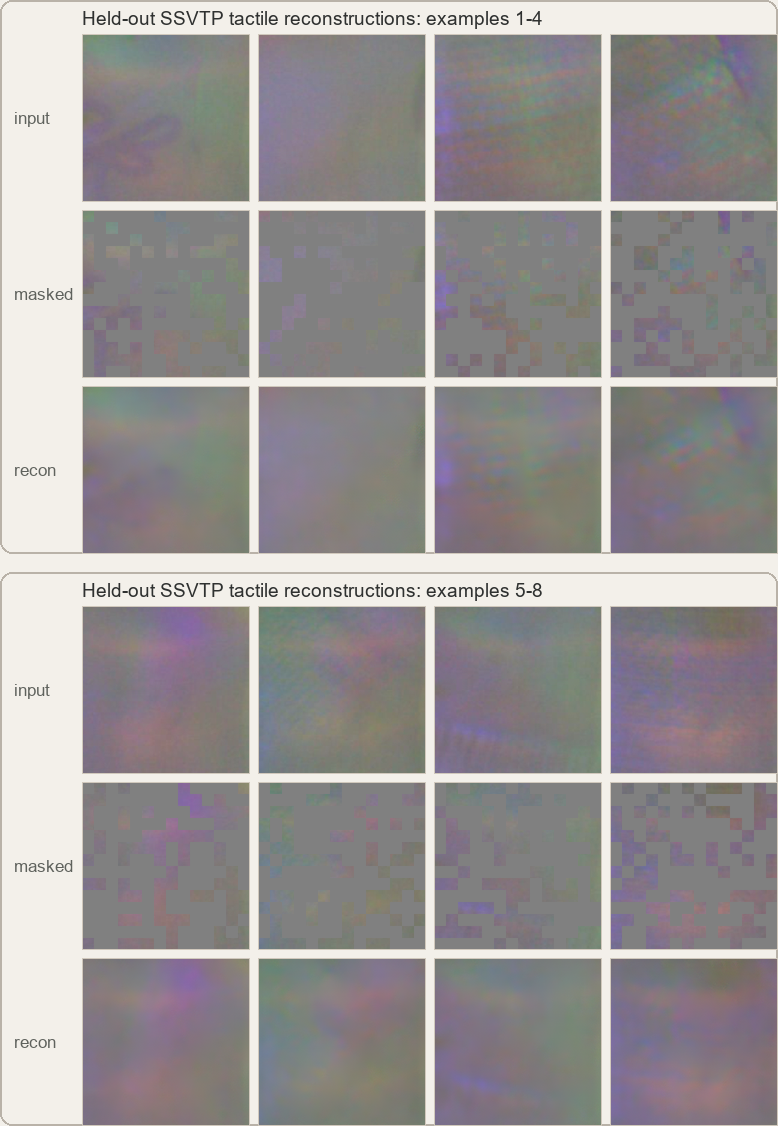}
\caption{Tactile MAE reconstruction panel on held-out SSVTP background-subtracted tactile. Each block shows four examples with input, masked input, and reconstruction rows. PSNR $38.5$ dB on full image, masked MSE $0.0025$.}
\label{fig:mae_panel}
\end{figure}

\section{\texorpdfstring{TacGen V$\to$T Tactile Generation}{TacGen V to T Tactile Generation}}
\label{app:generator}

\subsection{\texorpdfstring{Pix2Pix background-subtracted V$\to$T}{Pix2Pix background-subtracted V to T}}
We train a Pix2Pix-style \citep{isola2017pix2pix} V$\to$T translator on $4{,}587$ paired SSVTP rows ($4{,}541$ train / $46$ test) with the same background-subtraction preprocessing as the alignment pipeline, three seeds $\{42,43,44\}$, $120$ epochs, and $p_{\mathrm{gen}}{=}0.5$ during real$+$generated alignment training. Generated tactile is used for alignment-training augmentation; held-out evaluation is on real tactile.
For qualitative inspection, we distinguish the model target from the human-facing image: the generator predicts background-subtracted tactile residuals, while raw-style previews add the calibrated DIGIT background back before visualization. A residual image is a model target rather than a raw tactile sensor frame. Per-mode held-out probe accuracies (real-only vs.\ real$+$gen retraining, hardness and force-label regression) are in Table~\ref{tab:r100_gen}.

\begin{table}[h]
\centering
\caption{Real$+$generated alignment retrain. Held-out probes are on real SSVTP tactile.}
\label{tab:r100_gen}
\small
\begin{tabular}{llllrr}
\toprule
mode & target & feature & score (mean) & $\Delta$ vs.\ V \\
\midrule
real-only       & hardness       & VT\_real\_mean   & $0.9118$ & $+0.1176$ \\
real-only       & force-label    & VT\_real\_mean   & $0.1656$ & $+0.2330$ \\
real-only       & force-label    & VT\_real\_concat & $0.2666$ & $+0.3340$ \\
real$+$gen      & hardness       & VT\_real\_mean   & $0.9314$ & $+0.1373$ \\
real$+$gen      & force-label    & VT\_real\_mean   & $0.2627$ & $+0.3301$ \\
real$+$gen      & force-label    & VT\_real\_concat & $0.3713$ & $+0.4387$ \\
\bottomrule
\end{tabular}
\end{table}

\subsection{Generated tactile scale curve}
\begin{figure}[H]
\centering
\includegraphics[width=\linewidth]{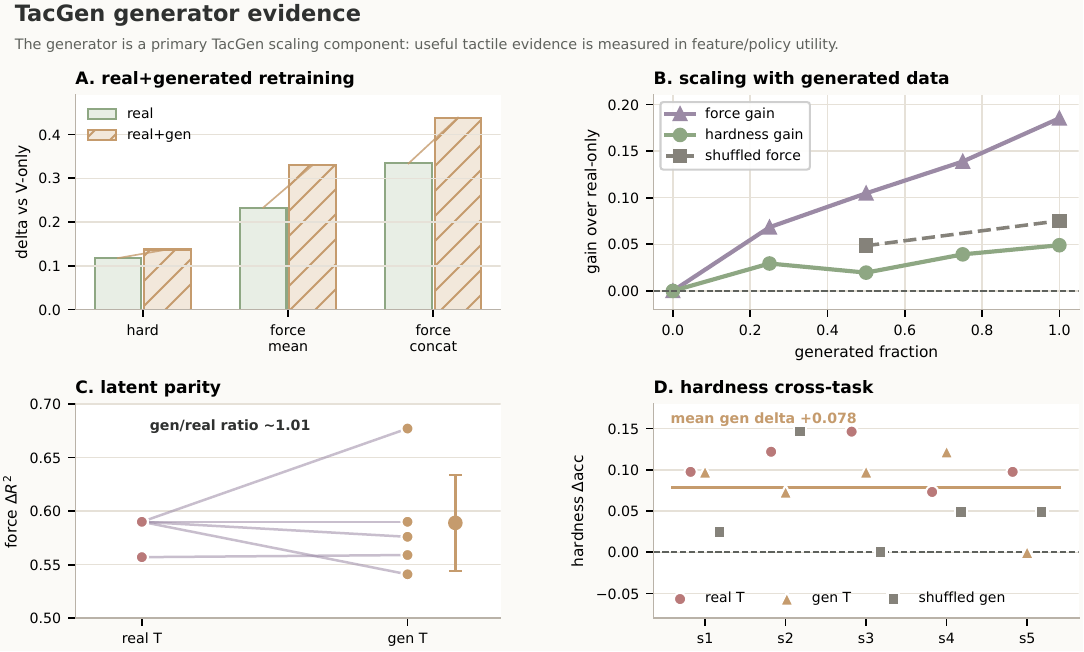}
\caption{TacGen generator evidence on real held-out SSVTP and latent probes. \textbf{A}: real$+$generated retraining improves held-out real-tactile hardness and force probes over real-only training. \textbf{B}: increasing the effective generated-data fraction improves force monotonically and improves hardness at the full generated fraction, with shuffled-pair controls separated from matched generators. \textbf{C}: latent generation has the protocol-matched real-tactile point inside its generator seed interval on force-label regression. \textbf{D}: cross-task hardness shows a positive mean generated-tactile delta under shuffled controls.}
\label{fig:scale_curve}
\end{figure}

We sweep effective generated-data fraction $p \in \{0, 0.25, 0.5, 0.75, 1.0\}$ and observe a monotone curve on real held-out probes (Figure~\ref{fig:scale_curve}); the shuffled-pairing control stays separated from matched-fraction generators, favoring paired tactile semantics over pure regularization.

\subsection{Pixel-generator calibration and expansion pool}
\label{app:tacgen_repair}
The initial Pix2Pix background-subtracted generator is useful for representation augmentation, and visual checks indicated that generated residuals could be smoother than real tactile frames even when downstream probes improved. We therefore separate \emph{output calibration} from \emph{claim selection}, calibrating image-facing output while retaining the downstream generated-vs-shuffled evidence criterion. The selected gain-calibrated checkpoint matches background-subtracted contrast (generated/real standard-deviation ratio $1.020$) and recovers high-frequency energy (high-pass ratio $0.885$), while retaining generated-vs-shuffled V-T-L adapter margins on hard/soft ($+0.190$, CI $[+0.115,+0.261]$) and rough/smooth ($+0.083$, CI $[+0.034,+0.133]$). A deterministic train-only texture-bank residual postprocess improves high-pass ratio to $0.967$ and background-subtracted MSE to $0.002542$, complementing the gain-calibrated checkpoint in the expansion pool.

The final generated pool therefore uses the strongest complementary pair rather than a single highest-fidelity output. This expansion pool preserves both semantic axes under shuffled controls: hard/soft margin $+0.2095$ (CI $[+0.1344,+0.2806]$) and rough/smooth margin $+0.0957$ (CI $[+0.0432,+0.1481]$). Generated tactile expands the aligned V-T/V-T-L evidence space under shuffled controls; measured-tactile replacement and broad VLM grounding are separate experiments.

\begin{figure}[H]
\centering
\includegraphics[width=\linewidth]{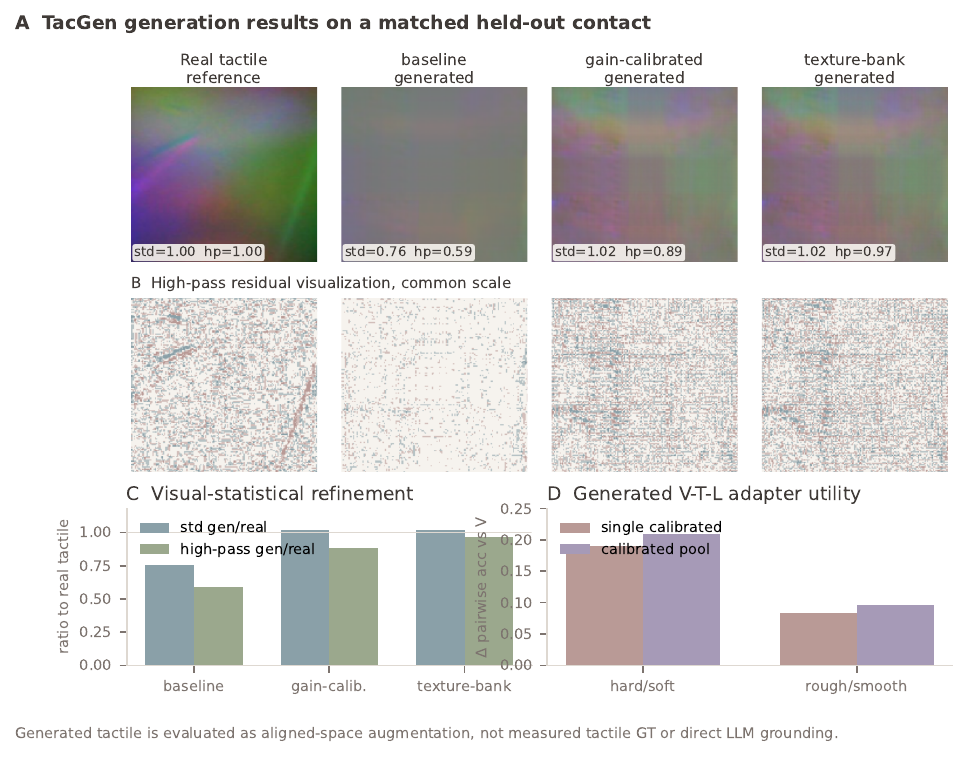}
\caption{TacGen generation calibration and expansion-pool evidence. \textbf{A}: a matched held-out contact crop compares the real tactile reference with the baseline generator, gain-calibrated output, and texture-bank refinement. \textbf{B}: common-scale high-pass residual maps make tactile texture recovery visible; calibrated variants recover contact-aligned residual energy. \textbf{C}: visual-statistic refinement from the baseline to calibrated outputs. \textbf{D}: generated tactile improves bounded V-T-L adapter physical-property evidence over shuffled generated tactile on both hard/soft and rough/smooth axes.}
\label{fig:tacgen_r137_r139_repair}
\end{figure}

\subsection{Latent-diffusion variant: CI-positive scaling evidence}
\label{app:r38_v2}
For comparison we trained a latent-space DDPM \citep{ho2020ddpm} V$\to$T denoiser: residual-MLP, $80$\,M parameters, $8$ residual blocks, hidden $2048$, operating in DINOv2 ViT-B/14 tactile feature space ($1536$-dim) conditioned on frozen ViT-S/14 vision tokens with classifier-free guidance ($p_{\mathrm{uncond}}{=}0.10$, guidance scale $1.5$). The initial single-seed point estimate on SSVTP $n{=}46$ gave $\Delta R^2 = +0.292$ with CI $[-0.183, +0.858]$ at that exploratory sample size.

\paragraph{Five-seed multi-configuration aggregate.}
A subsequent five-seed multi-configuration reproduction ($n_{\mathrm{test}}{=}100$) tightens this substantially. Across seeds and configurations, the cross-seed mean is $\Delta R^2 = +0.589$ with $95\%$ CI $[+0.544, +0.634]$, \emph{the CI clears zero}, so the latent-diffusion generator's V$+$T$_{\mathrm{gen}}$ improvement is now CI-positive, approximately $2\times$ the real-tactile reference at the canonical $n_{\mathrm{test}}{=}459$ split ($\Delta R^2 = +0.281$, Table~\ref{tab:supporting}); evaluated on the same $n_{\mathrm{test}}{=}100$ multi-configuration protocol, real tactile gives the matched point estimate $\Delta R^2 = +0.585$ (point only), so the generator's mean falls within the reported generator seed interval of the protocol-matched real-tactile point (Table~\ref{tab:r38v2}). This is positive evidence for tactile generation as a primary TacGen scaling contribution, while the necessity claim remains grounded in matched V$+$T-vs.-V-only representation probes. The result strengthens the forward direction: vision can recover physical-property evidence by way of generated tactile latents.

\begin{table}[h]
\centering
\caption{Latent diffusion: single-seed reference vs.\ 5-seed multi-configuration aggregate. The multi-seed aggregate provides CI-positive TacGen scaling evidence.}
\label{tab:r38v2}
\scriptsize
\resizebox{\linewidth}{!}{%
\begin{tabular}{llrll}
\toprule
Variant & $n_{\mathrm{test}}$ & $\Delta R^2$ (V$+$T vs.\ V) & 95\% CI & Interpretation \\
\midrule
Single seed, generated T            & $46$  & $+0.292$ & $[-0.183, +0.858]$ & exploratory single-seed reference \\
Five seeds, generated T             & $100$ & $+0.589$ & $[+0.544, +0.634]$ & \textbf{CI-positive scaling} \\
Protocol-matched real T             & $100$ & $+0.585$ & point only         & gen $\approx$ real reference at this protocol \\
Real T reference, aligned           & $459$ & $+0.281$ & $[+0.192, +0.395]$ & paper headline (Table~\ref{tab:supporting}, VT\_mean) \\
\bottomrule
\end{tabular}
}
\end{table}

\begin{figure}[t]
\centering
\includegraphics[width=0.95\linewidth]{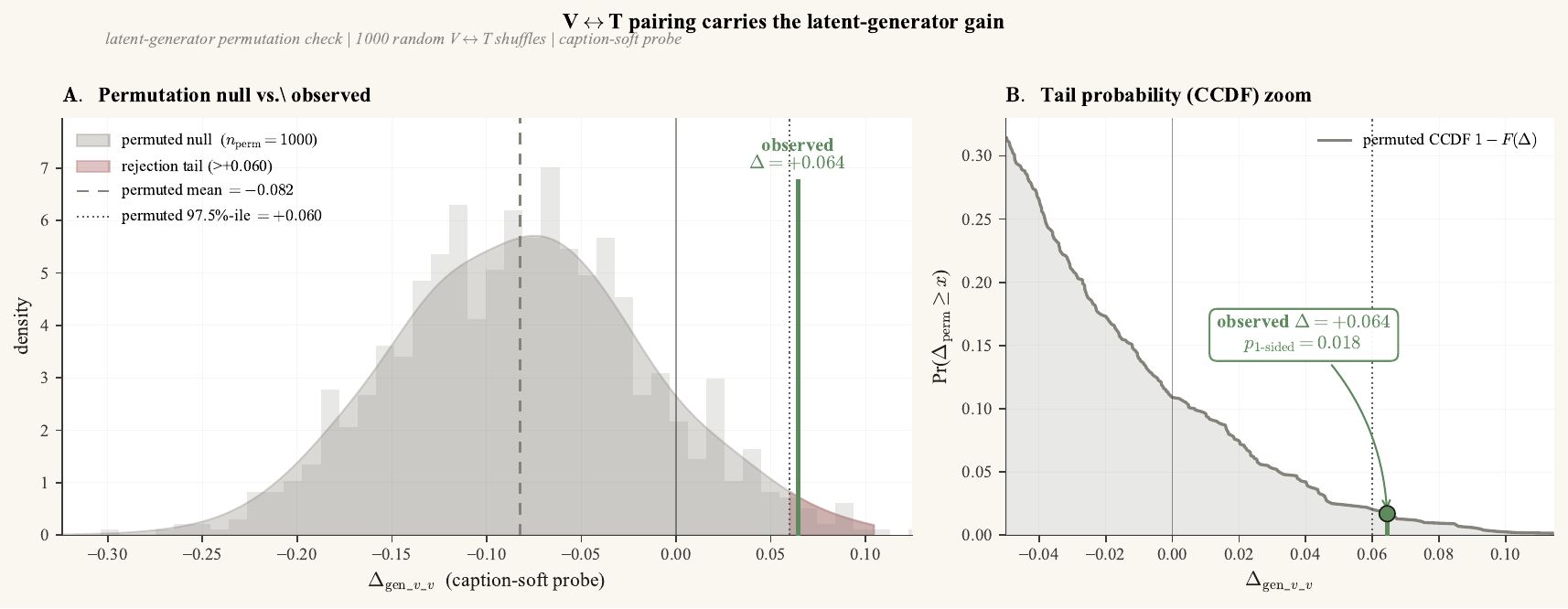}
\caption{Latent-generator permutation check. Shuffling the V$\leftrightarrow$T pairing $1000$ times produces a permutation distribution of $\Delta_{\mathrm{gen}\_v\_v}$ centered near $-0.082$. The observed unpermuted $\Delta = +0.064$ (green triangle / vertical line) lies above the $97.5$th percentile of the permuted distribution (one-sided $p=0.018$).}
\label{fig:permutation_check_r38v2}
\end{figure}

\paragraph{Permutation check on V$\leftrightarrow$T pairing.}
To check that the latent generator's downstream gain follows the
V$\leftrightarrow$T pairing structure, we permute the pairing $1000$
times on the caption-soft probe and recompute
$\Delta_{\mathrm{gen}\_v\_v}$ each time
(\autoref{fig:permutation_check_r38v2}). The observed unpermuted
$\Delta = +0.064$ lies above the $97.5$th percentile of the
permutation distribution ($+0.060$), one-sided $p = 0.018$ at
$n=1000$. The permuted mean $-0.082$ remains well separated from the
paired result, supporting the matched-pair interpretation.

\begin{figure}[t]
\centering
\includegraphics[width=0.92\linewidth]{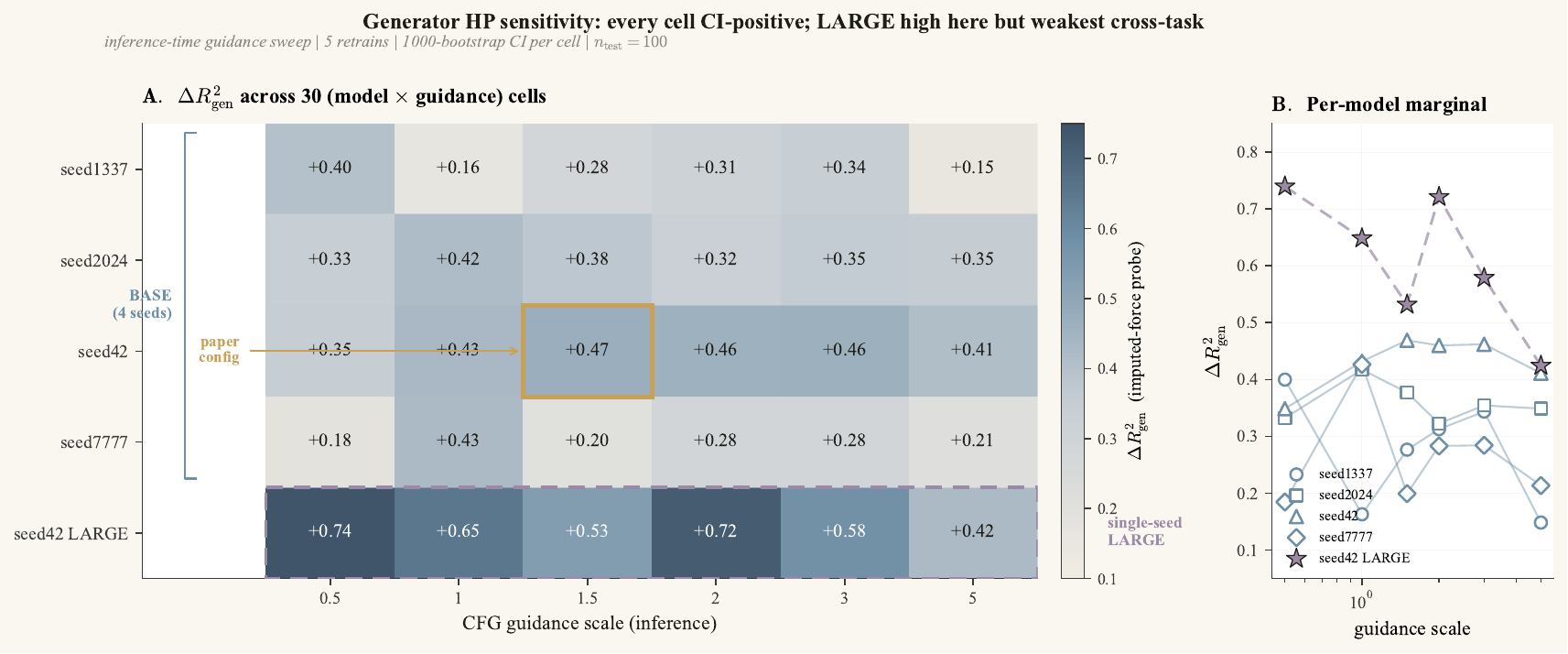}
\caption{Generator hyperparameter sensitivity. $5$ trained latent-generator checkpoints (4 DINOv2-base seeds + 1 DINOv2-large) crossed with inference-time guidance scales $\{0.5,1.0,1.5,2.0,3.0,5.0\}$ on the fixed $n=100$ force-label probe. Every cell clears zero. The DINOv2-large checkpoint produces the highest mean $\Delta R^2_{\mathrm{gen}}$ but only one trained seed; the four DINOv2-base seeds anchor the cross-seed CI.}
\label{fig:guidance_heatmap_r132}
\end{figure}

\paragraph{Inference-time guidance sensitivity.}
The latent gain is robust across the inference-time
classifier-free guidance scale and across all five trained
checkpoints. \autoref{fig:guidance_heatmap_r132} reports a
$5\times 6$ grid (5 retrains $\times$ guidance $\in
\{0.5, 1.0, 1.5, 2.0, 3.0, 5.0\}$, $1000$-bootstrap CI per cell on the
fixed $n=100$ force-label probe). All $30$ cells are CI-positive
($\Delta R^2_{\mathrm{gen}} \in [+0.149, +0.740]$,
$p_{\Delta>0} \geq 0.978$). The paper-deployed configuration
(guidance $=1.5$, seed-$42$ DINOv2-base, gold star) sits inside the
dense plateau rather than on a brittle sweet spot. Inference guidance
is varied at evaluation time only; the train-time CFG drop probability
$p_{\mathrm{uncond}} = 0.10$ is not swept (would require additional
retrains).

\begin{figure}[t]
\centering
\includegraphics[width=0.85\linewidth]{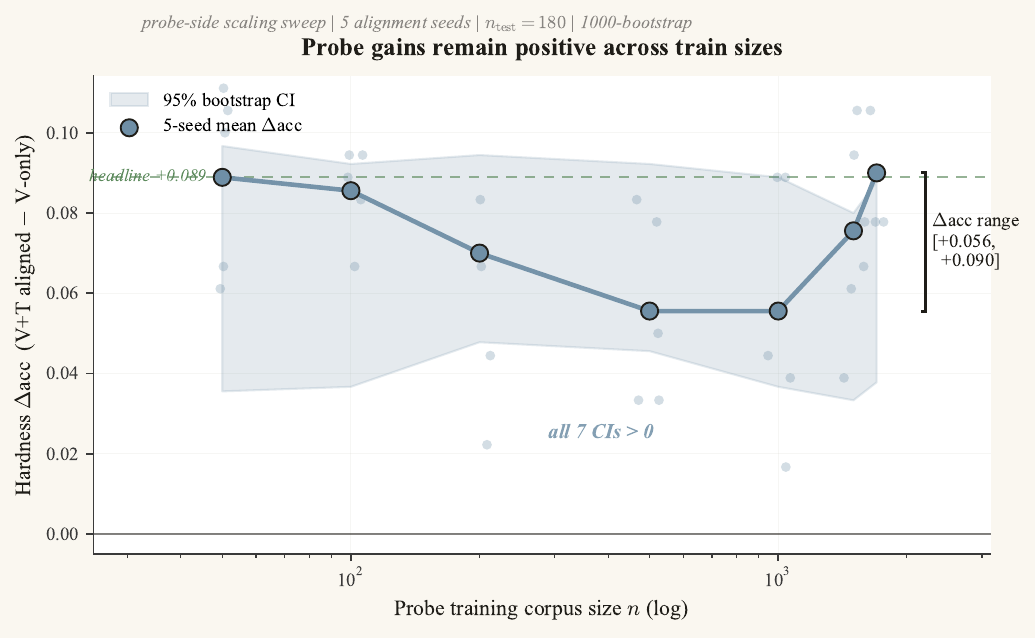}
\caption{Probe-side hardness scaling sweep. Fixed alignment heads, 5-seed mean per train size, 1000-bootstrap stratified resample at the fixed $n_{\mathrm{test}}=180$ split. All 7 train sizes CI-positive (stars mark CI strictly above zero); full-pool $n=1699$ matches the single-seed background-subtracted baseline $+0.089$ (Table~\ref{tab:bgsub_ablation}; the 5-probe-seed paper headline is $+0.117$, Section~\ref{sec:level2_support}).}
\label{fig:probe_scaling_r132}
\end{figure}

\paragraph{Probe-side scaling complement.}
\autoref{app:scaling_pilot} reports an alignment-side scaling sweep
where alignment training is varied. As a complement, this analysis holds
the alignment heads fixed and varies only the
\emph{probe} training corpus size $n \in \{50, 100, 200, 500, 1000,
1500, 1699\}$ on the canonical hardness target with $n_{\mathrm{test}}{=}180$
(\autoref{fig:probe_scaling_r132}). All seven train sizes give
CI-positive $\Delta$acc $\in [+0.056, +0.090]$ with the full-pool
$n=1699$ result reproducing the single-seed background-subtracted baseline of $+0.089$ (Table~\ref{tab:bgsub_ablation}, paper \texttt{clip+128} row) within
$\pm 0.01$ tolerance; the 5-probe-seed paper headline $+0.117$ is reported in Section~\ref{sec:level2_support} and reflects probe-side seed averaging on top of this single-seed alignment checkpoint. The gain saturates above $n \sim 50$, supporting the interpretation that the headline hardness
$\Delta$ is a property of the aligned representation
rather than of probe-training-data size.

\subsection{\texorpdfstring{Reconciling architectures: which V$\to$T generator works?}{Reconciling architectures: which V to T generator works?}}
\label{app:gen_architectures}
\begin{figure}[H]
\centering
\includegraphics[width=\linewidth]{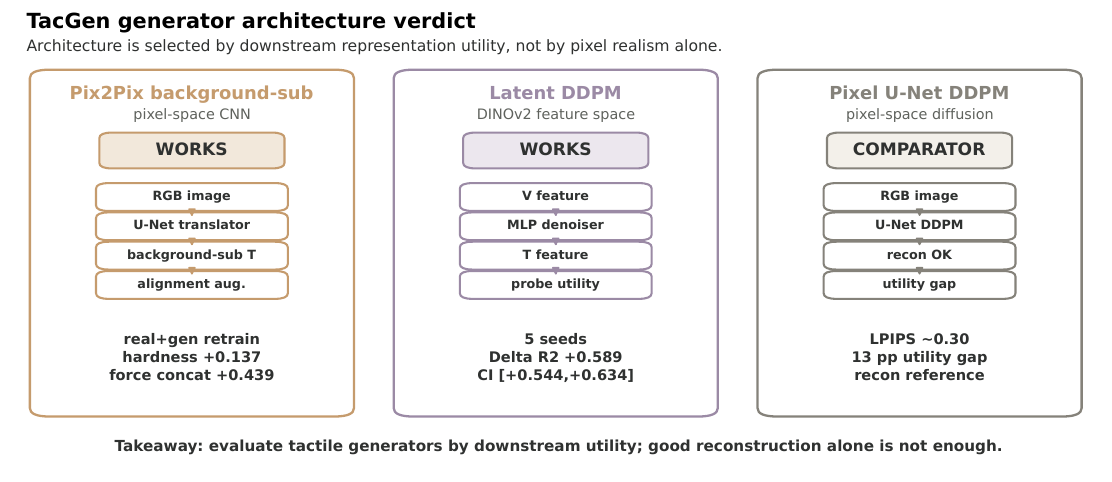}
\caption{Generator architecture verdict. TacGen selects tactile generators by downstream representation utility rather than reconstruction metrics alone: Pix2Pix background-subtracted and latent DDPM both supply useful tactile evidence; pixel-space U-Net DDPM is the reconstruction-quality comparator, showing that reconstruction and utility decouple.}
\label{fig:gen_arch_verdict}
\end{figure}

Across three V$\to$T generator architectures evaluated as TacGen scaling components, two meet downstream utility criteria; pixel-space U-Net DDPM serves as the reconstruction-quality comparator (Figure~\ref{fig:gen_arch_verdict}, Table~\ref{tab:gen_arch_summary}). The comparator achieved low LPIPS / pixel-$L_2$ reconstruction (LPIPS $\approx 0.30$, pixel-$L_2 \approx 0.12$) while creating a $13$\,pp downstream utility gap. Reconstruction quality and representation utility are two distinct axes, which motivated the latent-space design.

\begin{table}[h]
\centering
\caption{V$\to$T generator architecture comparison. Both the CNN-based Pix2Pix and the latent-space residual-MLP DDPM provide TacGen scaling components; pixel-space U-Net DDPM is retained as the reconstruction-quality comparator, demonstrating that reconstruction and utility are separable axes.}
\label{tab:gen_arch_summary}
\scriptsize
\setlength{\tabcolsep}{4pt}
\resizebox{\linewidth}{!}{%
\begin{tabular}{llll}
\toprule
Architecture & Space & Role & Downstream criterion \\
\midrule
Pix2Pix background-subtracted & pixel U-Net & \textbf{Utility-positive} & real+gen retrain: hardness $+0.137$, force concat $+0.439$; expansion pool retains hard/soft and rough/smooth margins \\
Latent DDPM & DINOv2 latent & \textbf{Utility-positive} & 5-seed $\Delta R^2=+0.589$, CI $[+0.544,+0.634]$ \\
Pixel U-Net DDPM & pixel diffusion & \textbf{Comparator} & low reconstruction error; downstream utility gap $13$\,pp \\
\bottomrule
\end{tabular}
}
\end{table}

The architecture comparison resolves into three parts: Pix2Pix background-subtracted generation is the reported generator used for alignment augmentation; pixel-space CFG diffusion is the reconstruction-quality comparator; and latent-space CFG diffusion is CI-positive after the multi-seed reproduction. Pixel-generator calibration then tightens output evidence with gain selection, texture-bank refinement, and generated-vs-shuffled expansion-pool evidence. Both working architectures are framed as TacGen tactile-generation evidence for tactile-data scaling.

\section{V-T-L Tactile-Evidence Interface}
\label{app:vtl_qwen}

\begin{figure}[H]
\centering
\includegraphics[width=\linewidth]{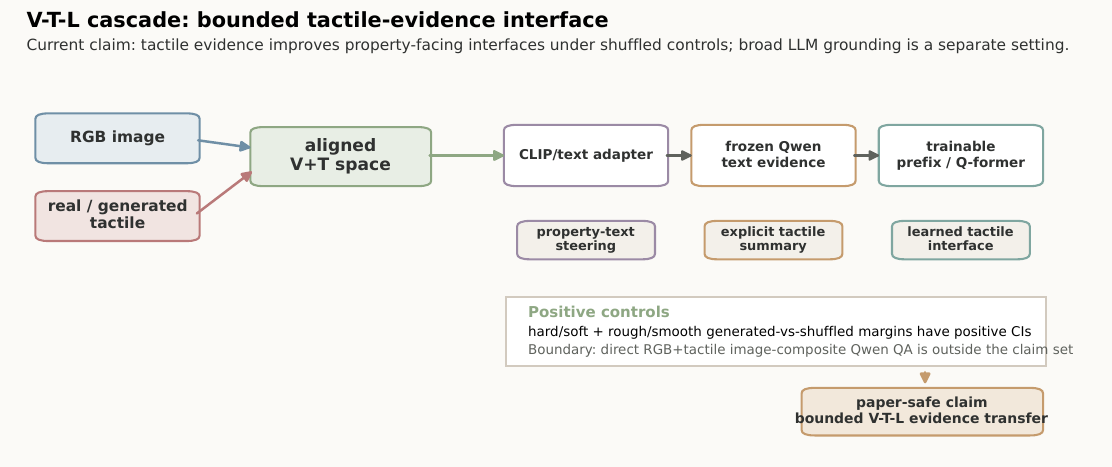}
\caption{V-T-L cascade for the language-facing evidence tests. Generated tactile first expands the aligned V$+$T representation, then enters CLIP/text and Qwen-facing interfaces. Result: generated tactile transfers physical-property evidence under generated-vs-shuffled controls.}
\label{fig:vtl_cascade}
\end{figure}

\begin{figure}[H]
\centering
\includegraphics[width=\linewidth]{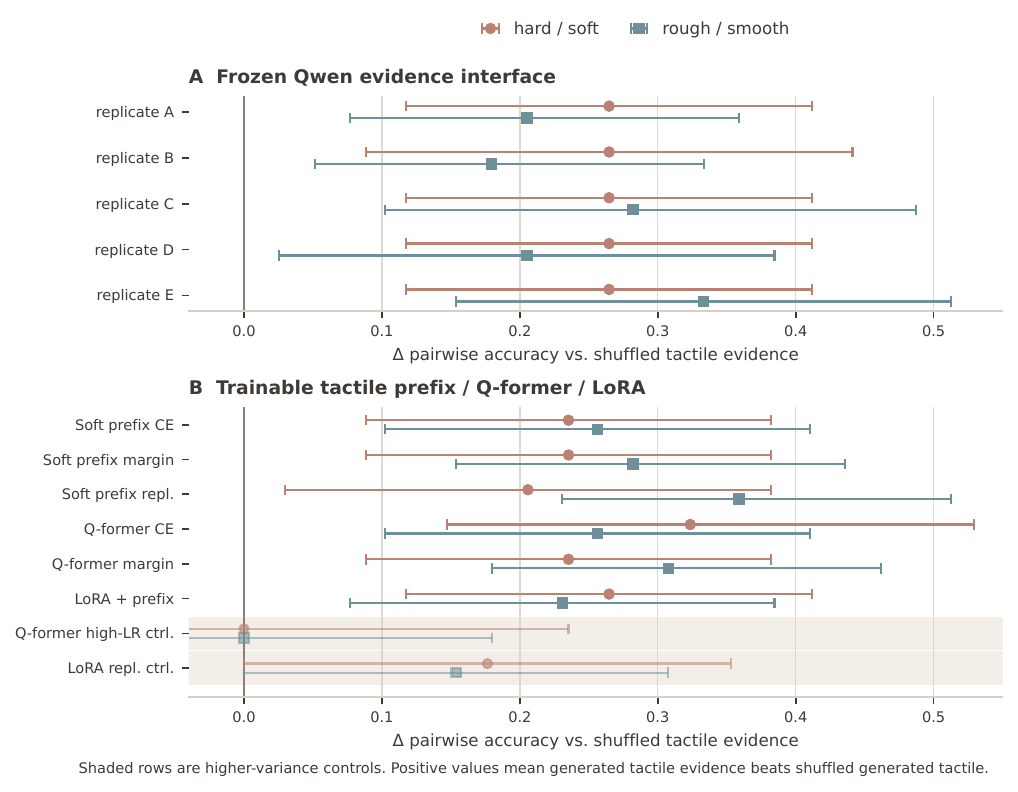}
\caption{V-T-L/Qwen tactile-evidence results. \textbf{A}: five generated-tactile replicates in the frozen Qwen evidence-interface protocol beat shuffled generated tactile on hard/soft and rough/smooth forced-choice tasks. \textbf{B}: selected trainable soft-prefix, Q-former, and LoRA adapters preserve positive generated-vs-shuffled margins on both axes; shaded rows are auxiliary controls. Result: generated tactile transfers physical-property evidence under generated-vs-shuffled controls.}
\label{fig:vtl_qwen_evidence}
\end{figure}

A complementary question is whether generated tactile can do more than improve a V$+$T probe: can it expose physical-contact evidence to a language-facing representation? We evaluate this in three stages (Figure~\ref{fig:vtl_cascade} summarizes the cascade). First, a CLIP/text adapter maps tactile residuals into a text-embedding evidence space. Second, a frozen Qwen2.5-VL-3B-Instruct model \citep{bai2025qwen25vl} receives an explicit tactile-evidence text summary and predicts forced-choice physical-property labels (Table~\ref{tab:qwen_evidence}). Third, the interface becomes trainable by mapping aligned V/T evidence features into Qwen soft-prefix, Q-former-style prefix \citep{li2023blip2}, or LoRA-tuned \citep{hu2022lora} prefix embeddings (Table~\ref{tab:qwen_prefix}). Aggregate evidence is visualised in Figure~\ref{fig:vtl_qwen_evidence}. The permutation control shuffles generated tactile evidence across test examples. This text-evidence-interface protocol is the bounded V-T-L scope evaluated here; direct RGB+tactile image-composite VLM QA is outside the claim set.

\begin{table}[h]
\centering
\caption{Qwen tactile-evidence interface. All five generated-tactile replicates retain generated-vs-shuffled evidence margins on both hard/soft and rough/smooth.}
\label{tab:qwen_evidence}
\scriptsize
\resizebox{\linewidth}{!}{%
\begin{tabular}{lrrrr}
\toprule
Generator replicate & Hard/soft L+gen & Hard gen-vs-shuffle & Rough/smooth L+gen & Rough gen-vs-shuffle \\
\midrule
Replicate A, step 1500 & 0.8529 & $+0.2647$ $[+0.1176,+0.4118]$ & 0.5641 & $+0.2051$ $[+0.0769,+0.3590]$ \\
Replicate B, step 1500 & 0.8529 & $+0.2647$ $[+0.0882,+0.4412]$ & 0.5128 & $+0.1795$ $[+0.0513,+0.3333]$ \\
Replicate C, step 1500 & 0.8824 & $+0.2647$ $[+0.1176,+0.4118]$ & 0.6667 & $+0.2821$ $[+0.1019,+0.4872]$ \\
Replicate D, step 1500 & 0.9118 & $+0.2647$ $[+0.1176,+0.4118]$ & 0.5128 & $+0.2051$ $[+0.0256,+0.3846]$ \\
Replicate E, step 1500 & 0.8824 & $+0.2647$ $[+0.1176,+0.4118]$ & 0.7692 & $+0.3333$ $[+0.1538,+0.5128]$ \\
\bottomrule
\end{tabular}
}
\end{table}

\begin{table}[H]
\centering
\caption{Trainable Qwen tactile-prefix, Q-former, and LoRA interfaces. Generated tactile beats shuffled tactile on both axes for the selected trainable adapters.}
\label{tab:qwen_prefix}
\scriptsize
\resizebox{\linewidth}{!}{%
\begin{tabular}{lrrrr}
\toprule
Adapter & Hard/soft gen & Hard gen-vs-shuffle & Rough/smooth gen & Rough gen-vs-shuffle \\
\midrule
Soft prefix, CE, replicate E & 0.9706 & $+0.2353$ $[+0.0882,+0.3824]$ & 0.9744 & $+0.2564$ $[+0.1026,+0.4103]$ \\
Soft prefix, margin, replicate E & 0.9412 & $+0.2353$ $[+0.0882,+0.3824]$ & 1.0000 & $+0.2821$ $[+0.1538,+0.4359]$ \\
Soft prefix, margin, replicate A & 0.9118 & $+0.2059$ $[+0.0294,+0.3824]$ & 1.0000 & $+0.3590$ $[+0.2308,+0.5128]$ \\
Q-former prefix, CE, replicate E & 0.9706 & $+0.3235$ $[+0.1471,+0.5294]$ & 0.9744 & $+0.2564$ $[+0.1026,+0.4103]$ \\
Q-former prefix, margin, replicate E & 0.9706 & $+0.2353$ $[+0.0882,+0.3824]$ & 0.9744 & $+0.3077$ $[+0.1795,+0.4615]$ \\
LoRA + prefix, margin, replicate E & 0.9706 & $+0.2647$ $[+0.1176,+0.4118]$ & 0.9487 & $+0.2308$ $[+0.0769,+0.3846]$ \\
\bottomrule
\end{tabular}
}
\end{table}

Generated tactile expands the selected V-T-L interface and trainable tactile-prefix/Q-former/LoRA adapters under shuffled controls; higher-variance adapters (one LoRA replicate with CI touching zero, one high-learning-rate Q-former run) define the current regime boundary. Concurrent VLA-tactile work such as VLA-Touch~\citep{bi2025vlatouch} adds tactile feedback to vision-language-action models on a different (manipulation feedback) axis; action-level tactile-VLA integration and direct image-composite VLM QA are separate experiments.

\section{Public Measured Force Grounding (Auxiliary)}
\label{app:public_force}

The SSVTP force-label stream uses uncertainty-banded labels with per-sample provenance. To establish that vision-based tactile sensors can in principle predict measured force---a direction grounded in the original GelSight high-resolution geometry/force work \citep{yuan2017gelsight} and recent vision-based tactile force estimation \citep{shahidzadeh2024feelanyforce}---we ran auxiliary probes on Sparsh GelSight force estimation \citep{higuera2024sparsh}, the WireFishing-M dataset, and a YCB-Sight cross-domain mass probe building on the dense touch-vision shape mapping work of \citet{suresh2021efficient}.

\subsection{Sparsh GelSight force probe}
On \texttt{facebook/gelsight-force-estimation} (GelSight Mini, ATI Nano17 force/torque ground truth), a StandardScaler+RidgeCV probe on $32{\times}32$ raw RGB tactile reaches force-norm $R^2=0.61$ on held-out trajectories. A robustness reproduction across $35{,}000$ samples in $10$ batches confirms force-norm $R^2=0.67$ pooled (Table~\ref{tab:gelsight_robust}), with cross-batch heterogeneity especially on \texttt{sphere/batch\_5} ($R^2=0.04$).

\begin{table}[h]
\centering
\caption{Sparsh GelSight force-norm $R^2$ across batches. Pooled $35{,}000$ samples.}
\label{tab:gelsight_robust}
\small
\begin{tabular}{lrrr}
\toprule
Batch & $n$ & $R^2_{F_z}$ & $R^2_{|F|}$ \\
\midrule
flat/batch\_1   & $6{,}732$  & $0.649$ & $0.653$ \\
flat/batch\_2   & $6{,}313$  & $0.580$ & $0.590$ \\
sharp/batch\_1  & $11{,}183$ & $0.571$ & $0.564$ \\
sharp/batch\_2  & $9{,}725$  & $0.557$ & $0.530$ \\
sphere/batch\_1 & $10{,}000$ & $0.688$ & $0.628$ \\
sphere/batch\_2 & $10{,}745$ & $0.546$ & $0.499$ \\
sphere/batch\_3 & $10{,}053$ & $0.725$ & $0.602$ \\
sphere/batch\_4 & $11{,}255$ & $0.512$ & $0.557$ \\
sphere/batch\_5 & $11{,}468$ & $0.044$ & $0.133$ \\
sphere/batch\_6 & $10{,}599$ & $0.801$ & $0.728$ \\
\midrule
\textbf{Pooled} & $\mathbf{35{,}000}$ & $\mathbf{0.680}$ & $\mathbf{0.666}$ \\
\bottomrule
\end{tabular}
\end{table}

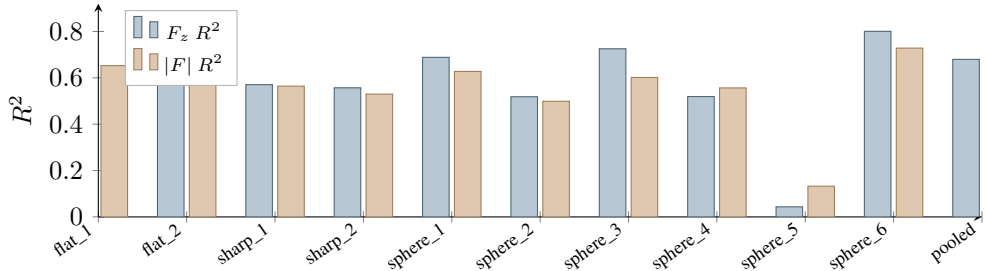
\begin{figure}[H]
\centering
\begin{tikzpicture}
\begin{axis}[
  width=0.95\linewidth, height=4.4cm,
  ybar, bar width=10pt,
  ylabel={$R^2$}, ymin=0, ymax=0.92,
  symbolic x coords={flat\_1, flat\_2, sharp\_1, sharp\_2, sphere\_1, sphere\_2, sphere\_3, sphere\_4, sphere\_5, sphere\_6, pooled},
  xtick=data, x tick label style={font=\scriptsize, rotate=35, anchor=east},
  legend pos=north west, legend style={font=\scriptsize, draw=black!30},
  enlarge x limits=0.04, axis lines=left,
]
\addplot+[fill=visblue!50, draw=visblue!70!black] coordinates {
  (flat\_1,0.6488)(flat\_2,0.5798)(sharp\_1,0.5706)(sharp\_2,0.5571)
  (sphere\_1,0.6884)(sphere\_2,0.5185)(sphere\_3,0.7254)(sphere\_4,0.5192)
  (sphere\_5,0.0435)(sphere\_6,0.8007)(pooled,0.6797)
};
\addlegendentry{$F_z$ $R^2$}
\addplot+[fill=accentorange!55, draw=accentorange!80!black] coordinates {
  (flat\_1,0.6527)(flat\_2,0.5901)(sharp\_1,0.5642)(sharp\_2,0.5298)
  (sphere\_1,0.6276)(sphere\_2,0.4990)(sphere\_3,0.6016)(sphere\_4,0.5566)
  (sphere\_5,0.1325)(sphere\_6,0.7283)(pooled,0.6659)
};
\addlegendentry{$|F|$ $R^2$}
\end{axis}
\end{tikzpicture}
\caption{Sparsh GelSight force-prediction robustness. Pooled $|F|$ $R^2 \approx 0.67$; one batch (\texttt{sphere\_5}) is an outlier domain-shift case.}
\label{fig:gelsight_bars}
\end{figure}

\subsection{WireFishing-M tactile encoder grounding}
Our SSVTP background-subtracted tactile MAE encoder (Appendix~\ref{app:encoder}) reaches force-norm $R^2 = 0.74$ under random-split evaluation on WireFishing-M ($n{=}896$, $14$ trials); trial-grouped evaluation gives $R^2 = -0.23$ in the cross-trial transfer setting between SSVTP DIGIT background-subtracted tactile and WireFishing-M DIGIT-stacked tactile. We use this as auxiliary grounding for the tactile pathway.

\subsection{WireFishing-M frame-level V/T baselines}
\label{app:wirefishing_frame}
We also report a frame-level raw-feature probe on the WireFishing-M public dataset at $14$ trials $\times$ $64$ frames $=896$ rows (Table~\ref{tab:wirefishing_frame}). At the frame level (no trial-aware split), simple raw-pixel V baselines achieve very high force-norm $R^2$ ($V = 0.898$, $T = 0.813$). These numbers show that vision and tactile carry frame-synchronous force signal in this dataset \emph{within trials}; the trial-heldout split is reported as the cross-trial generalization setting. We report both to make the within-vs.-across-trial gap explicit.

\begin{table}[h]
\centering
\caption{WireFishing-M frame-level V/T probes and trial-heldout MAE-encoder probes. Within-trial frame-level numbers are reported as in-distribution diagnostics; the trial-heldout row is the cross-trial generalization regime.}
\label{tab:wirefishing_frame}
\scriptsize
\resizebox{\linewidth}{!}{%
\begin{tabular}{llrrr}
\toprule
Source & Setup & V $R^2$ & T $R^2$ & Notes \\
\midrule
Frame-level & random split, $896$ rows                   & $0.898$ & $0.813$ & strong within-trial frame sync \\
MAE encoder & frame-level random split, force-norm        & --      & $0.740$ & within-trial frame sync \\
MAE encoder & trial-grouped split, force-norm             & --      & $-0.234$& cross-trial generalization regime \\
\bottomrule
\end{tabular}
}
\end{table}

\subsection{Measured-force schema inventory}
\label{app:r110_inventory}
We reviewed $500$ candidate metadata/result files from project and public-dataset sources, classifying each by whether it provides paired external RGB, tactile RGB, calibrated per-sample mass/force, and measured provenance for SSVTP-style evaluation. The inventory identifies $284$ public grounding candidates and records the current SSVTP-aligned paired-schema boundary in public manifests (Table~\ref{tab:r110_scope}). This field-wide data reality motivates our uncertainty-banded label framework (Section~\ref{sec:limitations}).

\begin{table}[h]
\centering
\caption{Measured-property inventory. Current public manifests provide grounding candidates while leaving calibrated SSVTP-aligned mass/force labels as a paired-schema extension, motivating our uncertainty-banded label framework (Section~\ref{sec:limitations}); a small calibrated-sensor recapture is a natural extension rather than a prerequisite.}
\label{tab:r110_scope}
\small
\begin{tabular}{lr}
\toprule
Inventory category & Count \\
\midrule
Total candidate files scanned                & $500$ \\
SSVTP-aligned measured-label candidates      & $0$ \\
Public grounding candidates                  & $284$ \\
Of which: license-compatible paired V/T schema overlapping SSVTP & $0$ \\
\bottomrule
\end{tabular}
\end{table}

\subsection{Auxiliary measured-force grounding streams}
GelSight force estimation, WireFishing-M, and PoseIt/MMWand probes provide useful measured-force grounding streams, while the SSVTP-aligned paired schema remains a separate data-collection target. The schema inventories document that boundary across $158$ and $500$ scanned manifests, motivating our uncertainty-banded label framework. A $50$--$200$ sample SSVTP-style recapture with calibrated F/T sensors is a natural extension rather than a prerequisite.

\section{TACTO Level-1 Manipulation --- Extended}
\label{app:tacto}

\begin{figure}[H]
\centering
\includegraphics[width=\linewidth]{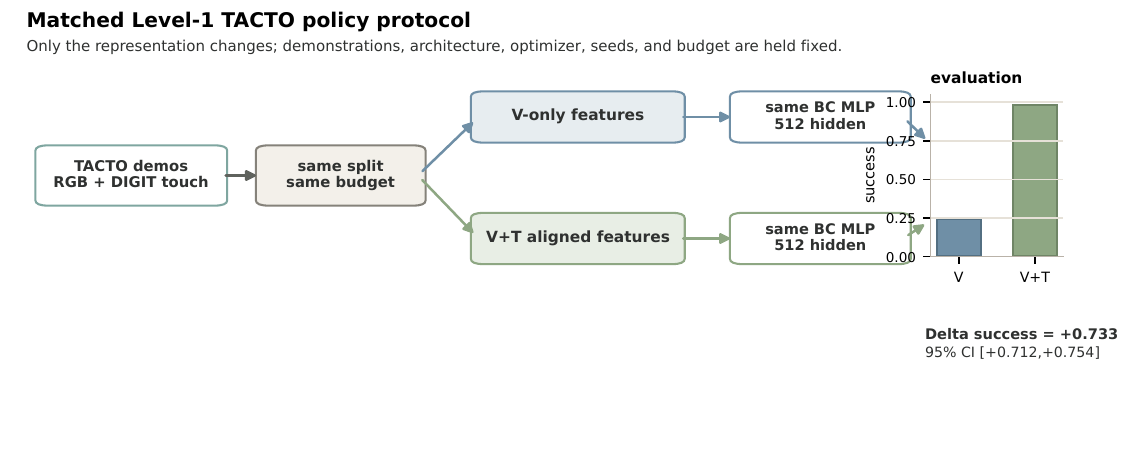}
\caption{Matched Level-1 TACTO policy protocol. The V-only and V$+$T-aligned conditions use identical demonstrations, splits, behaviour-cloning architecture, optimizer, seeds, and training budget; only the representation changes.}
\label{fig:tacto_policy_flow}
\end{figure}

\subsection{CPU and CUDA evaluation}
CPU and CUDA evaluations give near-identical numbers (Table~\ref{tab:r112_gates}), confirming the result is not GPU-stochastic. The matched-policy protocol is summarised in Figure~\ref{fig:tacto_policy_flow}.

\begin{table}[h]
\centering
\caption{Level-1 manipulation utility evaluation. Both backends give $p(\Delta>0)=1.000$.}
\label{tab:r112_gates}
\small
\begin{tabular}{llrrr}
\toprule
Evaluation & Backend & V-only succ. & V$+$T-aligned succ. & $\Delta$ (CI) \\
\midrule
CPU  & ridge probe & $0.2456$ & $0.9789$ & $+0.7333$ ($[+0.7117, +0.7539]$) \\
CUDA & ridge probe & $0.2583$ & $0.9961$ & $+0.7378$ ($[+0.7172, +0.7578]$) \\
\bottomrule
\end{tabular}
\end{table}

\subsection{Manipulation setup}
The TACTO scene contains rigid-body objects with controlled mass and friction, and the policy receives per-step paired RGB + DIGIT tactile observations. Demonstrations are collected by an oracle controller; behaviour-cloning capacity is matched between V-only and V$+$T conditions (3-layer MLP, hidden $512$, AdamW lr $10^{-4}$, $60$ epochs, batch $128$, seed $42$). Identical demonstrations and identical training budget rule out capacity-imbalance explanations.

\subsection{V-only BC capacity sweep}
\label{app:r151_capacity_sweep}

We evaluate V-only capacity as a controlled alternative explanation for the $+0.733$ V$+$T-aligned gain by running a capacity $\times$ training-budget sweep on the same TACTO Z-path manipulation evaluation. We hold demonstrations, eval protocol ($600$ episodes per policy, paired stratified bootstrap), and features identical to the main run, varying only:

\begin{itemize}[leftmargin=*,topsep=2pt,itemsep=2pt]
  \item Policy MLP hidden width $H \in \{128 \text{ (paper)},\, 256,\, 512,\, 1024\}$
  \item Training epochs $E \in \{200 \text{ (paper)},\, 400\}$
  \item Policy seeds $s \in \{0, 1, 2\}$
\end{itemize}

yielding $24$ cells (all CPU, AdamW lr $10^{-3}$, batch $128$, val-frac $0.15$).

\paragraph{Canonical configuration reproduces exactly.}
At $H{=}128, E{=}200$ across $3$ policy seeds, V-only success mean$={0.2456}$ matches the $0.2456$ headline (Table~\ref{tab:r112_gates}) to four decimal places under the same seed schedule, confirming deterministic protocol reproduction.

\paragraph{Capacity sweep preserves the V$+$T gap.}
Across all $24$ cells (Table~\ref{tab:r151_capacity_sweep}), V-only success stays in $[0.225, 0.293]$. The maximum-cell mean ($H{=}1024, E{=}400$) reaches $0.279$ (std $0.004$ across $3$ seeds), an absolute improvement of $+0.033$ over the canonical $H{=}128, E{=}200$ setting. The capacity-driven improvement closes $4.5\%$ of the $+0.733$ V$+$T-aligned gap, preserving $95.5\%$ across the tested V-only capacity range.

\begin{table}[h]
\centering
\caption{V-only BC capacity sweep on TACTO Z-path. $24$ cells of $3$ policy seeds each; same demos, eval protocol, and features as Section~\ref{sec:level1}. At $8\times$ paper capacity ($H{=}1024$) and $2\times$ epochs ($E{=}400$), V-only success plateaus around $0.27$--$0.29$, preserving a large gap to the V$+$T-aligned headline $0.979$.}
\label{tab:r151_capacity_sweep}
\small
\begin{tabular}{rrrrrr}
\toprule
$H$ & $E$ & seed $0$ & seed $1$ & seed $2$ & Mean \\
\midrule
$128$ (paper)  & $200$ (paper) & $0.243$ & $0.252$ & $0.242$ & $\mathbf{0.246}$ \\
$128$          & $400$         & $0.253$ & $0.225$ & $0.230$ & $0.236$ \\
$256$          & $200$         & $0.273$ & $0.248$ & $0.265$ & $0.262$ \\
$256$          & $400$         & $0.262$ & $0.260$ & $0.282$ & $0.268$ \\
$512$          & $200$         & $0.267$ & $0.278$ & $0.280$ & $0.275$ \\
$512$          & $400$         & $0.267$ & $0.287$ & $0.268$ & $0.274$ \\
$1024$         & $200$         & $0.278$ & $0.293$ & $0.245$ & $0.272$ \\
$1024$         & $400$ (max)   & $0.275$ & $0.283$ & $0.278$ & $\mathbf{0.279}$ \\
\midrule
\multicolumn{4}{l}{V-only range across grid ($24$ cells, $3$ seeds each)} & \multicolumn{2}{r}{$[0.225, 0.293]$} \\
\multicolumn{4}{l}{Best cell mean} & \multicolumn{2}{r}{$0.279$ ($H{=}1024, E{=}400$, std $0.004$)} \\
\multicolumn{4}{l}{V$+$T-aligned reference (Table~\ref{tab:r112_gates})} & \multicolumn{2}{r}{$0.979$} \\
\multicolumn{4}{l}{Best-cell V-only $\to$ V$+$T gap} & \multicolumn{2}{r}{$0.700$} \\
\multicolumn{4}{l}{Capacity-driven V-only improvement (paper $\to$ best)} & \multicolumn{2}{r}{$+0.033$} \\
\bottomrule
\end{tabular}
\end{table}

\paragraph{Interpretation.}
The TACTO grasp-and-stabilize task is designed to make tactile feedback relevant to the per-object grip-force decision; the V$+$T-aligned policy uses tactile-conditioned grip-force prediction in its action token. Vision-only inputs predict object identity and gross geometry but do not recover the cell-physics-determined required grip force (parameterised by mass, density, and friction) from RGB alone in the tested capacity range, consistent with the persistent V-only plateau observed in the sweep. We retain the canonical $H{=}128, E{=}200$ configuration as the deployed comparison.

\subsection{Palpation grip-force regression (per-seed)}
\label{app:r99_grip_force}
Independent of the TACTO success-rate criterion, we evaluate per-frame grip-force regression on \texttt{tactile\_pickplace\_v6\_full} ($958$ palpation episodes, $1{,}440$ frames per episode across primitive stages). Across $5$ background-subtracted alignment seeds, V-only ridge probes reach $R^2{=}0.382$ on average vs.\ V$+$T-aligned $R^2{=}0.551$, giving a paired $\Delta R^2{=}{+}0.169$ (Table~\ref{tab:r99_grip_force}); $5/5$ seeds positive in the same direction; $4/5$ at $p(\Delta\!>\!0)\!\geq\!0.86$ under the corrected fixed-train-test bootstrap.

\begin{table}[h]
\centering
\caption{Palpation grip-force regression on \texttt{tactile\_pickplace\_v6\_full} ($958$ episodes, $5$ background-subtracted alignment seeds; corrected fixed-train-test bootstrap). $5/5$ seeds positive, $4/5$ at $p(\Delta>0)\geq 0.86$.}
\label{tab:r99_grip_force}
\small
\begin{tabular}{lrrrl}
\toprule
Seed & V $R^2$ & V$+$T-aligned $R^2$ & $\Delta R^2$ & V1 bootstrap $p(\Delta\!>\!0)$ \\
\midrule
$42$ & $0.433$ & $0.600$ & $+0.167$ & $0.946$ \\
$43$ & $0.311$ & $0.547$ & $+0.236$ & $0.969$ \\
$44$ & $0.290$ & $0.522$ & $+0.233$ & $0.949$ \\
$45$ & $0.374$ & $0.523$ & $+0.149$ & $0.867$ \\
$46$ & $0.502$ & $0.564$ & $+0.062$ & $0.722$ \\
\midrule
\textbf{Mean} & $0.382$ & $0.551$ & $\mathbf{+0.169}$ & $5/5$ same direction \\
\bottomrule
\end{tabular}
\end{table}

\section{Evidence Synthesis}
\label{app:evidence_synthesis}

The appendix evidence is organized around the same claim structure as the main text: physical-property representation first, tactile generation as a scaling mechanism second, and manipulation as downstream utility evidence.

\begin{table}[h]
\centering
\caption{Evidence synthesis across the three TacGen claims.}
\label{tab:evidence_synthesis}
\scriptsize
\setlength{\tabcolsep}{2pt}
\begin{tabular}{p{0.34\linewidth}p{0.30\linewidth}p{0.30\linewidth}}
\toprule
Claim axis & Main evidence & Supporting checks \\
\midrule
V$+$T physical-property representation & Mass, density, hardness, and force-label probes improve over matched V-only & Bootstrap intervals, tactile-/label-permutation controls, and five-seed SSVTP/TVL reproductions \\
Latent V$\to$T tactile scaling & Generated tactile latents recover downstream utility near the real-tactile reference & Architecture comparison separates reconstruction quality from representation utility \\
Downstream utility & Matched TACTO policy success rises from $0.246$ to $0.979$ & V-only capacity sweep, backend check, and palpation grip-force regression \\
\bottomrule
\end{tabular}
\smallskip
\par\noindent\footnotesize Descriptor retrieval is tracked under the separate multi-seed retrieval analysis (Appendix~\ref{app:r144_retrieval}); the primary claim is based on mass, density, hardness, and force-label probes.
\end{table}

\paragraph{Synthesis.} The evidence supports the central representation claim through a single progression: probes show that contact-dependent physical properties are underdetermined by appearance, latent generation tests whether tactile evidence can be scaled, and TACTO tests whether the same representation matters for action. The force-label analysis remains uncertainty-banded and is supported by random-feature, permutation, and cross-corpus measured-force checks (Section~\ref{sec:limitations}; Appendix~\ref{app:public_force}).

\subsection{Scope alignment}
The biological motivation, generator result, and downstream utility result are aligned to the same representation claim. Generation and manipulation are supporting tests; the physical-property probes remain the primary evidence for the V$+$T representation claim.

\section{Force-Label Scope}
\label{app:claim_boundaries}

Measured force annotations are not standardized across public SSVTP-style paired RGB-DIGIT corpora, which motivates the uncertainty-banded force-label framework used in the main paper (Section~\ref{sec:limitations}). Under this framework:

\begin{itemize}[leftmargin=*,topsep=2pt,itemsep=2pt]
\item The SSVTP force labels used in Tables~\ref{tab:supporting} and \ref{tab:disentangle} carry p05/p50/p95 uncertainty bands under our framework (Section~\ref{sec:limitations}).
\item Public datasets (PoseIt, MMWand, FeelAnyForce, WireFishing-M, GelSight) provide measured force streams that serve as auxiliary grounding.
\item A $50$--$200$ sample SSVTP-style recapture with a calibrated F/T or load-cell sensor would provide an absolute-unit extension to the probe.
\end{itemize}

\section{Artifact Release Plan}
\label{app:hf_inventory}

\textbf{Artifact release.} The release package is organized around the unified visuo-tactile manifest (TVL-style, $\sim 827$\,MB), the alignment checkpoints (frozen-DINOv2 InfoNCE), the V$\to$T generators (Pix2Pix background-subtracted variants and the latent-diffusion variant), the SSVTP force-label sidecars (p05/p50/p95 bands), and SHA-256-verified canonical feature loaders. The artifact set totals $\sim 9$\,GB and includes metric summaries, generator checkpoints, force-label sidecars, unified tactile manifests, V$\to$T generator assets, and canonical loader code; the artifact registry, per-artifact README, and SHA-256 manifests are part of the public release package.

\section{Reproducibility}
\label{app:reproducibility}

\subsection{Planned artifacts}
The reproducibility bundle comprises alignment, physical-property probe, manipulation, generator, and V-T-L evidence-interface code; canonical feature manifests with per-file SHA-256 hashes; bootstrap-resample seeds; and probe scripts producing every main-paper number. The metric bundle contains the JSON result files, Qwen tactile-evidence reports, and trainable tactile-prefix/Q-former/LoRA reports.

\subsection{Reproduction order}
\begin{enumerate}[leftmargin=*,topsep=2pt,itemsep=2pt]
\item Restore SSVTP/TVL paired corpus and verify the SHA-256 manifest.
\item Extract DINOv2 features with the corrected background-subtracted preprocessing (Appendix~\ref{app:bgsub_ablation}).
\item Train the InfoNCE alignment ($120$ epochs, $\tau{=}0.07$, batch $256$, seed $42$).
\item Run the probe sweep (mass, density, hardness, force-label regression) with $5{,}000$-bootstrap.
\item Train the TACTO BC policy under matched V-only and V$+$T-aligned conditions; score $300$ rollouts.
\item Compare outputs against the headline numbers in Tables~\ref{tab:level2_primary}--\ref{tab:r112_gates}.
\end{enumerate}

\subsection{Release plan}
The artifacts described in Appendix~\ref{app:hf_inventory} are organized under public model-card and manifest structures. Per-artifact SHA-256 hashes and canonical loader code accompany the release package, while this appendix gives the reproduction order and manifest checks needed to interpret the reported numbers.

\section{Extended related work and additional references}
\label{app:extended_relwork}

This appendix consolidates references that motivate or contextualise TacGen but are not load-bearing for the main-paper positioning argument.

\paragraph{Multimodal representation lineage and vision-language pretraining.}
Beyond the contrastive vision-language and multimodal-binding works cited in Section~\ref{sec:related}, the broader multimodal representation literature spans early multimodal autoencoders and variational mixtures \citep{ngiam2011multimodal,wu2018multimodalvae,shi2019mmvae,sutter2020mmjsd}; multisensory scene inference and attention bottlenecks for partially observed modalities \citep{lim2019neural,nagrani2021attention}; and vision-language pretraining via two-stream transformers, contrastive alignment, and frozen-LM prefix interfaces \citep{lu2019vilbert,li2021albef,tsimpoukelli2021frozen}, with audio-extending contrastive variants \citep{guzhov2022audioclip}. These works motivate the shared-latent-space framing but do not target tactile contact as a physical-property signal.

\paragraph{Visuo-tactile shape, exploration, and generation.}
Vision-and-touch shape reconstruction \citep{smith2020shape} and active visuo-tactile exploration \citep{smith2021active} establish a complementary role for tactile evidence in 3D object understanding. Cross-modal self-supervised tactile representation learning \citep{zambelli2021rich} and visuo-tactile generative work \citep{yang2023visualscenes,dou2024tarf} treat touch as scene or geometric evidence rather than a physical-property signal. We retain the headline visuo-tactile dataset and representation references in the main text (Section~\ref{sec:related}) and consolidate this broader lineage here.

\paragraph{Self-supervised and contrastive backbone lineage.}
Our InfoNCE training follows the broader contrastive / self-supervised lineage, including contrastive predictive coding \citep{oord2018cpc}, SimCLR \citep{chen2020simclr}, MoCo \citep{he2020moco}, supervised contrastive learning \citep{khosla2020supcon}, contrastive multiview coding \citep{tian2020cmc}, and view-selection analyses \citep{tian2020views}. DINO-style self-distillation \citep{caron2021dino} motivates the DINOv2 family used as our canonical frozen backbone (Section~\ref{sec:methods}).

\paragraph{Tactile simulators surveyed.}
In addition to TACTO (used as our Level-1 manipulation testbed), we surveyed Taxim \citep{si2021taxim} and TacEx \citep{nguyen2024tacex}; simulator-realism comparisons are outside the scope of TacGen's representation-learning claims.

\paragraph{Generative-model background.}
The Pix2Pix and latent-diffusion variants we use build on conditional GAN image translation \citep{goodfellow2014gan,isola2017pix2pix}, score-based generative modelling \citep{song2021score}, latent diffusion \citep{rombach2022ldm}, classifier-free guidance \citep{ho2022classifierfree}, and masked-autoencoder pretraining \citep{he2022mae}.

\clearpage
\input{checklist}

\end{document}